%% file: iclr2021_conference.tex
\documentclass{article}
\pdfoutput=1 
\usepackage{iclr2021_conference,times}
\input{math_commands.tex}

\usepackage{hyperref}
\usepackage{url}
\usepackage{booktabs}
\usepackage{algorithm}
\usepackage{algorithmic}
\usepackage{graphics}
\usepackage{wrapfig}
\usepackage{mathtools}
\usepackage[capitalize]{cleveref}
\usepackage{subcaption}
\usepackage{enumitem}
\usepackage{bbm}
\usepackage{tabularx}
\usepackage{glossaries-extra}
\setabbreviationstyle[acronym]{long-short}
\glssetcategoryattribute{acronym}{nohyperfirst}{true}

\newacronym{vrnn}{VRNN}{variational recurrent neural network}
\newacronym{gmm}{GMM}{gaussian mixture model}
\newacronym{rkn}{RKN}{recurrent Kalman network}
\newacronym{aesmc}{AESMC}{auto-encoding sequential Monte Carlo}
\newacronym{elbo}{ELBO}{evidence lower bound}
\newacronym{cfvae}{CF-VAE}{conditional flow variational autoencoder}
\newacronym{ssm}{SSM}{state-space model}
\newacronym{vdm}{VDM}{variational dynamic mixtures}
\newacronym{ca}{CA}{cubature approximation}
\newacronym{sca}{SCA}{stochastic cubature approximation}
\newacronym{smc}{SMC}{sequential Monte Carlo}
\usepackage{amsthm}
\newtheorem{claim}{Claim}
\newtheorem*{claim-non}{Claim}
\title{Variational Dynamic Mixtures}

\iclrfinalcopy 

\author{
	Chen Qiu\thanks{Correspondence to:
		\texttt{chen.qiu@de.bosch.com}}\\
	Bosch Center for AI 
	\And
	Stephan Mandt\\
	UC Irvine
	\And
	Maja Rudolph\\
	Bosch Center for AI
}

\begin{document}
\maketitle

\begin{abstract}
Deep probabilistic time series forecasting models have become an integral part of machine learning. While several powerful generative models have been proposed, we provide evidence that their associated inference models are oftentimes too limited and cause the generative model to predict mode-averaged dynamics. Mode-averaging is problematic since many real-world sequences are highly multi-modal, and their averaged dynamics are unphysical (e.g., predicted taxi trajectories might run through buildings on the street map). To better capture multi-modality, we develop \gls{vdm}: a new variational family to infer sequential latent variables. 
The \gls{vdm} approximate posterior at each time step is a mixture density network, whose parameters come from propagating multiple samples through a recurrent architecture. This results in an expressive  multi-modal posterior approximation. In an empirical study, we show that \gls{vdm} outperforms competing approaches on highly multi-modal datasets from different domains. 
\end{abstract}

\glsresetall
\input{introduction}

\input{related_work}

\input{methods}

\input{experiments}

\input{conclusion}
\subsubsection*{Acknowledgments}
Stephan Mandt acknowledges support by DARPA (Contract No. HR001120C0021), the National Science Foundation under Grants 1928718, 2003237 and 2007719, as well as Intel and Qualcomm.
\bibliography{iclr2021_conference}
\bibliographystyle{iclr2021_conference}
\newpage
\appendix
\input{supplement}

\end{document}

%% file: math_commands.tex
\usepackage{amsmath,amsfonts,bm}

\def\eqref#1{equation~\ref{#1}}

\def\1{\bm{1}}

\DeclareMathAlphabet{\mathsfit}{\encodingdefault}{\sfdefault}{m}{sl}
\SetMathAlphabet{\mathsfit}{bold}{\encodingdefault}{\sfdefault}{bx}{n}

\DeclareMathOperator*{\argmax}{arg\,max}

%% file: introduction.tex
\section{Introduction}
\label{sec:intro}

Making sense of time series data is an important challenge in various domains, including ML for climate change.
One important milestone to reach the climate goals is to significantly reduce the $\text{CO}_2$ emissions from mobility \citep{rogelj2016paris}. Accurate forecasting models of typical driving behavior and of typical pollution levels over time
can help both lawmakers and automotive engineers to develop solutions for cleaner mobility. In these applications, no accurate physical model of the entire dynamic system is known or available. Instead, data-driven models,
specifically deep probabilistic time series models, can be used to solve the necessary tasks including forecasting.

\begin{figure}[t]
	\centering
	\begin{subfigure}[b]{0.16\textwidth}
	\includegraphics[width=\textwidth]{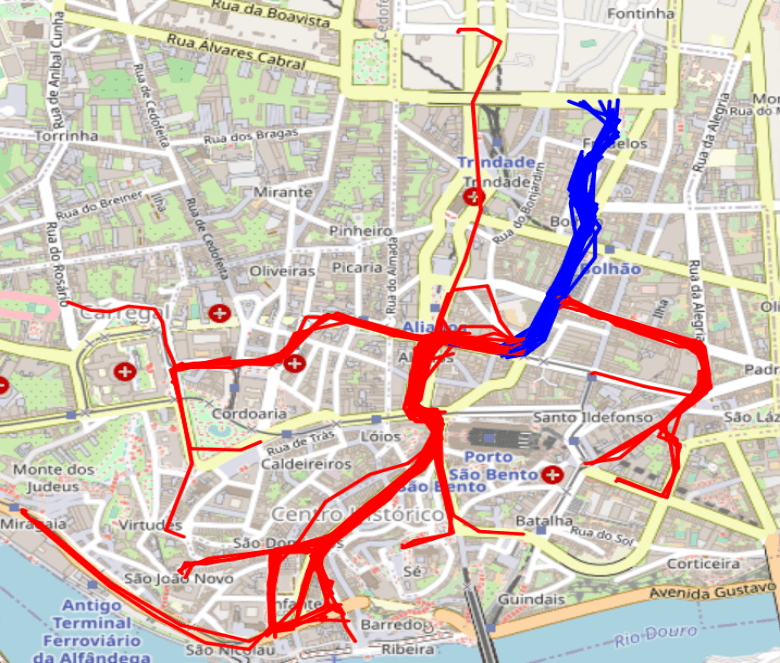}
	\end{subfigure}
	\begin{subfigure}[b]{0.16\textwidth}
		\includegraphics[width=\textwidth]{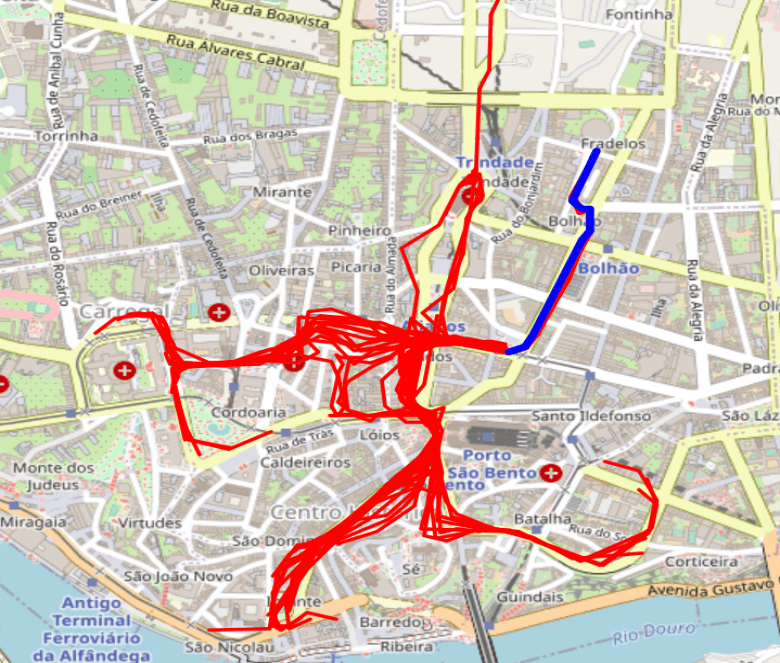}
	\end{subfigure}
	\begin{subfigure}[b]{0.16\textwidth}
		\includegraphics[width=\textwidth]{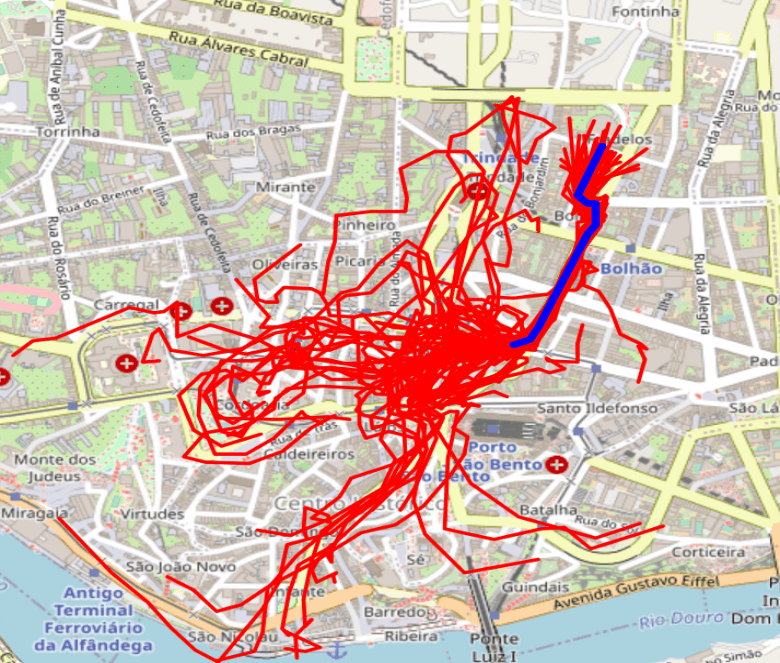}
	\end{subfigure}
	\begin{subfigure}[b]{0.16\textwidth}
		\includegraphics[width=\textwidth]{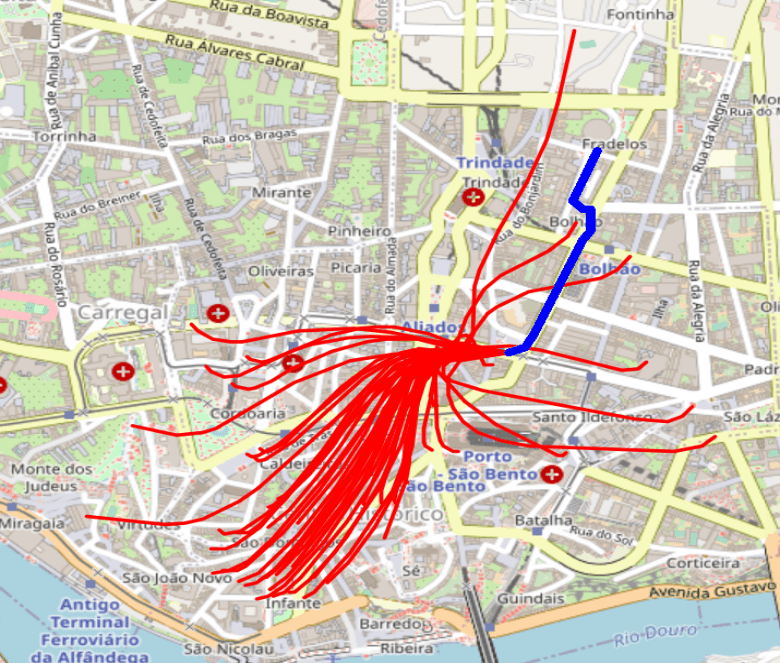}
	\end{subfigure}
		\begin{subfigure}[b]{0.16\textwidth}
		\includegraphics[width=\textwidth]{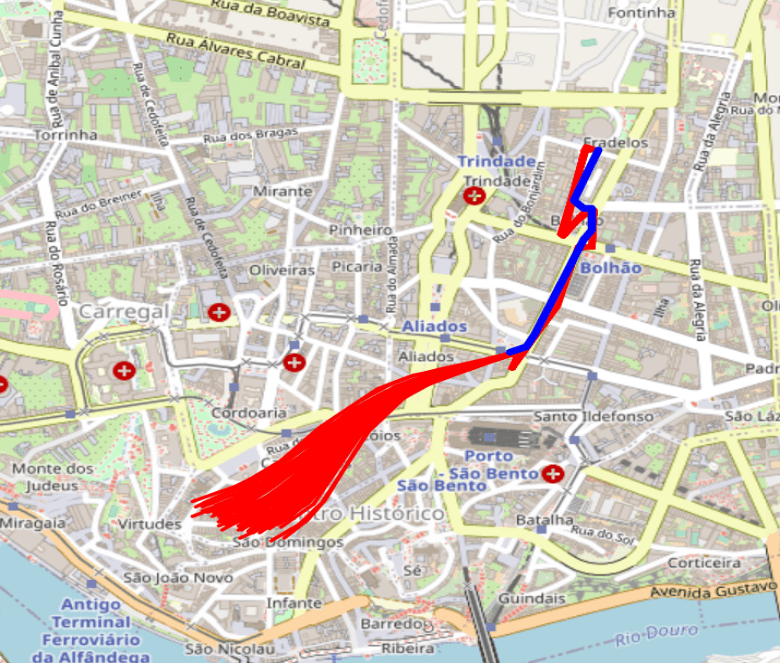}
	\end{subfigure}
	\begin{subfigure}[b]{0.16\textwidth}
		\includegraphics[width=\textwidth]{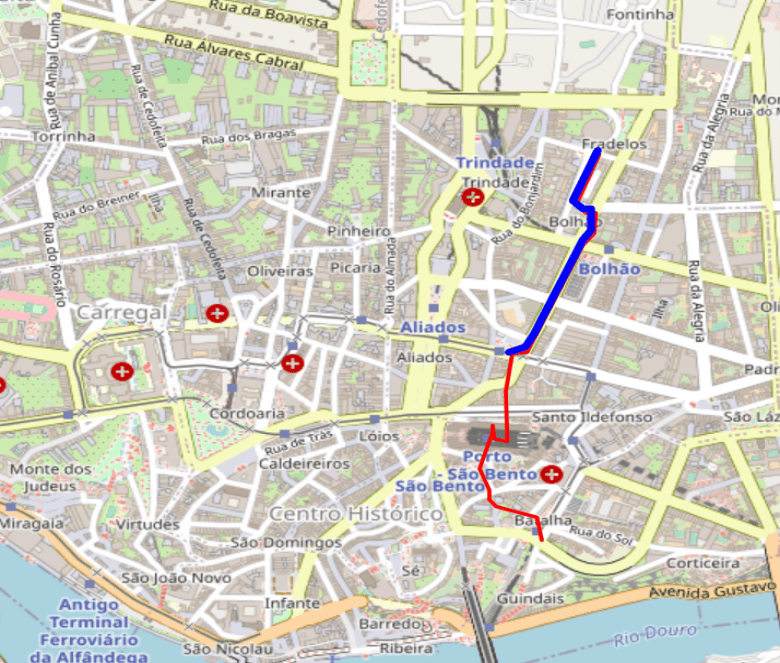}
	\end{subfigure}\\
	\begin{subfigure}[b]{0.16\textwidth}
		\includegraphics[width=\textwidth]{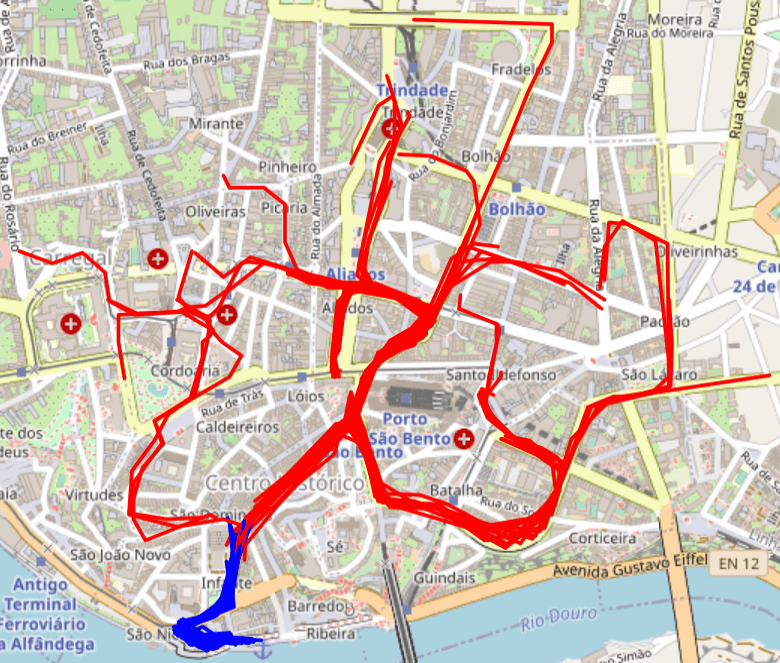}
		\caption{Taxi Data}
		\label{fig:dat}
	\end{subfigure}
		\begin{subfigure}[b]{0.16\textwidth}
		\includegraphics[width=\textwidth]{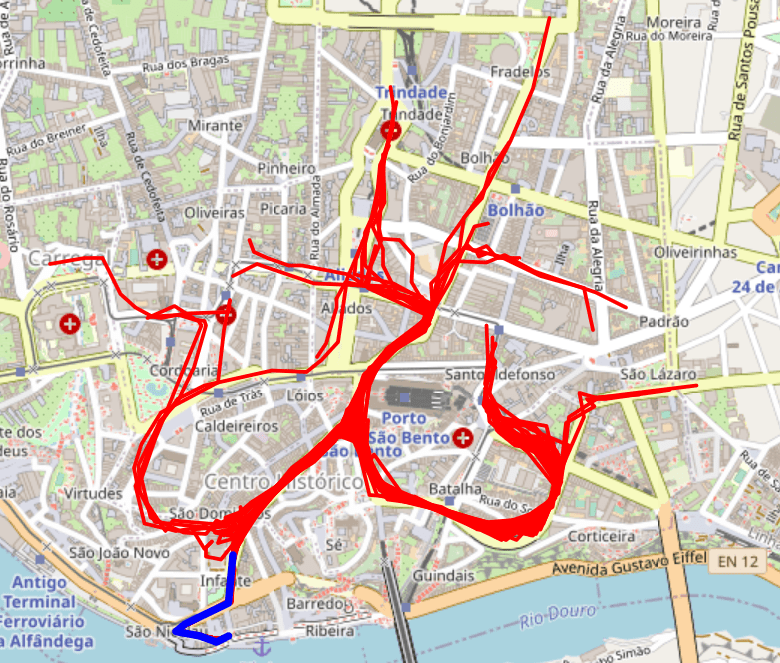}
		\caption{\acrshort{vdm} (ours)}
		\label{fig:ours}
	\end{subfigure}
	\begin{subfigure}[b]{0.16\textwidth}
		\includegraphics[width=\textwidth]{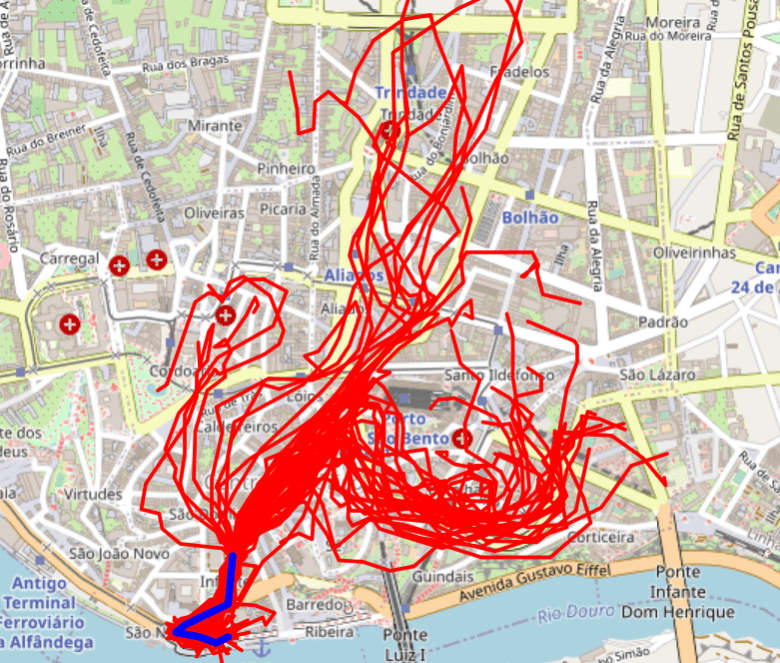}
		\caption{\acrshort{aesmc} 
		}
		\label{fig:aesmc}
	\end{subfigure}
	\begin{subfigure}[b]{0.16\textwidth}
		\includegraphics[width=\textwidth]{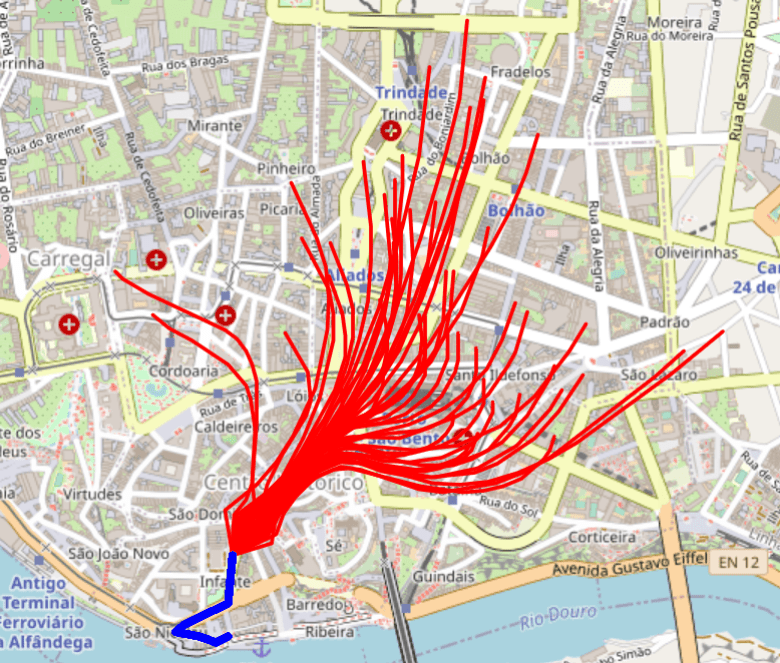}
		\caption{\acrshort{cfvae} 
		}
		\label{fig:cfvae}
	\end{subfigure}
	\begin{subfigure}[b]{0.16\textwidth}
		\includegraphics[width=\textwidth]{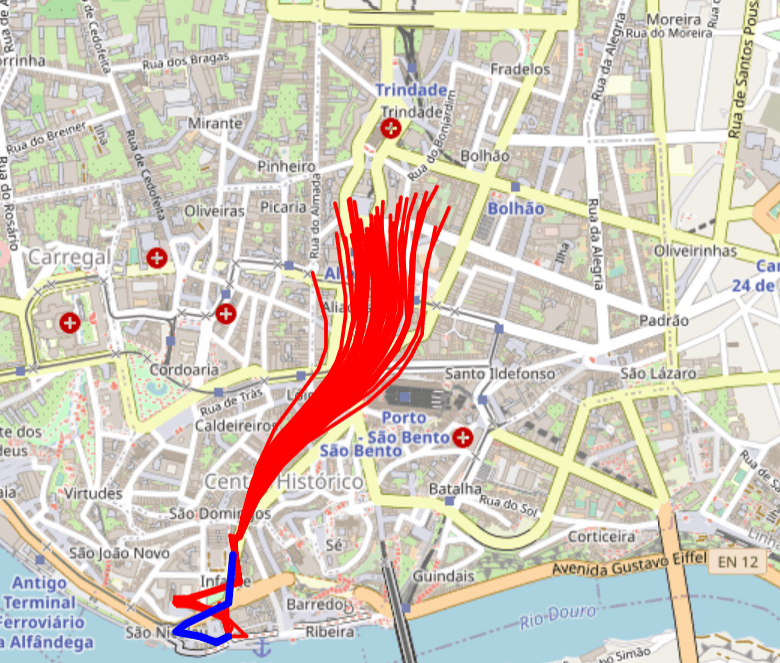}
		\caption{\acrshort{vrnn} 
		}
		\label{fig:lstm}
	\end{subfigure}
		\begin{subfigure}[b]{0.16\textwidth}
		\includegraphics[width=\textwidth]{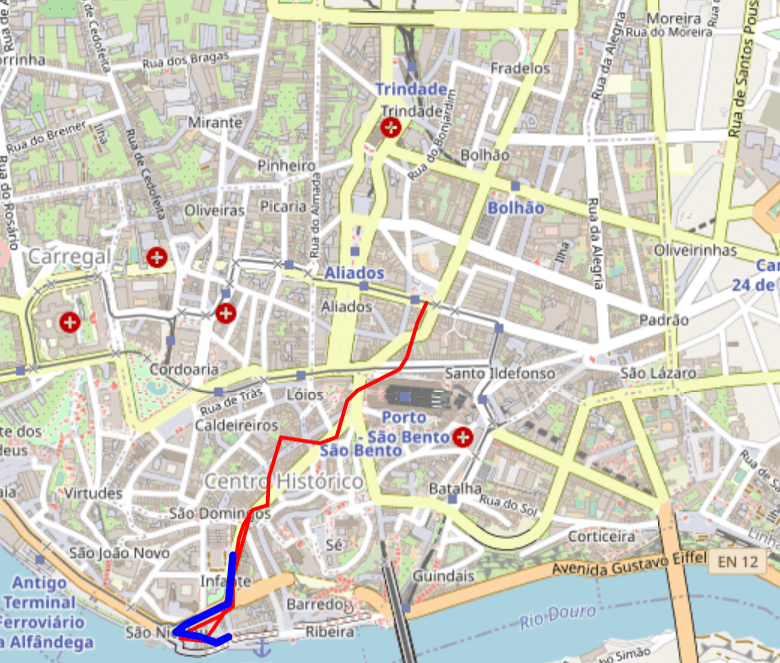}
		\caption{\acrshort{rkn} 
		}
		\label{fig:rkn}
	\end{subfigure}
	
	\caption{Forecasting taxi trajectories is challenging due to the highly multi-modal nature of the data (\cref{fig:dat}). \gls{vdm} (\cref{fig:ours}) succeeds in generating diverse plausible predictions (red), based the beginning of a trajectory (blue). The other methods, \acrshort{aesmc} \citep{le2018auto}, \acrshort{cfvae} \citep{bhattacharyya2019conditional}, \acrshort{vrnn} \cite{chung2015recurrent}, \acrshort{rkn} \cite{becker2019recurrent}, suffer from mode averaging. }
	\label{fig:taxi_traj}
	\vspace{-10pt}
\end{figure}

The dynamics in such data can be highly multi-modal.
At any given part of the observed sequence, there might be multiple distinct continuations of the data that are plausible, but the average of these behaviors is unlikely, or even physically impossible. Consider for example a dataset of taxi trajectories\footnote{https://www.kaggle.com/crailtap/taxi-trajectory}. In each row of \cref{fig:dat}, we have selected 50 routes from the dataset with similar starting behavior (blue). Even though these routes are quite similar to each other in the first 10 way points, the continuations of the trajectories (red) can exhibit quite distinct behaviors and lead to points on any far edge of the map. The trajectories follow a few main traffic arteries, these could be considered the main modes of the data distribution. We would like to learn a generative model of the data, that based on some initial way points, can forecast plausible continuations for the trajectories.

Many existing methods make restricting modeling assumptions such as Gaussianity to make learning tractable and efficient. 
But trying to capture the dynamics through unimodal distributions can lead either to ``over-generalization'', (i.e. putting probability mass in spurious regions) or on focusing only on the dominant mode and thereby neglecting important structure of the data.
Even neural approaches, with very flexible generative models can fail to fully capture this multi-modality because their capacity is often limited through the assumptions of their {\em inference model}.
To address this, we develop \gls{vdm}. Its generative process is a sequential latent variable model. The main novelty is a new multi-modal variational family which makes learning and inference multi-modal yet tractable.
In summary, our contributions are

\begin{itemize}[topsep=0pt,leftmargin=*]
\itemsep0em 
\item {\bf A new inference model.} We establish a new type of variational family for variational inference of sequential latent variables. By successively marginalizing over previous latent states, the procedure can be efficiently carried-out in a single forward pass and induces a multi-modal posterior approximation. We can see in \cref{fig:ours}, that \gls{vdm} trained on a dataset of taxi trajectories produces forecasts with the desired multi-modality while other methods overgeneralize. 
\item {\bf An evaluation metric for multi-modal tasks.} The negative log-likelihood measures predictive accuracy but neglects an important aspect of multi-modal forecasts -- sample diversity. 
In \cref{sec:exp}, we derive a score based on the Wasserstein distance \citep{villani2008optimal} which evaluates both sample quality and diversity. 
This metric complements our evaluation based on log-likelihoods.
\item {\bf An extensive empirical study.} in \cref{sec:exp}, we use \gls{vdm} to study various datasets, including a synthetic data with four modes, a stochastic Lorenz attractor, the taxi trajectories, and a U.S. pollution dataset with the measurements of various pollutants over time. We illustrate \gls{vdm}'s ability in modeling multi-modal dynamics, and provide quantitative comparisons to other methods showing that \gls{vdm}
compares favorably to previous work.
\end{itemize}

%% file: related_work.tex
\section{Related Work}
\label{sec:related_work}
\paragraph*{Neural recurrent models.} Recurrent neural networks (RNNs) such as LSTMs \citep{hochreiter1997long} and GRUs \citep{chung2014empirical} have proven successful on many time series modeling tasks. However, as deterministic models they cannot capture uncertainties in their dynamic predictions. Stochastic RNNs make these sequence models non-deterministic \citep{chung2015recurrent,fraccaro2016sequential,gemici2017generative,li2018disentangled}. For example, the \gls{vrnn} \citep{chung2015recurrent} enables multiple stochastic forecasts due to its stochastic transition dynamics. An extension of \gls{vrnn} \citep{goyal2017z} uses an auxiliary cost to alleviate the KL-vanishing problem. It improves on \gls{vrnn} inference by forcing the latent variables to also be predictive of future observations. Another line of  related methods rely on particle filtering \citep{naesseth2018variational,le2018auto,hirt2019scalable} and in particular \gls{smc} to improve the evidence lower bound. 
In contrast, \gls{vdm} adopts an explicitly multi-modal posterior approximation. Another \gls{smc}-based work \citep{saeedi2017variational} employs search-based techniques for multi-modality but is limited to models with finite discrete states.
Recent works \citep{schmidt2018deep,schmidt2019autoregressive,ziegler2019latent} use normalizing flows in the latent space to model the transition dynamics. 
A normalizing flow requires many layers to transform its base distribution into a truly multi-modal distribution in practice. In contrast, mixture density networks (as used by \gls{vdm}) achieve multi-modality by mixing only one layer of neural networks. A task orthogonal to multi-modal inference is learning disentangled representations. Here too, mixture models are used \citep{chen2016infogan,li2017infogail}. These papers use discrete variables and a mutual information based term to disentangle different aspects of the data.

VAE-like models \citep{bhattacharyya2018accurate,bhattacharyya2019conditional} and GAN-like models \citep{sadeghian2019sophie,kosaraju2019social} only have global, time independent latent variables. Yet, they show good results on various tasks, including forecasting. 
With a deterministic decoder, these models focus on average dynamics and don't capture local details (including multi-modal transitions) very well. Sequential latent variable models are described next.

\paragraph*{Deep state-space models.} Classical \Glspl{ssm} are popular due to their tractable inference and interpretable predictions.  Similarly, \emph{deep} \glspl{ssm} with locally linear transition dynamics enjoy tractable inference \citep{karl2016deep,fraccaro2017disentangled,rangapuram2018deep,becker2019recurrent}. However, these models are often not expressive enough to capture complex (or highly multi-modal) dynamics. Nonlinear deep \glspl{ssm} \citep{krishnan2017structured,zheng2017state,doerr2018probabilistic,de2019gru,gedon2020deep} are more flexible. Their inference is often no longer tractable and requires variational approximations. Unfortunately, in order for the inference model to be tractable, the variational approximations are often simplistic and don't approximate multi-modal posteriors well with negative effects on the trained models. Multi-modality can be incorporated via additional discrete switching latent variables, such as recurrent switching linear dynamical systems \citep{linderman2017bayesian,nassar2018tree,becker2019switching}. However, these discrete states make inference more involved.  

%% file: methods.tex
\section{Variational Dynamic Mixtures}
\label{sec:vdm}
We develop \gls{vdm}, a new sequential latent variable model for multi-modal dynamics. Given sequential observations $\mathbf{x}_{1:T} =(\mathbf{x}_1,\ldots,\mathbf{x}_T)$, \gls{vdm} assumes that the underlying dynamics are governed by latent states $\mathbf{z}_{1:T} =(\mathbf{z}_1,\ldots,\mathbf{z}_T)$. We first present the generative process and the multi-modal inference model of \gls{vdm}. We then derive a new variational objective that encourages multi-modal posterior approximations and we explain how it is regularized via hybrid-training. Finally, we introduce a new sampling method used in the inference procedure.
\paragraph*{Generative model.}
The generative process consists of a transition model and an emission model. The transition model $p(\mathbf{z}_t \mid \mathbf{z}_{< t})$ describes the temporal evolution of the latent states and the emission model $p(\mathbf{x}_t \mid \mathbf{z}_{\leq t})$ maps the states to observations. 
We assume they are parameterized by two separate neural networks, the transition network $\phi^{tra}$ and the emission network $\phi^{dec}$. To give the model the capacity to capture longer range temporal correlations we parametrize the transition model with a recurrent architecture $\phi^{\text{GRU}}$ \citep{auger2016state,zheng2017state} such as a GRU \citep{chung2014empirical}. The latent states $\mathbf{z}_t$ are sampled recursively from
\begin{align}
\mathbf{z}_t  \mid \mathbf{z}_{<t}
\sim \mathcal{N}(\mathbf{\mu}_{0,t}, \mathbf{\sigma}_{0,t}^2\mathbb{I}), \quad \text{where} \quad [\mathbf{\mu}_{0,t}, \mathbf{\sigma}_{0,t}^2] = \phi^{tra}(\mathbf{h}_{t-1}),\, \mathbf{h}_{t-1} = \phi^{\text{GRU}}(\mathbf{z}_{t-1}, \mathbf{h}_{t-2}),
\label{eqn:transition}
\end{align}
and are then decoded such that the observations can be sampled from the emission model,
\begin{align}
\mathbf{x}_t \mid \mathbf{z}_{\leq t}
\sim \mathcal{N}(\mathbf{\mu}_{x,t}, \mathbf{\sigma}_{x,t}^2\mathbb{I}), \quad \text{where} \quad [\mathbf{\mu}_{x,t}, \mathbf{\sigma}_{x,t}^2] = \phi^{dec}(\mathbf{z}_{t},\mathbf{h}_{t-1}).
\label{eqn:emission}
\end{align}
This generative process is similar to \citep{chung2015recurrent}, though we did not incorporate autoregressive feedback due to its negative impact on long-term generation \citep{ranzato2015sequence,lamb2016professor}. The competitive advantage of \gls{vdm} comes from a more expressive inference model.
\paragraph*{Inference model.}
\begin{figure}[t]
\centering
	\begin{subfigure}[b]{0.34\textwidth}
		\includegraphics[width=\textwidth]{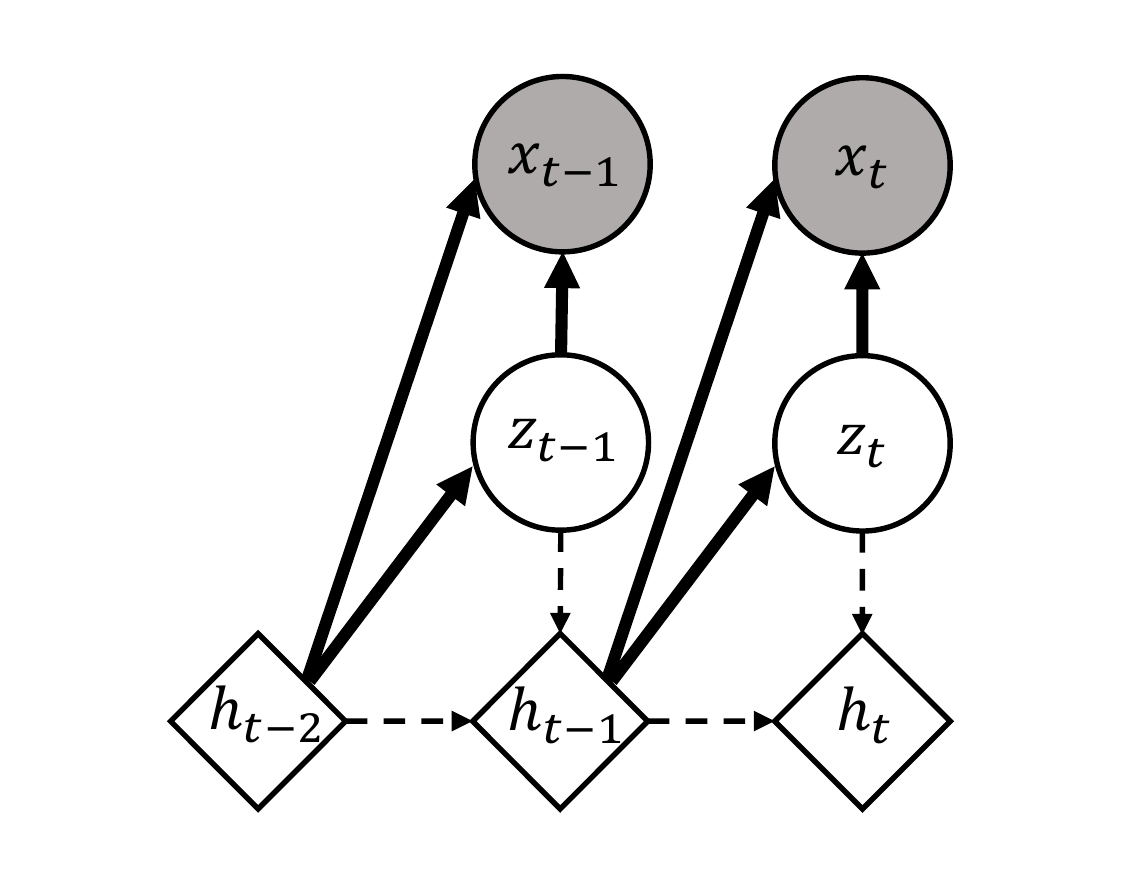}
	\caption{Generation (\cref{eqn:transition,eqn:emission})}
	\label{fig:generation}
	\end{subfigure}
	\begin{subfigure}[b]{0.53\textwidth}
		\includegraphics[width=\textwidth]{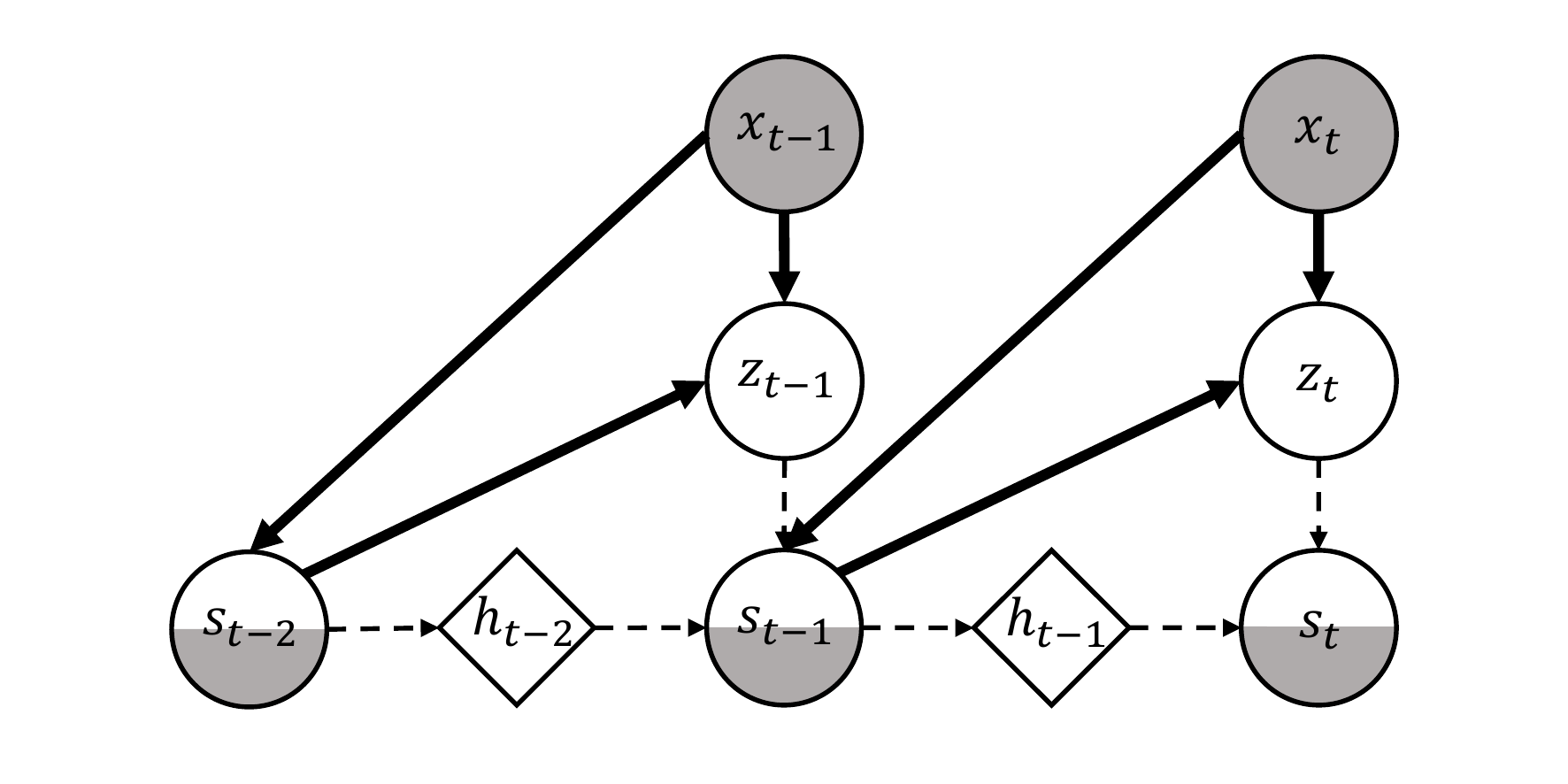}
	\caption{Inference (\cref{eqn:qtilde,eqn:inference,eqn:expected_h})}
	\label{fig:inference}
	\end{subfigure}
		\caption{Graphical illustrations of \gls{vdm}. Dashed lines denote deterministic dependencies such as transformations, marginalization, or computing the mean, as explained in the main text, while bold lines denote stochastic dependencies. The half-shaded node for $\mathbf{s}_t$ indicates that $\mathbf{s}_t$ is being marginalized out as opposed to conditioned on.}
		\label{fig:schematic_illustration}
		\vspace{-10pt}
\end{figure}
\gls{vdm} is based on a new procedure for multi-modal inference. The main idea is that to approximate the posterior at time $t$, we can use the posterior approximation of the previous time step and exploit the generative model’s transition model $\phi^{\text{GRU}}$. This leads to a sequential inference procedure. We first use the forward model to transform the approximate posterior at time $t-1$ into a distribution at time $t$. In a second step, we use samples from the resulting transformed distribution and combine each sample with data evidence $\mathbf{x}_t$, where every sample parameterizes a Gaussian mixture component. As a result, we obtain a multi-modal posterior distribution that depends on data evidence, but also on the previous time step’s posterior. 

In more detail, for every $\mathbf{z}_{t}$, we define its corresponding recurrent state as the transformed random variable $\mathbf{s}_{t} = \phi^{\text{GRU}}(\mathbf{z}_{t}, \mathbf{h}_{t-1})$, using a deterministic hidden state $\mathbf{h}_{t-1} = \mathbb{E}\left[\mathbf{s}_{t-1}\right]$. The variational family of \gls{vdm} is defined as follows:
\begin{align}
\label{eqn:posterior}
 q(\mathbf{z}_{1:T} \mid \mathbf{x}_{1:T}) = \prod_{t=1}^T q(\mathbf{z}_{t} \mid \mathbf{x}_{\leq t}) = 
 \prod_{t=1}^T \int q(\mathbf{z}_{t} \mid \mathbf{s}_{t-1},\mathbf{x}_{t}) q(\mathbf{s}_{t-1} \mid \mathbf{x}_{\leq t}) \mathrm{d}\mathbf{s}_{t-1}.
\end{align} 
\cite{chung2015recurrent} also use a sequential inference procedure, but without considering the distribution of $\mathbf{s}_t$. Only a single sample is propagated through the recurrent network and all other information about the distribution of previous latent states $\mathbf{z}_{<t}$ is lost. In contrast, \gls{vdm} explicitly maintains $\mathbf{s}_t$ as part of the inference model. Through marginalization, the entire distribution is taken into account for inferring the next state $\mathbf{z}_t$. Beyond the factorization assumption and the marginal consistency constraint of \cref{eqn:posterior}, the variational family of \gls{vdm} needs two more choices to be fully specified; First, one has to choose the parametrizations of $q(\mathbf{z}_{t} \mid \mathbf{s}_{t-1},\mathbf{x}_{t})$ and $q(\mathbf{s}_{t-1} \mid \mathbf{x}_{\leq t})$ and second, one has to choose a sampling method to approximate the marginalization in \cref{eqn:posterior}. These choices determine the resulting factors $q(\mathbf{z}_{t} \mid \mathbf{x}_{\leq t})$ of the variational family.

We assume that the variational distribution of the recurrent state factorizes as $q(\mathbf{s}_{t-1} \mid \mathbf{x}_{\leq t}) = \omega(\mathbf{s}_{t-1}, \mathbf{x}_{t})\Tilde{q}(\mathbf{s}_{t-1} \mid \mathbf{x}_{<t})$, i.e. it is the distribution of the recurrent state given the past observation\footnote{ $\Tilde{q}(\mathbf{s}_{t-1} \mid \mathbf{x}_{<t})$ is the distribution obtained by transforming the previous $z_{t-1} \sim q(\mathbf{z}_{t-1} | \mathbf{x}_{<t} )$ through the RNN. It can be expressed analytically using the Kronecker $\delta$ to compare whether the stochastic variable $\mathbf{s}_{t-1}$ equals the output of the RNN: $\Tilde{q}(\mathbf{s}_{t-1} \mid \mathbf{x}_{< t})\propto \int \delta(\mathbf{s}_{t-1}-\phi^{\text{GRU}}(\mathbf{z}_{t-1},\mathbf{h}_{t-2}))q(\mathbf{z}_{t-1} \mid \mathbf{x}_{t-1}, \lambda_{t-1}) \mathrm{d}\mathbf{z}_{t-1}$.}, re-weighted by a weighting function $\omega(\mathbf{s}_{t-1}, \mathbf{x}_{t})$ which involves only the current observations. For \gls{vdm}, we only need samples from $\Tilde{q}(\mathbf{s}_{t-1} \mid \mathbf{x}_{<t})$, which are obtained by sampling from the previous posterior approximation  $q(\mathbf{z}_{t-1} \mid \mathbf{x}_{<t})$ and transforming the sample with the RNN,
\begin{align}
\label{eqn:qtilde}
 \mathbf{s}_{t-1}^{(i)} \sim \Tilde{q}(\mathbf{s}_{t-1} \mid \mathbf{x}_{<t})  \quad \text{equiv. to} \quad \mathbf{s}_{t-1}^{(i)} = \phi^{\text{GRU}}(\mathbf{z}_{t-1}^{(i)}, \mathbf{h}_{t-2}),\quad \mathbf{z}_{t-1}^{(i)} \sim q(\mathbf{z}_{t-1} \mid \mathbf{x}_{<t}),
\end{align}
where $i$ indexes the samples. The RNN $\phi^{\text{GRU}}$ has the same parameters as in the generative model.

Augmenting the variational model with the recurrent state has another advantage; approximating the marginalization in \cref{eqn:posterior} with $k$ samples from $q(\mathbf{s}_{t-1} \mid \mathbf{x}_{\leq t})$ and choosing a Gaussian parametrization for $q(\mathbf{z}_{t} \mid \mathbf{s}_{t-1},\mathbf{x}_{ t})$ results in a q-distribution $q(\mathbf{z}_{t} \mid \mathbf{x}_{\leq t})$ that resembles a mixture density network \citep{bishop2006pattern}, which is a convenient choice to model multi-modal distributions. 
\begin{align}
\label{eqn:inference}
q(\mathbf{z}_{t} \mid \mathbf{x}_{\leq t}) = \sum_i^k \omega_t^{(i)} \mathcal{N}(\mathbf{\mu}_{z,t}^{(i)}, \mathbf{\sigma}_{z,t}^{(i)2}\mathbb{I}),\qquad
[\mathbf{\mu}_{z,t}^{(i)}, \mathbf{\sigma}_{z,t}^{(i)2}] = \phi^{inf}( \mathbf{s}_{t-1}^{(i)},\mathbf{x}_{t}).
\end{align} 
We assume $q(\mathbf{z}_{t} \mid \mathbf{s}_{t-1},\mathbf{x}_{t})$ to be Gaussian and use an inference network $\phi^{inf}$ to model the effect of the observation $\mathbf{x}_{ t}$ and recurrent state $\mathbf{s}_{t-1}$ on the mean and variance of the mixture components. 

The mixture weights $\omega_t^{(i)} \coloneqq \omega(\mathbf{s}_{t-1}^{(i)}, \mathbf{x}_{t})/k$ come from the variational distribution $q(\mathbf{s}_{t-1} \mid \mathbf{x}_{\leq t})=\omega(\mathbf{s}_{t-1}, \mathbf{x}_{t})\Tilde{q}(\mathbf{s}_{t-1} \mid \mathbf{x}_{<t})$ and importance sampling\footnote{the $\omega$ adjusts for using samples from $\Tilde{q}(\mathbf{s}_{t-1} \mid \mathbf{x}_{<t})$ when marginalizing over $\omega(\mathbf{s}_{t-1}, \mathbf{x}_{t})\Tilde{q}(\mathbf{s}_{t-1} \mid \mathbf{x}_{<t})$}. We are free to choose how to parametrize the weights, as long as all variational distributions are properly normalized. Setting
\begin{align}
\label{eqn:pi}
    \omega_t^{(i)}&=\omega(\mathbf{s}_{t-1}^{(i)}, \mathbf{x}_{t})/k \coloneqq \mathbbm{1}(i=\argmax_{j} p(\mathbf{x}_t\mid \mathbf{h}_{t-1}=\mathbf{s}_{t-1}^{(j)})),
\end{align}
achieves this.
In \cref{sec:omega_appendix}, we explain this choice with importance sampling and in \cref{sec:appendix_results}, we compare the performance of \gls{vdm} under alternative variational choices for the weights. 

In the next time-step, plugging the variational distribution $q(\mathbf{z}_{t} \mid \mathbf{x}_{\leq t})$ into \cref{eqn:qtilde} yields the next distribution over recurrent states $\Tilde{q}(\mathbf{s}_{t} \mid \mathbf{x}_{\leq t})$. For this, the expected recurrent state $\mathbf{h}_{t-1}$ is required. We approximate the update using the same $k$ samples (and therefore the same weights) as in \cref{eqn:inference}.
\begin{align}
\label{eqn:expected_h}
    \mathbf{h}_{t-1} = \mathbb{E}[\mathbf{s}_{t-1}] = \int \mathbf{s}_{t-1}\, q(\mathbf{s}_{t-1} \mid \mathbf{x}_{\leq t}) \mathrm{d}\mathbf{s}_{t-1} \approx \sum_i^k \omega_t^{(i)}\mathbf{s}_{t-1}^{(i)}.
\end{align}
A schematic view of the generative and inference model of \gls{vdm} is shown in \cref{fig:schematic_illustration}. In summary, the inference model of \gls{vdm} alternates between \cref{eqn:qtilde,eqn:inference,eqn:pi,eqn:expected_h}. Latent states are sampled from the posterior approximation of the previous time-step and transformed  by \cref{eqn:qtilde} into samples of the recurrent state of the RNN. These are then combined with the new observation $\mathbf{x}_t$ to produce the next variational posterior \cref{eqn:inference} and the expected recurrent state is updated (\cref{eqn:expected_h}). These are then used in \cref{eqn:qtilde} again.
Approximating the marginalization in \cref{eqn:posterior} with a single sample, recovers the inference model of \gls{vrnn} \citep{chung2015recurrent}, and fails in modeling multi-modal dynamics as shown in \cref{fig:toy}. In comparison, \gls{vdm}'s approximate marginalization over the recurrent states with multiple samples succeeds in modeling multi-modal dynamics.

\paragraph*{Variational objective.} We develop an objective to optimize the variational parameters of \gls{vdm} $\phi = [\phi^{tra}, \phi^{dec}, \phi^{\text{GRU}}, \phi^{inf}]$. The \gls{elbo} at each time step is
\begin{align}
 \label{eqn:elbo}
     \mathcal{L}_{\mathrm{ELBO}}(\mathbf{x}_{\leq t},\phi)
&:=\frac{1}{k}\sum_i^k \omega(\mathbf{s}_{t-1}^{(i)},\mathbf{x}_{t}) \mathbb{E}_{q(\mathbf{z}_{t}\mid \mathbf{s}_{t-1}^{(i)},\mathbf{x}_{ t})}\left[\log p(\mathbf{x}_{t} \mid \mathbf{z}_{t},\mathbf{h}_{t-1} = \mathbf{s}_{t-1}^{(i)})\right]\nonumber\\
   & +  \frac{1}{k}\sum_i^k \omega(\mathbf{s}_{t-1}^{(i)},\mathbf{x}_{t}) \mathbb{E}_{q(\mathbf{z}_{t}\mid \mathbf{s}_{t-1}^{(i)},\mathbf{x}_{ t})}\left[\log\frac{p(\mathbf{z}_{t} \mid  \mathbf{h}_{t-1} = \mathbf{s}_{t-1}^{(i)})}{q(\mathbf{z}_{t}\mid \mathbf{s}_{t-1}^{(i)},\mathbf{x}_{t})}\right]\nonumber\\
    &- \frac{1}{k}\sum_i^k \omega(\mathbf{s}_{t-1}^{(i)},\mathbf{x}_{t})\left[\log \omega(\mathbf{s}_{t-1}^{(i)},\mathbf{x}_{t})+\mathbf{C}\right] 
\end{align}

\begin{claim}
\label{theo:claim}
The \gls{elbo} in \cref{eqn:elbo} is a lower bound on the log evidence $\log p(\mathbf{x}_{t} \mid \mathbf{x}_{<t})$,
\begin{equation}
\label{eqn:bound}
        \log p(\mathbf{x}_{t} \mid \mathbf{x}_{<t})\geq \mathcal{L}_{\mathrm{ELBO}}(\mathbf{x}_{\leq t},\phi), \qquad \textit{(see proof in \cref{sec:elbo_appendix})}\,.
\end{equation}
\end{claim}

In addition to the \gls{elbo}, the objective of \gls{vdm} has two regularization terms,
\begin{align}
    \label{eqn:full_objective}
  \mathcal{L}_{\mathrm{VDM}}(\phi)
  = \sum_{t=1}^T \mathbb{E}_{p_{\mathcal{D}}}\left[-\mathcal{L}_{\mathrm{ELBO}}(\mathbf{x}_{\leq t},\phi)- \omega_1\mathcal{L}_{pred}(\mathbf{x}_{\leq t},\phi) \right]
  + \omega_2\mathcal{L}_{adv}(\mathbf{x}_{\leq t},\phi)\,.
\end{align}

In an ablation study in \cref{sec:study_regularizations}, we compare the effect of including and excluding the regularization terms in the objective. 
\gls{vdm} is competitive without these terms, but we got the strongest results by setting $\omega_{1,2}=1$ (this is the only nonzero value we tried. This hyperparameter could be tuned even further.)
The first regularization term $\mathcal{L}_{pred}$, encourages the variational posterior (from the previous time step) to produce samples that maximize the predictive likelihood,
\begin{align}
\label{eqn:pred_loss}
\mathcal{L}_{pred}(\mathbf{x}_{\leq t},\phi) 
=\log \mathbb{E}_{q(\mathbf{s}_{t-1} \mid \mathbf{x}_{<t})}\left[p(\mathbf{x}_{t} \mid \mathbf{s}_{t-1},\mathbf{x}_{<t})\right]\approx\log\frac{1}{k}\sum_{i}^{k}p(\mathbf{x}_{t} \mid \mathbf{s}_{t-1}^{(i)})\,.
\end{align} 
This regularization term is helpful to improve the prediction performance, since it depends on the predictive likelihood of samples, which isn't involved in the \gls{elbo}. 
The second optional regularization term $\mathcal{L}_{adv}$ (\cref{eqn:adv_loss}) is based on ideas from hybrid adversarial-likelihood training \citep{grover2018flow,lucas2019adaptive}. These training strategies have been developed for generative models of images to generate sharper samples while avoiding ``mode collapse''.
We adapt these ideas to generative models of dynamics. The adversarial term $\mathcal{L}_{adv}$ uses a forward KL-divergence, which enables ``quality-driven training'' to discourage probability mass in spurious areas.
\begin{align}
\label{eqn:adv_loss}
\mathcal{L}_{adv}(\mathbf{x}_{\leq t},\phi)  
= \mathcal{D}_{\mathrm{KL}}(p(\mathbf{x}_t\mid \mathbf{x}_{<t}) \| p_{\mathcal{D}}(\mathbf{x}_t\mid \mathbf{x}_{<t})) 
= \mathbb{E}\left[ \log p(\mathbf{x}_t \mid \mathbf{x}_{<t}) -  \log p_{\mathcal{D}}(\mathbf{x}_t \mid \mathbf{x}_{<t})\right]
\end{align}
The expectation is taken w.r.t. $p(\mathbf{x}_t \mid \mathbf{x}_{<t})$. The true predictive distribution $p_{\mathcal{D}}(\mathbf{x}_t \mid \mathbf{x}_{<t})$ is unknown. Instead, we can train the generator of a conditional GAN \citep{mirza2014conditional}, while assuming an optimal discriminator. As a result, we optimize \cref{eqn:adv_loss} in an adversarial manner, conditioning on $\mathbf{x}_{<t}$ at each time step. Details about the discriminator are in \cref{sec:imp_appendix}.
\paragraph*{\Gls{sca}.} The variational family of \gls{vdm} is defined by a number of modeling choices, including the factorization and marginal consistency assumptions of \cref{eqn:posterior}, the parametrization of the transition and inference networks \cref{eqn:qtilde,eqn:inference}, and the choice of weighting function $\omega(\cdot)$. It is also sensitive to the choice of sampling method which we discuss here. In principle, we could use Monte-Carlo methods. However, for a relatively small number of samples $k$, Monte-Carlo methods don't have a mechanism to control the quality of samples. We instead develop a semi-stochastic approach based on the cubature approximation \citep{wan2000unscented,wu2006numerical,arasaratnam2009cubature}, which chooses samples more carefully. The cubature approximation proceeds by constructing $k=2d+1$ so-called sigma points, which are optimally spread out on the $d$-dimensional Gaussian with the same mean and covariance as the distribution we need samples from. In \gls{sca}, the deterministic sigma points are infused with Gaussian noise to obtain stochastic {\em sigma variables}. A detailed derivation of \gls{sca} is in \cref{sec:cubature_appendix}. 

We use \gls{sca} for various reasons: First, it typically requires fewer samples than Monte-Carlo methods because the sigma points are carefully chosen to capture the first two moments of the underlying distribution. Second, it ensures a persistence of the mixture components; when we resample, we sample another nearby point from the mixture component and not an entirely new location.

%% file: experiments.tex
\section{Evaluation and Experiments}
\label{sec:exp}
In this empirical study, we evaluate \gls{vdm}'s ability to model multi-modal dynamics and show its competitive forecasting performance in various domains. 
We first introduce the evaluation metrics and baselines. Experiments on synthetic data demonstrate that \gls{vdm} is truly multi-modal thereby supporting the modeling choices of \cref{sec:vdm}, especially for the inference model. Then, experiments on real-world datasets with challenging multi-modal dynamics show the benefit of \gls{vdm} over state-of-the art (deep) probabilistic time-series models.

\paragraph*{Evaluation metrics.} 
In the experiments, we always create a training set, a validation set, and a test set. During validation and test, each trajectory is split into two parts; initial observations (given to the models for inference) and continuations of the trajectories (to be predicted and not accessible to the models). The inference models are used to process the initial observations and to infer latent states. These are then processed by the generative models to produce forecasts.

We use 3 criteria to evaluate these forecasts (i) multi-steps ahead prediction $p(\mathbf{x}_{t+1:t+\tau} \mid \mathbf{x}_{1:t})$, (ii) one-step-ahead prediction $p(\mathbf{x}_{t+1} \mid \mathbf{x}_{1:t})$, and (iii) empirical Wasserstein distance. As in other work \citep{lee2017desire,bhattacharyya2018accurate,bhattacharyya2019conditional}, (i) and (ii) are reported in terms of negative log-likelihood. While the predictive distribution for one-step-ahead prediction is in closed-form, the long-term forecasts have to be computed using samples. For each ground truth trajectory $\mathbf{x}$ we generate $n=1000$ forecasts $\hat{\mathbf{x}}_i$ given initial observations from the beginning of the trajectory 
\begin{equation}
\label{eqn:nll}
    NLL = -\log\left(\frac{1}{n}\sum_{i}^{n}\frac{1}{\sqrt{2\pi}}\exp\left(-\frac{(\hat{\mathbf{x}}_i-\mathbf{x})^2}{2}\right)\right)\,,
\end{equation}
This evaluates the predictive accuracy but neglects a key aspect of multi-modal forecasts -- diversity. 

We propose a new evaluation metric, which takes both diversity and accuracy of predictions into account. 
It relies on computing the Wasserstein distance between two empirical distributions $P$, $Q$
\begin{equation}
\label{eqn:wasserstein_eval}
	W(P,Q) = \inf_{\pi}\biggl(\frac{1}{n}\sum_{i}^{n}\lVert(\mathbf{x}_i-\mathbf{y}_{\pi(i)}\rVert_2\biggr)\,,
\end{equation}
where $\mathbf{x}$ and $\mathbf{y}$ are the discrete samples of $P$ and $Q$, and $\pi$ denotes all permutations \citep{villani2008optimal}. 
To use this as an evaluation measure for multi-modal forecasts, we do the following. We select $n$ samples from the test set with similar initial observations. If the dynamics in the data are multi-modal the continuations of those $n$ trajectories will be diverse and this should be reflected in the forecasts. 
For each of the $n$ samples, the model generates $10$ forecasts and we get $n$ groups of samples. With \cref{eqn:wasserstein_eval} the empirical W-distance between the $n$ true samples, and each group of generated samples can be calculated. The averaged empirical W-distance over groups evaluates how well the generated samples match the ground truth. Repeating this procedure with different initial trajectories evaluates the distance between the modeled distribution and the data distribution.

\paragraph*{Baselines.} 
We choose baselines from three classes of models. 
Two stochastic recurrent models are \glsfirst{vrnn} \citep{chung2015recurrent} and \gls{aesmc} \citep{le2018auto}. \gls{vrnn} has a similar but more powerful generative model than \gls{vdm}, and \gls{aesmc} uses \gls{smc} to achieve a tighter lower bound. But compared to \gls{vdm}, both methods have a less powerful inference model which limits their capacity to capture multi-modal distributions.
The third baseline is a deep \gls{ssm}. The \gls{rkn} \citep{becker2019recurrent} models the latent space with a locally linear \glspl{ssm}, which makes the prediction step and update step analytic (as for Kalman filters \citep{kalman1960new}). A final baseline is the \gls{cfvae} \citep{bhattacharyya2019conditional}, which uses conditional normalizing flows to model a global prior for the future continuations and achieves state-of-the-art performances.

To investigate the necessity of taking multiple samples in the \gls{vdm} inference model, we also compared to $\text{VDM}(k=1)$ which uses only a single sample in \cref{eqn:inference}. $\text{VDM}(k=1)$ has a simpler generative model than \gls{vrnn} (it considers no autoregressive feedback of the observations $\mathbf{x}$), but the same inference model. More ablations for the modeling choices of \gls{vdm} are in \cref{sec:appendix_results}. 

For fair comparison, we fix the dimension of the latent variables $\mathbf{z}_t$ and $\mathbf{h}_t$ to be the same for  \gls{vdm}, \gls{aesmc}, and \gls{vrnn} which have the same resulting model size (except for the additional autoregressive feedback in \gls{vrnn}).  \gls{aesmc} and \gls{vdm} always use the same number of particles/samples. \gls{rkn} does not have recurrent states, so we choose a higher latent dimension to make model size comparable. In contrast, \gls{cfvae} has only one global latent variable which needs more capacity and we make it higher-dimensional than $\mathbf{z}_t$. Details for each experiment are in \cref{sec:imp_appendix}.   

\begin{figure}[t!]
\vspace{-10pt}
	\centering
	\begin{subfigure}[t]{0.15\textwidth}
		\includegraphics[width=\textwidth]{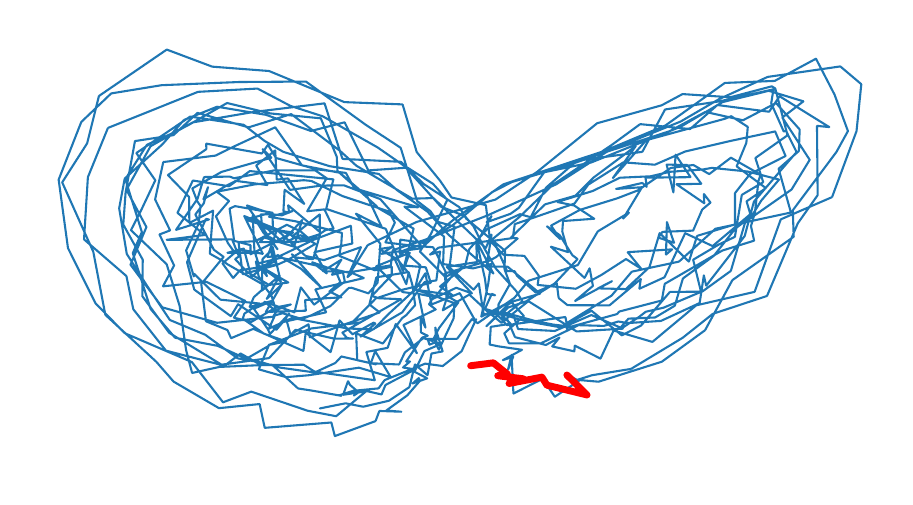}
		\caption{True sample}
	\end{subfigure}
	\begin{subfigure}[t]{0.15\textwidth}
		\includegraphics[width=\textwidth]{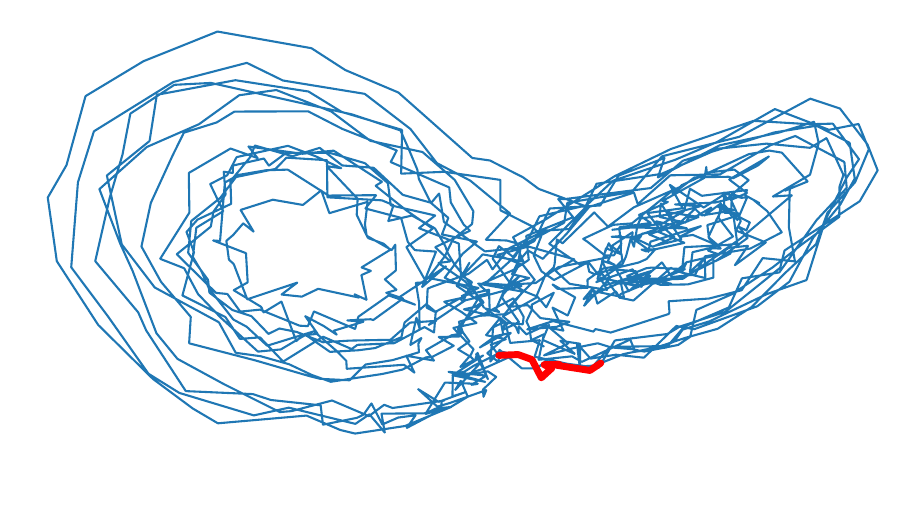}
		\caption{\acrshort{vdm} (ours)}
		\label{fig:lorenz_vdm}
	\end{subfigure}
		\begin{subfigure}[t]{0.15\textwidth}
		\includegraphics[width=\textwidth]{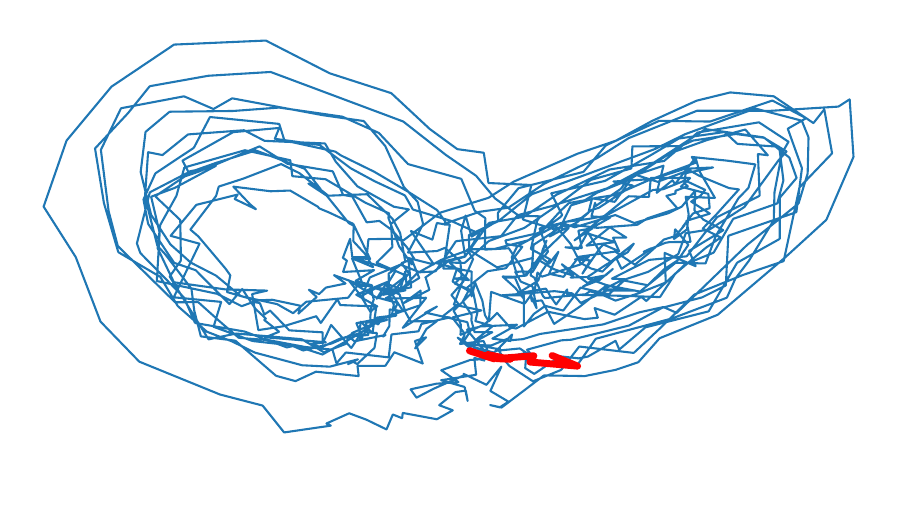}
		\caption{\acrshort{aesmc}}
		\label{fig:lorenz_smc}
	\end{subfigure}
	\begin{subfigure}[t]{0.15\textwidth}
		\includegraphics[width=\textwidth]{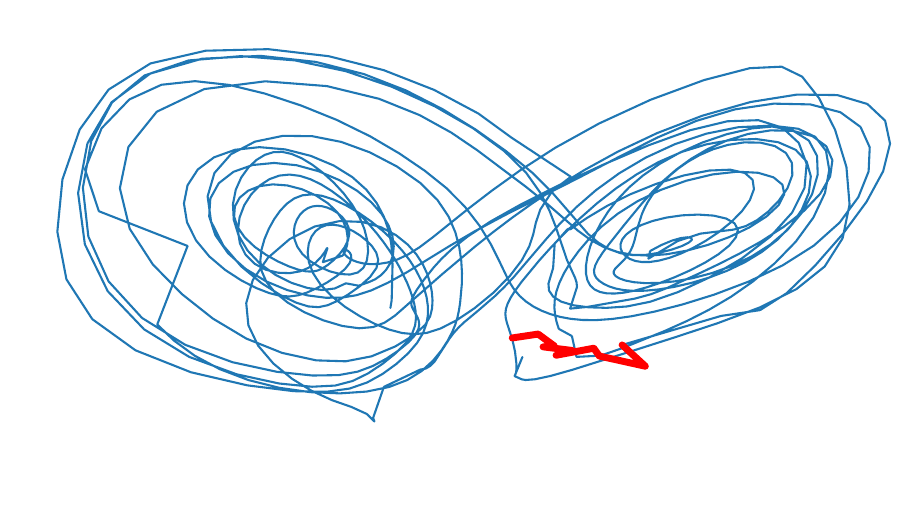}
		\caption{\acrshort{cfvae}}
		\label{fig:lorenz_vae}
	\end{subfigure}
	\begin{subfigure}[t]{0.15\textwidth}
		\includegraphics[width=\textwidth]{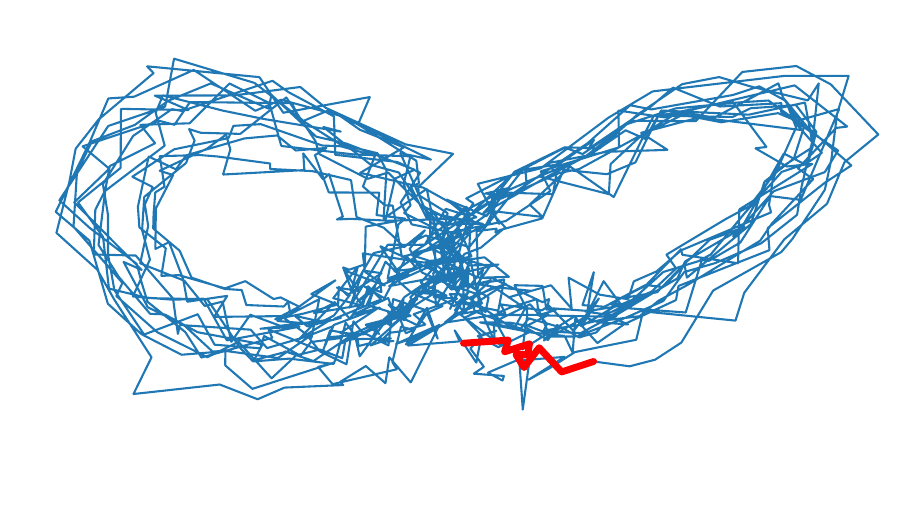}
		\caption{\acrshort{vrnn}}
		\label{fig:lorenz_vrnn}
	\end{subfigure}
		\begin{subfigure}[t]{0.15\textwidth}
		\includegraphics[width=\textwidth]{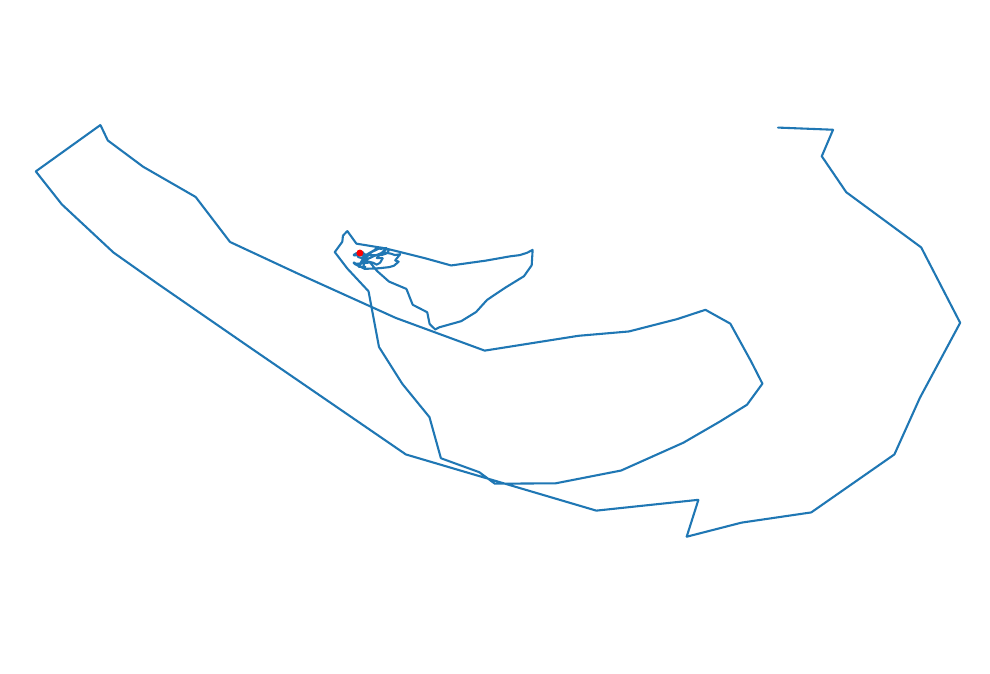}
		\caption{\acrshort{rkn} }
		\label{fig:lorenz_rkn}
	\end{subfigure}
	\caption{Generated samples from \gls{vdm} and baselines for stochastic Lorenz attractor. The models generate the remaining 990 observations (blue) based on the first 10 observations (red). Due to the chaotic property, the reconstruction is impossible even the model learns the right dynamics. \gls{vdm} and \gls{aesmc} capture the dynamics very well, while \gls{rkn} fails in capturing the stochastic dynamics.}
	\label{fig:lorenz}
\end{figure}

\paragraph*{Synthetic data with multi-modal dynamics.}
\begin{figure}[t]
	\captionsetup[subfigure]{labelformat=empty}
	\centering
	\resizebox{\linewidth}{!}{
	\begin{tabular}{c|c|c}
	$\text{VDM}(k=9)$&$\text{VDM}(k=1)$&\gls{aesmc}$
	(k=9)$\\
	\hskip -1ex
	\begin{subfigure}[b]{0.11\textwidth}
	\includegraphics[width=\textwidth]{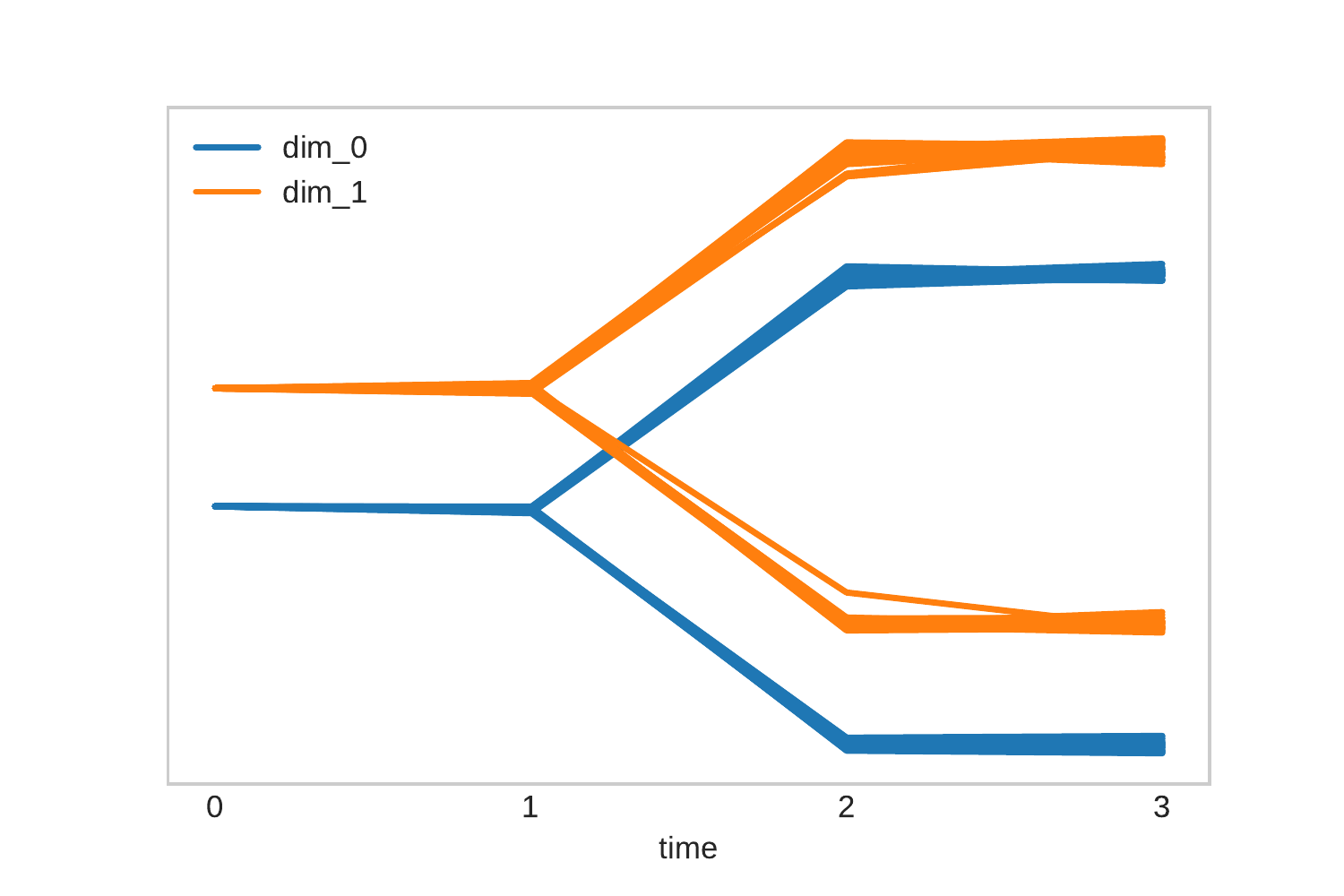}
	\caption{$\Tilde{\mathbf{D}}$}
	\end{subfigure}
	\hskip -1ex
	\begin{subfigure}[b]{0.11\textwidth}
		\includegraphics[width=\textwidth]{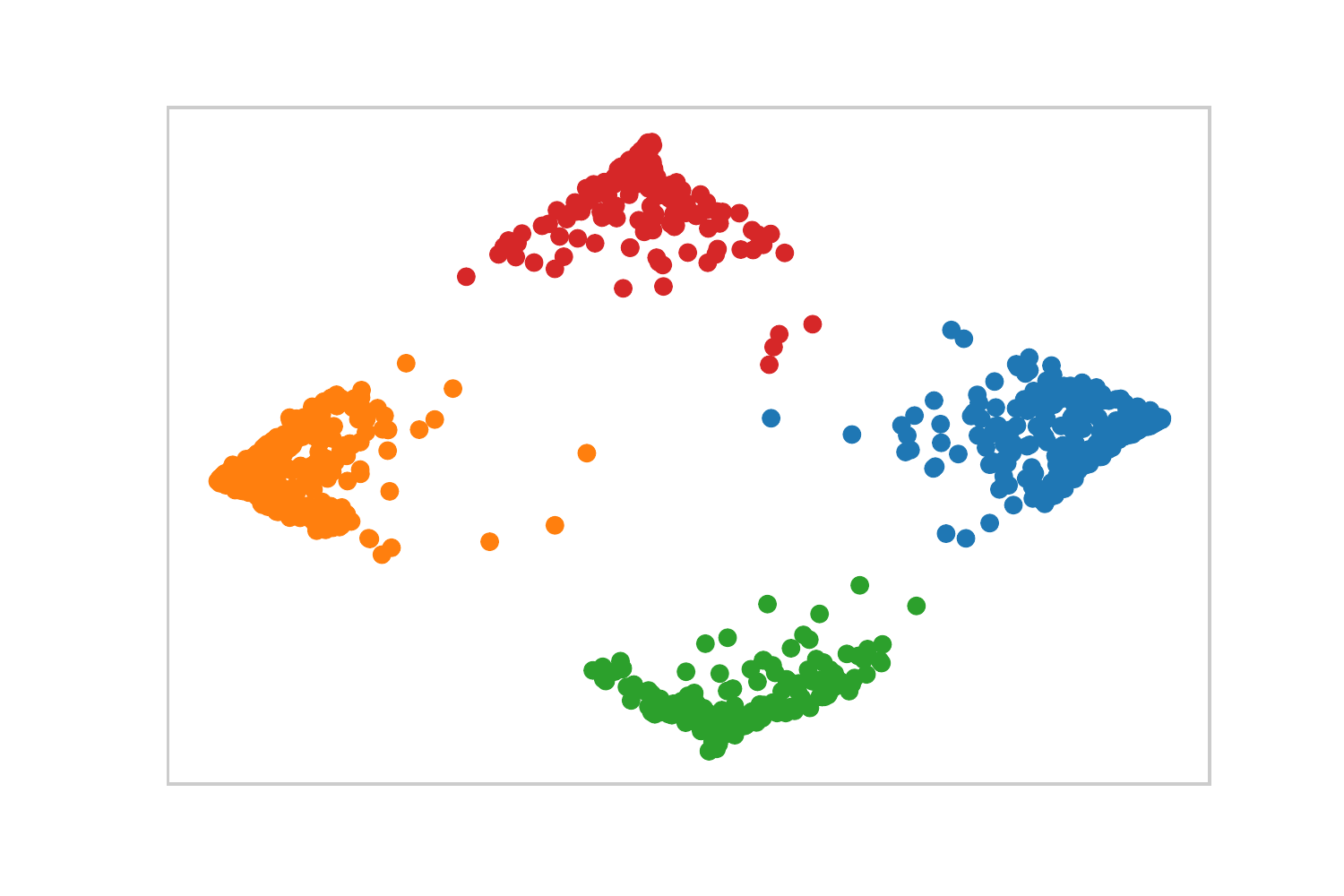}
	\caption{$p(\mathbf{z}_2|\mathbf{D})$}
	\end{subfigure}
	\hskip -1ex
	\begin{subfigure}[b]{0.11\textwidth}
		\includegraphics[width=\textwidth]{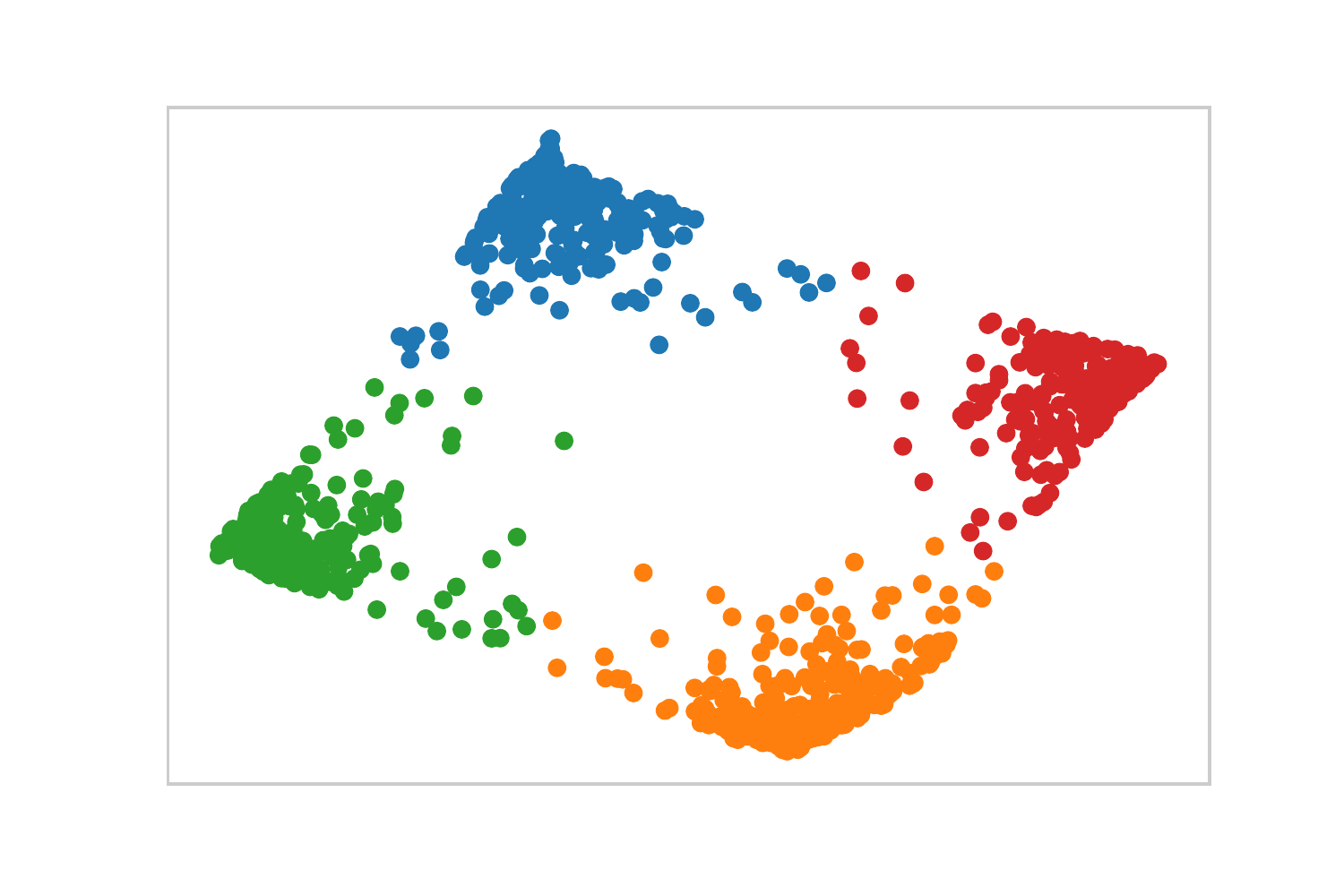}
	\caption{$p(\mathbf{z}_2|\mathbf{x}_{\leq1})$}

	\end{subfigure}
	\hskip -1ex
	 &
	 \begin{subfigure}[b]{0.11\textwidth}
	\includegraphics[width=\textwidth]{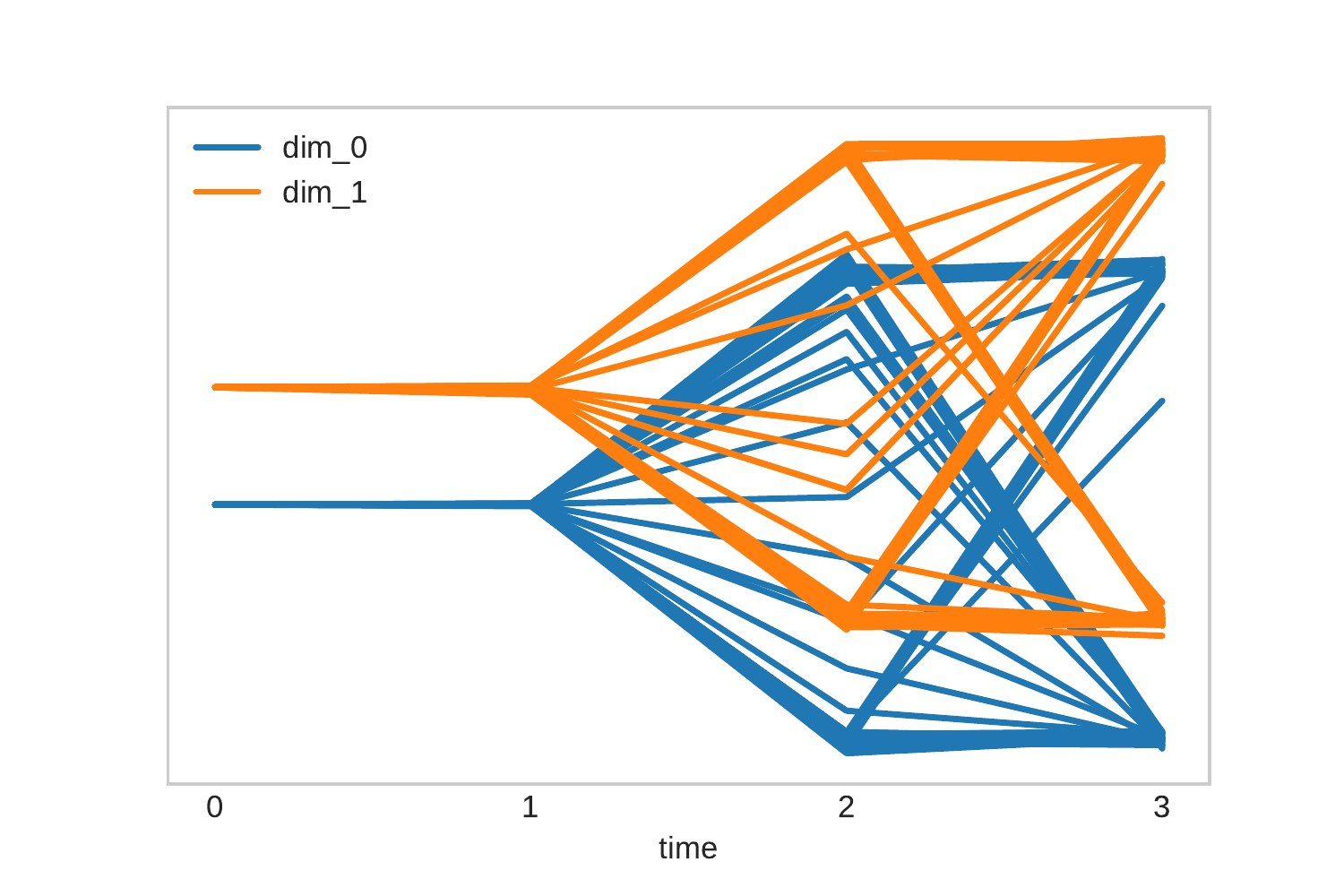}
	\caption{$\Tilde{\mathbf{D}}$}
	\end{subfigure}
	\hskip -1ex
	\begin{subfigure}[b]{0.11\textwidth}
		\includegraphics[width=\textwidth]{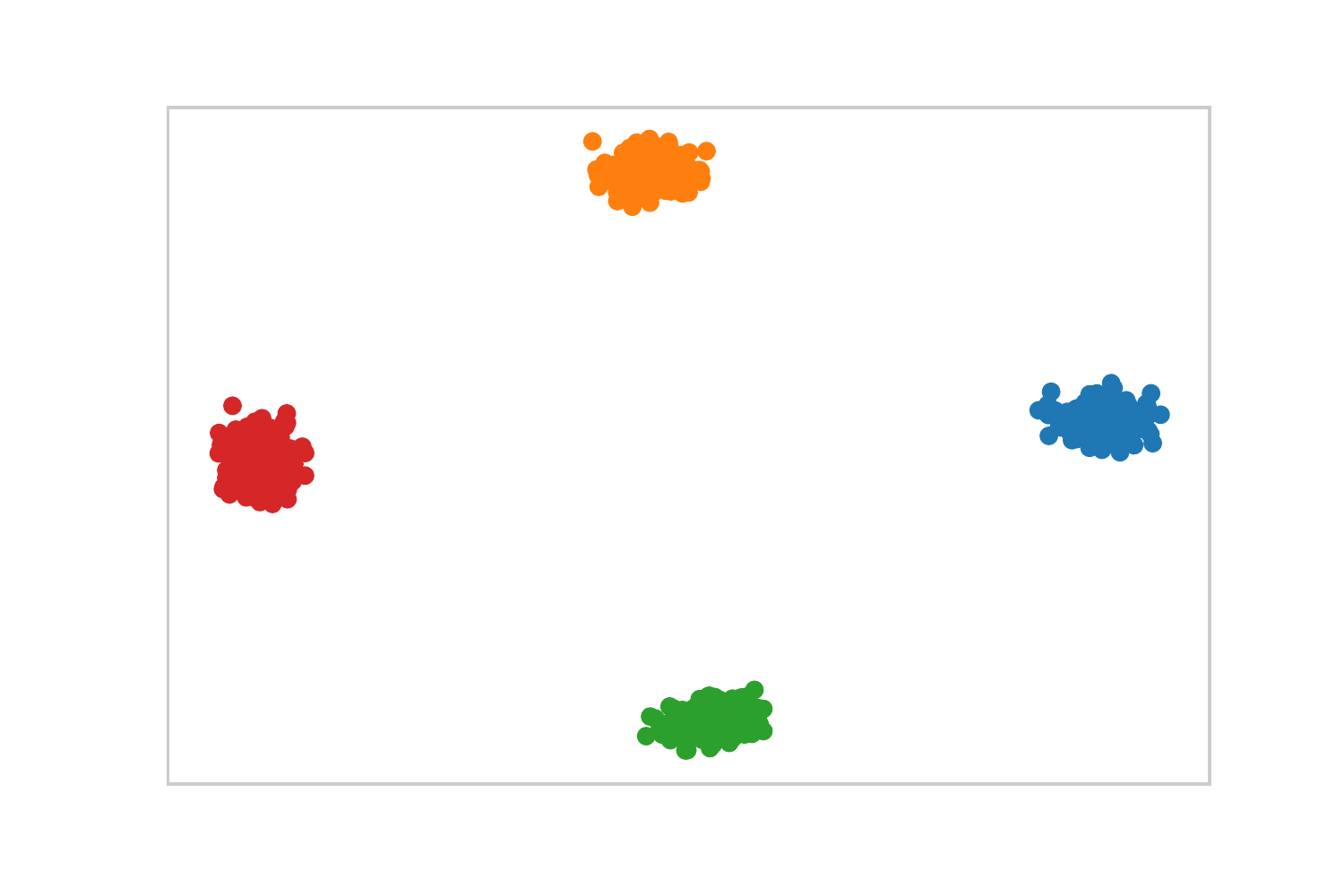}
	\caption{$p(\mathbf{z}_2|\mathbf{D})$}
	\end{subfigure}

    \hskip -1ex
	\begin{subfigure}[b]{0.11\textwidth}
		\includegraphics[width=\textwidth]{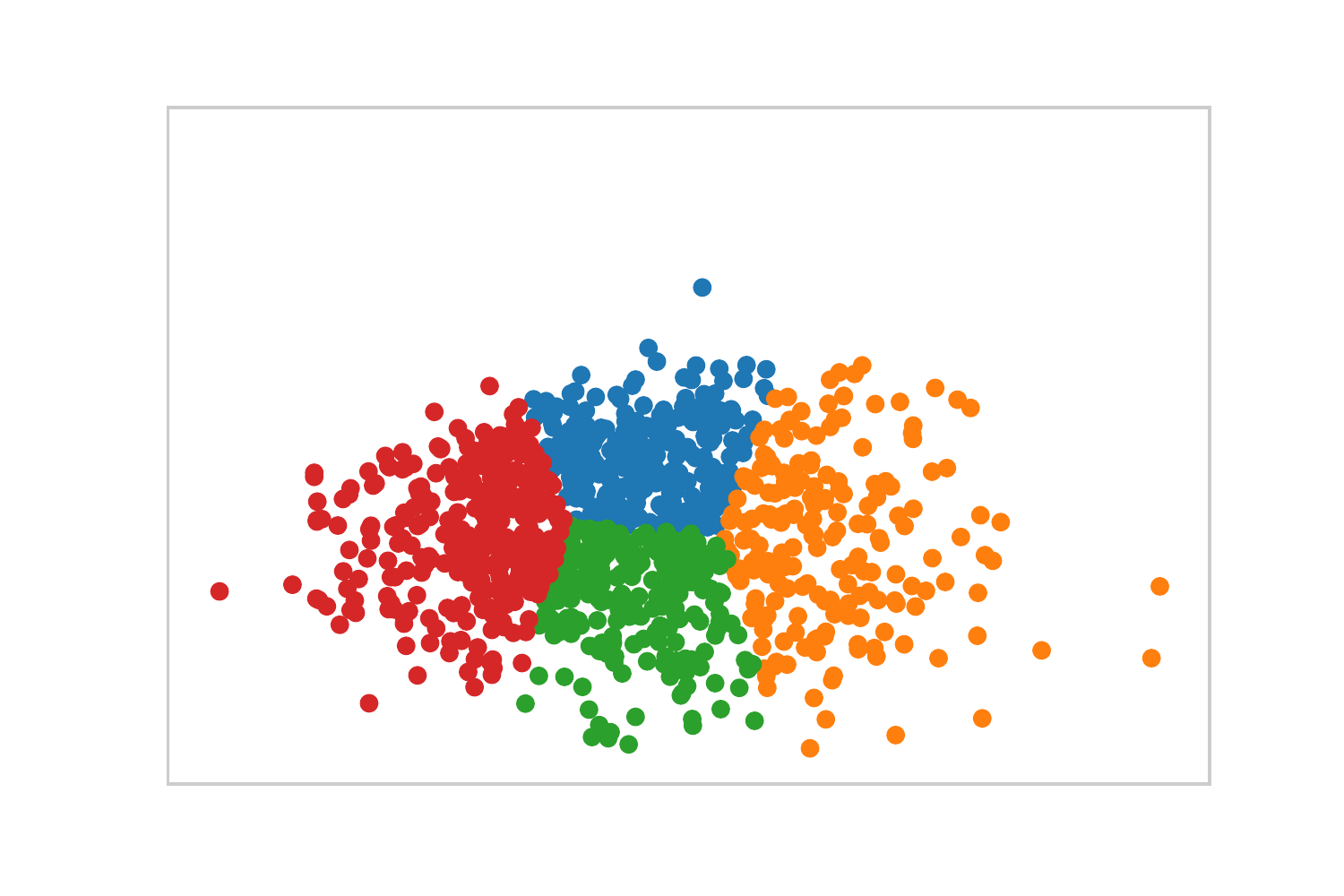}
	\caption{$p(\mathbf{z}_2|\mathbf{x}_{\leq1})$}
	\end{subfigure}
	\hskip -1ex
		 & 	
		 	 \begin{subfigure}[b]{0.11\textwidth}
	\includegraphics[width=\textwidth]{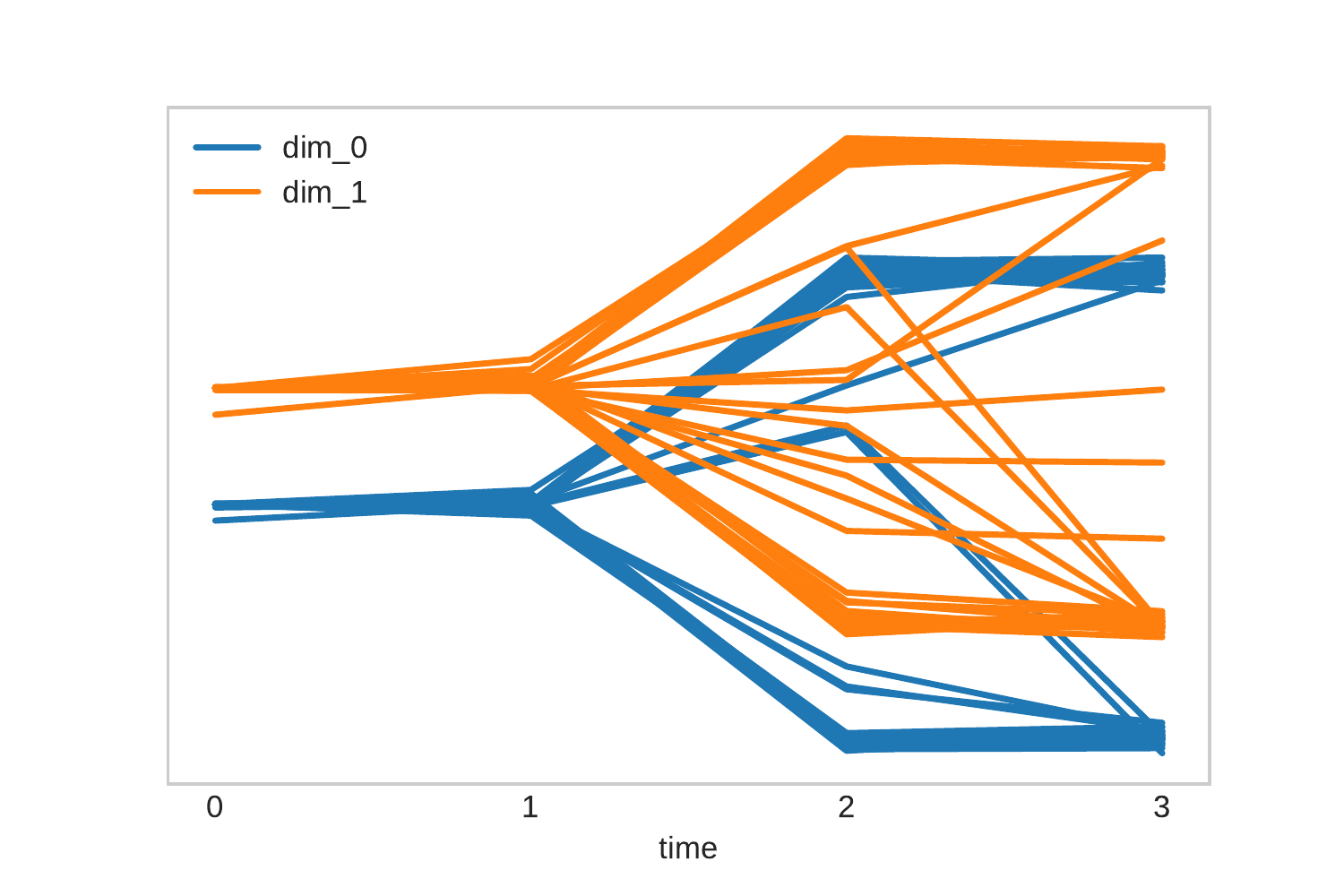}
	\caption{$\Tilde{\mathbf{D}}$}
	\end{subfigure}
	\hskip -1ex
	\begin{subfigure}[b]{0.11\textwidth}
		\includegraphics[width=\textwidth]{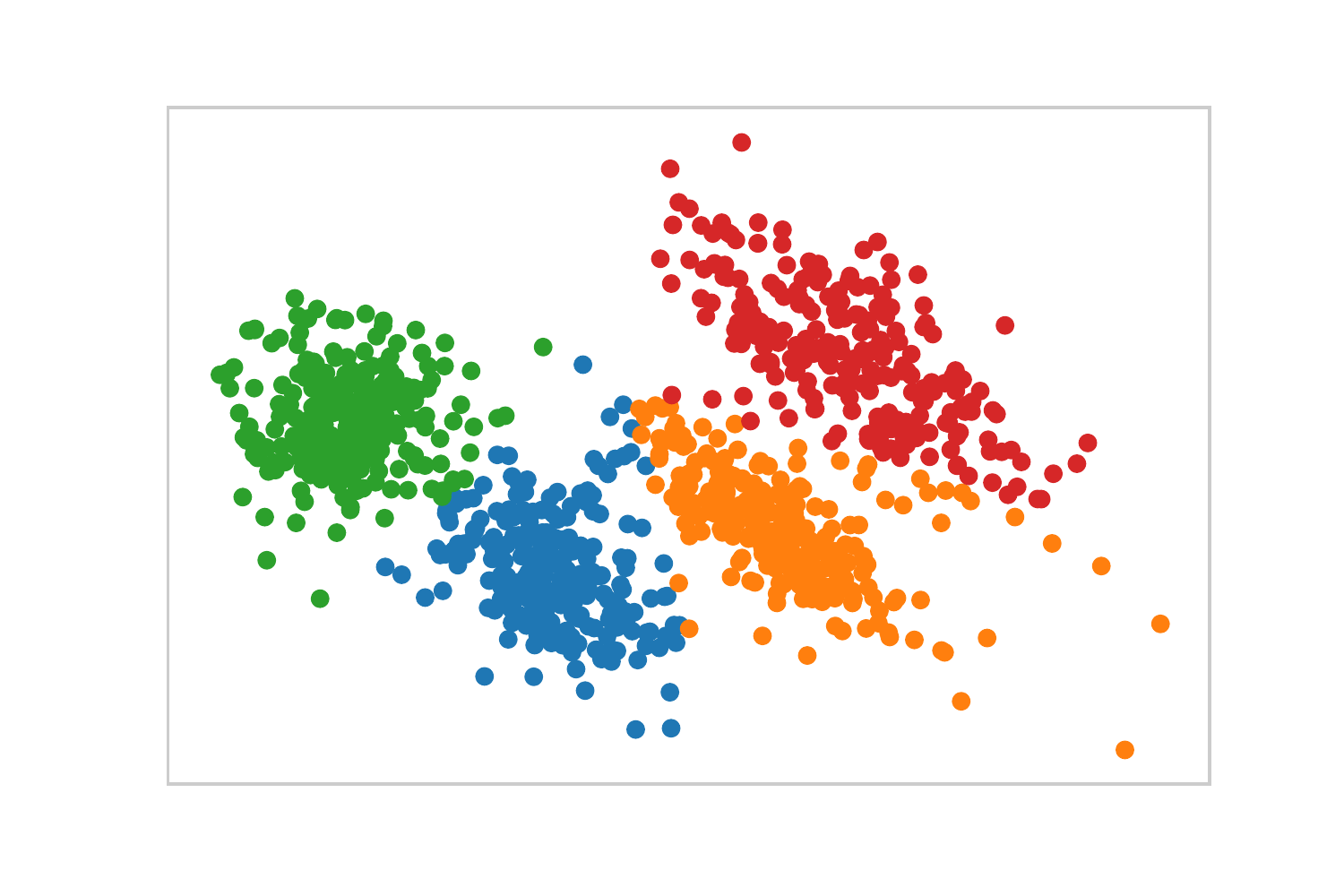}
	\caption{$p(\mathbf{z}_2|\mathbf{D})$}
	\end{subfigure}

    \hskip -1ex
	\begin{subfigure}[b]{0.11\textwidth}
		\includegraphics[width=\textwidth]{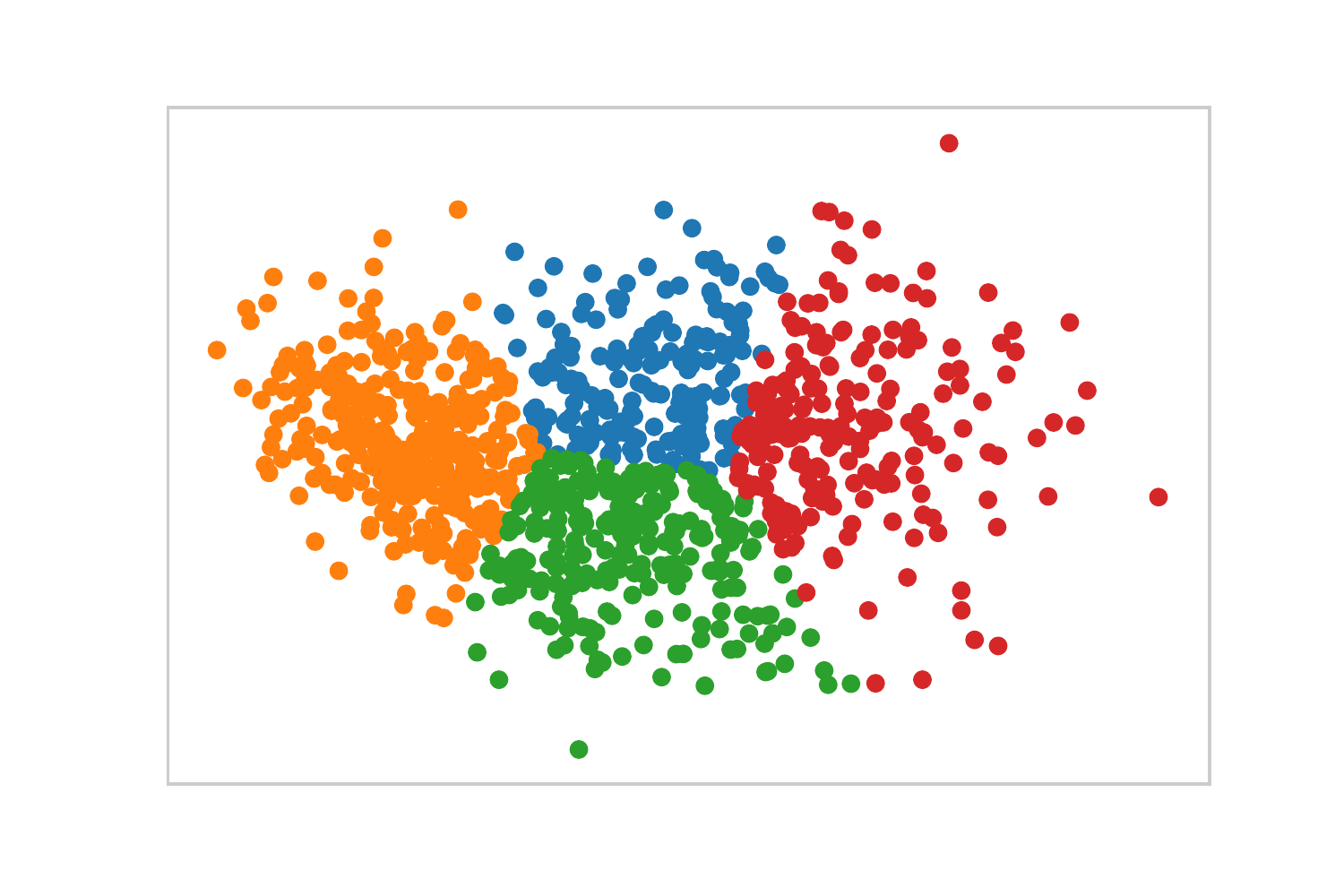}
	\caption{$p(\mathbf{z}_2|\mathbf{x}_{\leq1})$}

	\end{subfigure}
	\hskip -1ex
	\end{tabular}}
	\caption{
	{\bf Experiments on 2d synthetic data with 4 modes highlight the multi-modality of \gls{vdm}.} We train $\text{VDM}(k=9)$ (left), $\text{VDM}(k=1)$ (middle), and \gls{aesmc}$(k=9)$ (right) on a training set of trajectories $\mathbf{D}$ of length 4, and plot generated trajectories $\Tilde{\mathbf{D}}$ (2 colors for 2 dimensions). We also plot the aggregated posterior $p(\mathbf{z}_2|\mathbf{D})$, and the predictive prior $p(\mathbf{z}_2|\mathbf{x}_{\leq1})$ (4 colors for 4 clusters, and not related to the colors in the trajectories plot) at the second time step. Only \gls{vdm} learns a multi-modal predictive prior, which explains its success in modeling multi-modal dynamics.}
		\label{fig:toy}
		\vspace{-5pt}
\end{figure}
We generate synthetic data with two dimensions and four modes and compare the performance of \gls{vdm} with $9$ samples (\cref{fig:toy}, left), \gls{vdm} with a single sample (\cref{fig:toy}, middle), and \gls{aesmc} using $9$ particles (\cref{fig:toy}, right). Since variational inference is known to try to match the aggregated posterior with the predictive prior \citep{tomczak2017vae}, it is instructive to fit all three models and to look at their predictive prior $p(\mathbf{z}_2|\mathbf{x}_{\leq1})$ and the aggregated posterior $p(\mathbf{z}_2|\mathbf{D})$. Because of the multi-modal nature of the problem, all 3 aggregated posteriors are multi-modal, but only \gls{vdm}($k=9$) learns a multi-modal predictive prior (thanks to its choice of variational family).
Although \gls{aesmc} achieves a good match between the prior and the aggregated posterior, the predictive prior does not clearly separate into different modes. In contrast, the inference model of \gls{vdm} successfully uses the weights (\cref{eqn:pi}), which contain information about the incoming observation, to separate the latent states into separate modes.
\begin{table}[t!]
	\caption{Prediction error on stochastic Lorenz attractor for three evaluation metrics (details in main text). $\text{VDM}(k=13)$ achieves the best performance, and \gls{aesmc} also gives comparable results.}
	\label{tab:lorenz_eva}
	\centering
	\vspace{-5pt}
	\resizebox{\linewidth}{!}{
	\begin{tabular}{l|cccc|cc}

		     & \acrshort{rkn}   &\acrshort{vrnn}   &\acrshort{cfvae} &\acrshort{aesmc}   &$\text{VDM}(k=1)$&$\text{VDM}(k=13)$\\
		\hline
        Multi-steps & 104.41 & 65.89$\pm$0.21 & 32.41$\pm$0.13 &25.01$\pm$0.22&25.03$\pm$0.28& \textbf{24.46}$\pm$0.12 \\
        One-step & 1.88 & -1.63 & n.a &-1.69&-1.81& \textbf{-1.81} \\
        W-distance &16.16 &16.14$\pm$0.006  &8.44$\pm$0.005&7.29$\pm$0.005 &7.31$\pm$0.002&\textbf{7.28}$\pm$0.002 \\

	\end{tabular}}
		\vspace{-12pt}
\end{table}
\paragraph*{Stochastic Lorenz attractor.}
The Lorenz attractor is a system governed by ordinary differential equations. We add noise to the transition and emission function to make it stochastic (details in \cref{sec:lorenz_exp_appendix}). Under certain parameter settings it is chaotic -- even small errors can cause considerable differences in the future. This makes forecasting its dynamics very challenging.
All models are trained and then tasked to predict $90$ future observations given $10$ initial observations.
\cref{fig:lorenz} illustrates qualitatively that \gls{vdm} (\cref{fig:lorenz_vdm}) and \gls{aesmc} (\cref{fig:lorenz_smc}) succeed in modeling the chaotic dynamics of the stochastic Lorenz attractor, while \gls{cfvae} (\cref{fig:lorenz_vae}) and \gls{vrnn} (\cref{fig:lorenz_vrnn}) miss local details, and \gls{rkn} (\cref{fig:lorenz_rkn}) which lacks the capacity for stochastic transitions does not work at all. \gls{vdm} achieves the best scores on all metrics (\cref{tab:lorenz_eva}). Since the dynamics of the Lorenz attractor are governed by ordinary differential equations, the transition dynamics at each time step are not obviously multi-modal, which explains why all models with stochastic transitions do reasonably well. Next, we will show the advantages of \gls{vdm} on real-world data with multi-modal dynamics.
\begin{figure}[t!]
	\captionsetup[subfigure]{labelformat=empty}
	\centering
	\resizebox{\linewidth}{!}{
	\begin{tabular}{c|c|c|c}
	Trajectory&$\text{VDM}(k=13)$&$\text{VDM}(k=1)$&\gls{aesmc}$
	(k=13)$\\
		\hskip -1ex
	\begin{subfigure}[b]{0.09\textwidth}
	\includegraphics[width=\textwidth]{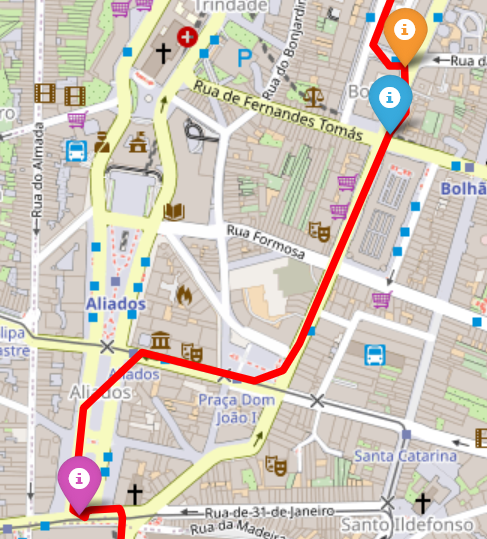}
	\end{subfigure}
	\hskip -1ex
	&
	\hskip -1ex
	\begin{subfigure}[b]{0.1\textwidth}
	\includegraphics[width=\textwidth]{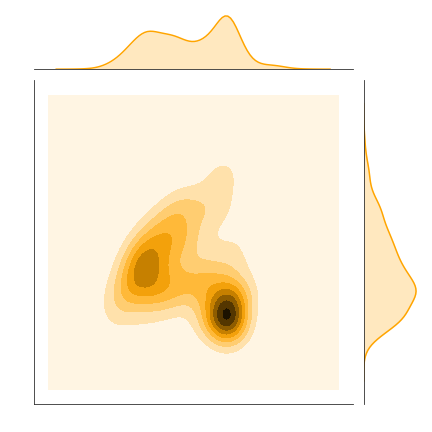}
	\end{subfigure}
	\hskip -1ex
	\begin{subfigure}[b]{0.1\textwidth}
		\includegraphics[width=\textwidth]{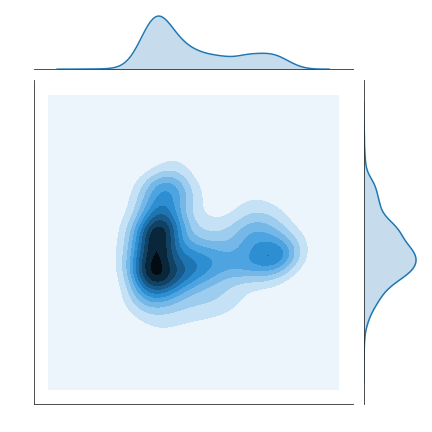}
	\end{subfigure}
	\hskip -1ex
	\begin{subfigure}[b]{0.1\textwidth}
		\includegraphics[width=\textwidth]{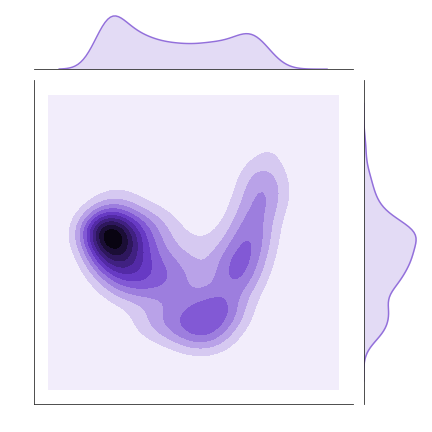}
	\end{subfigure}
	\hskip -1ex
	 &
	 \begin{subfigure}[b]{0.1\textwidth}
	\includegraphics[width=\textwidth]{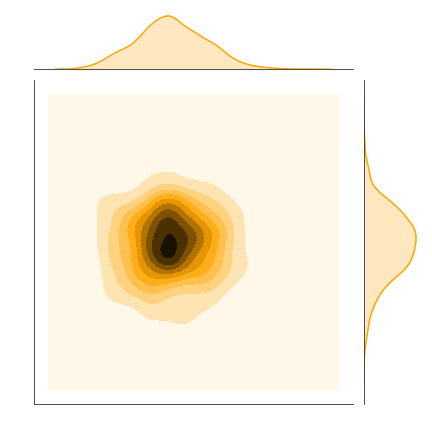}
	\end{subfigure}
	\hskip -1ex
	\begin{subfigure}[b]{0.1\textwidth}
		\includegraphics[width=\textwidth]{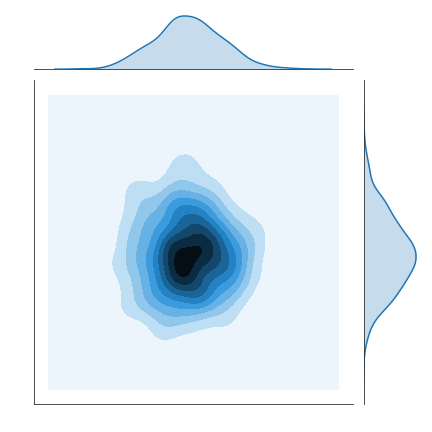}
	\end{subfigure}
    \hskip -1ex
	\begin{subfigure}[b]{0.1\textwidth}
		\includegraphics[width=\textwidth]{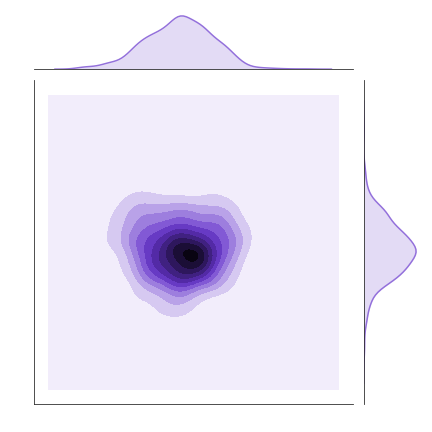}
	\end{subfigure}
	\hskip -1ex
		 & 			 	
		\begin{subfigure}[b]{0.1\textwidth}
	\includegraphics[width=\textwidth]{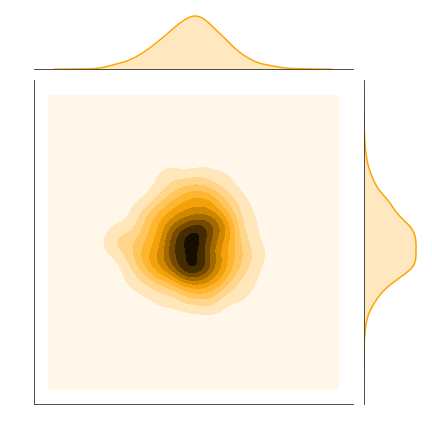}
	\end{subfigure}
	\hskip -1ex
	\begin{subfigure}[b]{0.1\textwidth}
		\includegraphics[width=\textwidth]{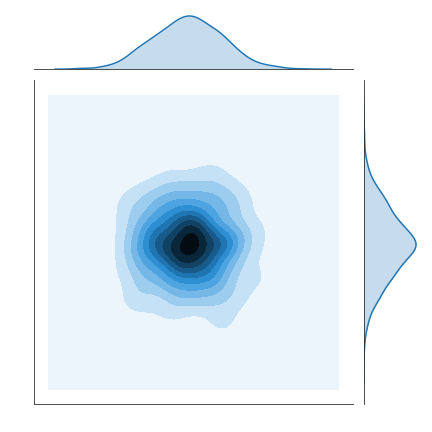}
	\end{subfigure}
    \hskip -1ex
	\begin{subfigure}[b]{0.1\textwidth}
		\includegraphics[width=\textwidth]{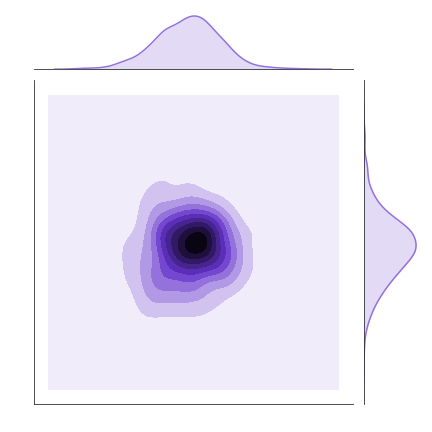}
	\end{subfigure}
	\hskip -1ex
	\end{tabular}}
	\caption{An illustration of predictive priors $p(\mathbf{z}_t | \mathbf{x}_{<t})$ of taxi trajectories from $\text{VDM}(k=13)$, $\text{VDM}(k=1)$, and \gls{aesmc}$(k=13)$ at 3 forks in the road marked on the map (left). $\text{VDM}(k=13)$ succeeds in capturing the multi-modal distributions, while the other methods approximate them with uni-modal distributions. For visualization, the distributions have been projected to 2d with KDE.}
		\label{fig:kde_taxi}
		\vspace{-5pt}
\end{figure}

\begin{table}[t!]
	\caption{Prediction error on taxi trajectories for three evaluation metrics (details in main text). \gls{cfvae} gives the best result in multi-steps prediction, since it uses one global latent variable, while sequential models rely on a sequence of local latent variables. Meanwhile, $\text{VDM}(k=13)$ outperforms all sequential models, and performs better in other metrics than \gls{cfvae}.}
	\label{tab:taxi_eva}
	\vspace{-5pt}
	\centering
	\resizebox{\linewidth}{!}{
	\begin{tabular}{l|cccc|cc}

		     & \acrshort{rkn}    &\acrshort{vrnn}  &\acrshort{cfvae} &\acrshort{aesmc} &$\text{VDM}(k=1)$&$\text{VDM}(k=13)$\\
		\hline
        Multi-steps & 4.25  & 5.51$\pm$0.002 & \textbf{2.77}$\pm$0.001 &3.31$\pm$0.001&3.26$\pm$0.001& 2.85$\pm$0.002 \\
        One-step &  -2.90 & -2.77 & n.a &-2.87&-2.99 & \textbf{-3.62}\\
        W-distance &2.07 &2.43$\pm$0.0002 &0.76$\pm$0.0003 &0.66$\pm$0.0004&0.69$\pm$0.0005 &\textbf{0.56}$\pm$0.0005 \\

	\end{tabular}}
	\vspace{-10pt}
\end{table}
\paragraph*{Taxi trajectories.}
The taxi trajectory dataset involves taxi trajectories with variable lengths in Porto, Portugal. Each trajectory is a sequence of two dimensional locations over time. Here, we cut the trajectories to a fixed length of 30 to simplify the comparison (details in \cref{sec:taxi_exp_appendix}). The task is to predict the next $20$ observations given $10$ initial observations. 
Ideally, the forecasts should follow the street map (though the map is not accessible to the models). 

The results in \cref{tab:taxi_eva} show that \gls{vdm} outperforms the other {\em sequential} latent variable models in all evaluations. However, it turns out that for multi-step forecasting learning global structure is advantageous, and \gls{cfvae} which is a global latent variable model, achieves the highest results. However, this value doesn't match the qualitative results in \cref{fig:taxi_traj}. Since \gls{cfvae} has to encode the entire structure of the trajectory forcast into a single latent variable, its predictions seem to average over plausible continuations but are locally neither plausible nor accurate. In comparison, \gls{vdm} and the other models involve a sequence of latent variables. As the forecasting progresses, the methods update their distribution over latest states, and the impact of the initial observations becomes weaker and weaker. As a result, local structure is captured more accurately. While the forecasts are plausible and can be highly diverse, they potentially evolve into other directions than the ground truth. For this reason, their multi-step prediction results are worse in terms of log-likelihood. That's why the empirical W-distance is useful to complement the evaluation of multi-modal tasks. It reflects that the forecasts of \gls{vdm} are diverse and plausible. Additionally, we illustrate the predictive prior $p(\mathbf{z}_t|\mathbf{x}_{<t})$ at different time steps in \cref{fig:kde_taxi}. \gls{vdm}($k=13$) learns a multi-modal predictive prior, which \gls{vdm}($k=1$) and \gls{aesmc} approximate it with an uni-modal Gaussian.

\paragraph*{U.S. pollution data.}
In this experiment, we study \gls{vdm} on the U.S. pollution dataset (details in \cref{sec:pollution_exp_appendix}). 
The data is collected from counties in different states from 2000 to 2016. Each observation has 12 dimensions (mean, max value, and air quality index of NO2, O3, SO2, and O3). The goal is to predict monthly pollution values for the coming 18 months, given observations of the previous six months. We ignore the geographical location and time information to treat the development tendency of pollution in different counties and different times as i.i.d.. The unknown context information makes the dynamics multi-modal and challenging to predict accurately. Due to the small size and high dimensionality of the dataset, there are not enough samples with very similar initial observations. Thus, we cannot evaluate empirical W-distance in this experiment. In multi-step predictions and one-step predictions, $\text{VDM}$ outperforms the other methods. 

\begin{figure}[t!]
\vspace{-5pt}
	\centering
	\begin{subfigure}[t]{0.16\textwidth}
		\includegraphics[width=\textwidth]{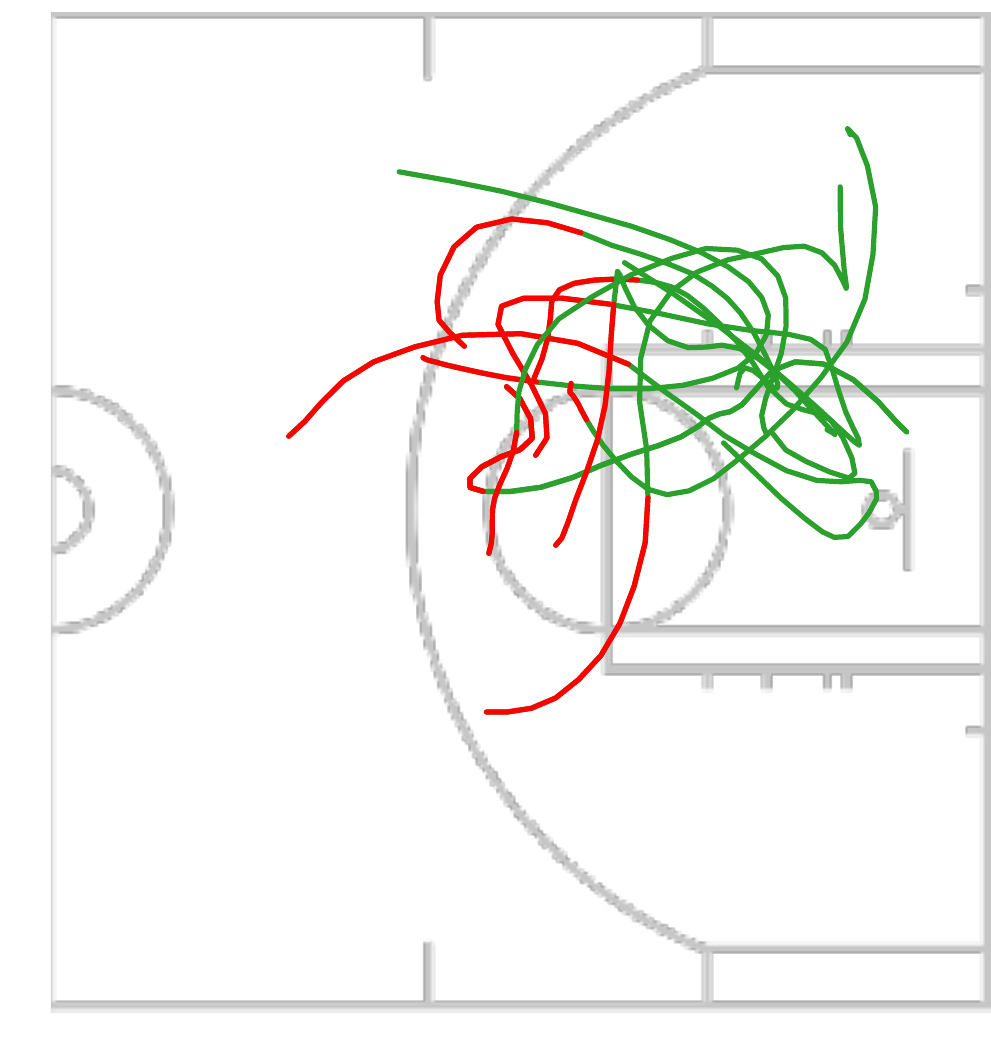}
		\caption{NBA data}
		\label{fig:player_gt}
	\end{subfigure}
	\begin{subfigure}[t]{0.16\textwidth}
		\includegraphics[width=\textwidth]{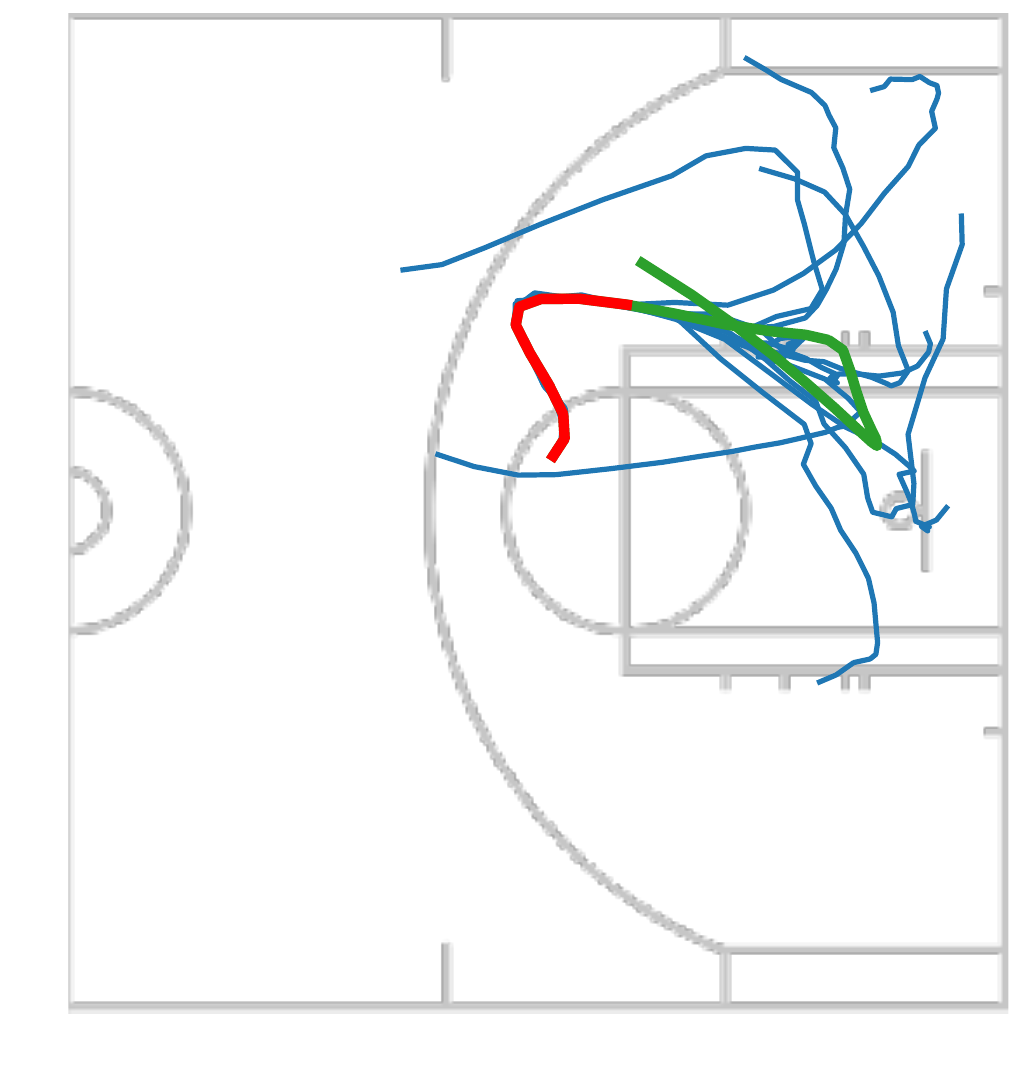}
		\caption{\acrshort{vdm} (ours)}
		\label{fig:player_vdm}
	\end{subfigure}
		\begin{subfigure}[t]{0.16\textwidth}
		\includegraphics[width=\textwidth]{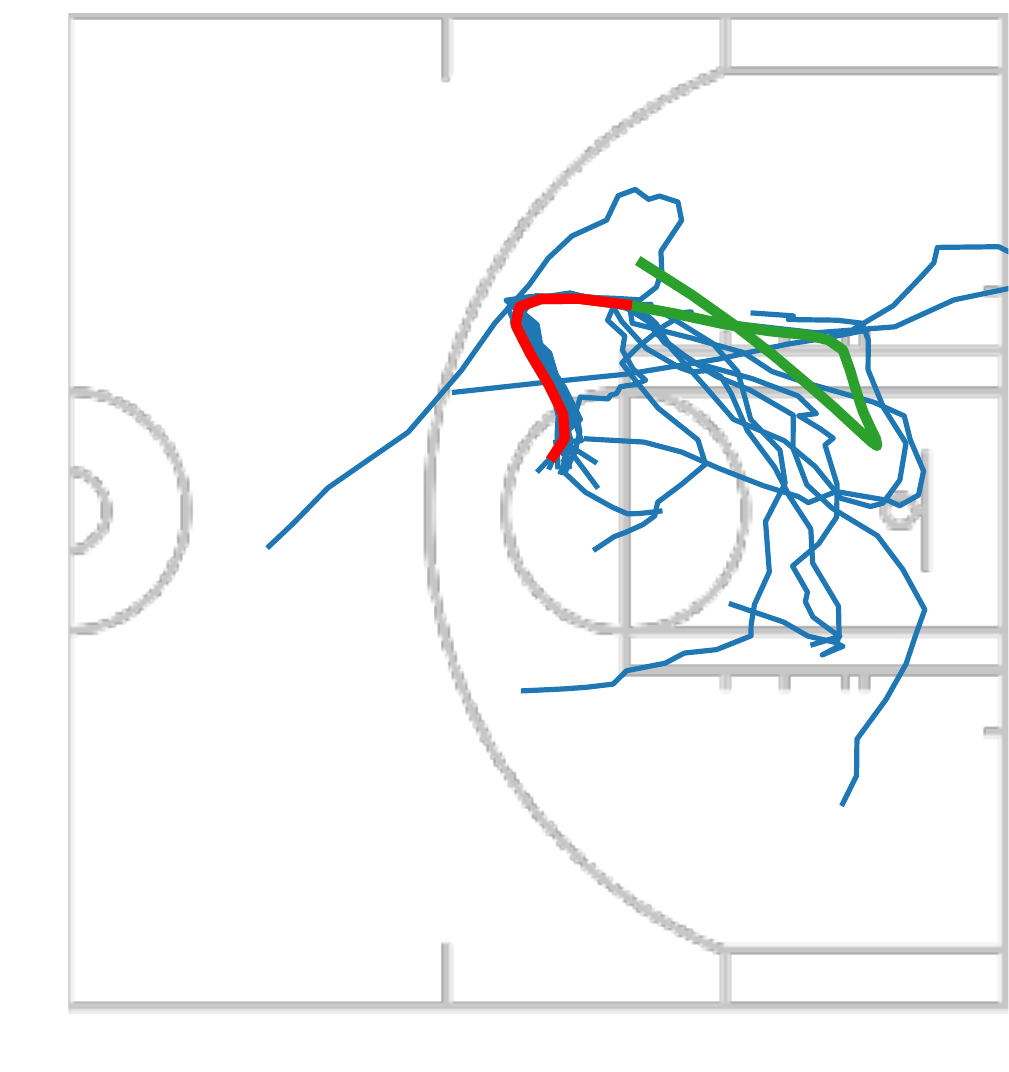}
		\caption{\acrshort{aesmc}}
		\label{fig:player_smc}
	\end{subfigure}
	\begin{subfigure}[t]{0.16\textwidth}
		\includegraphics[width=\textwidth]{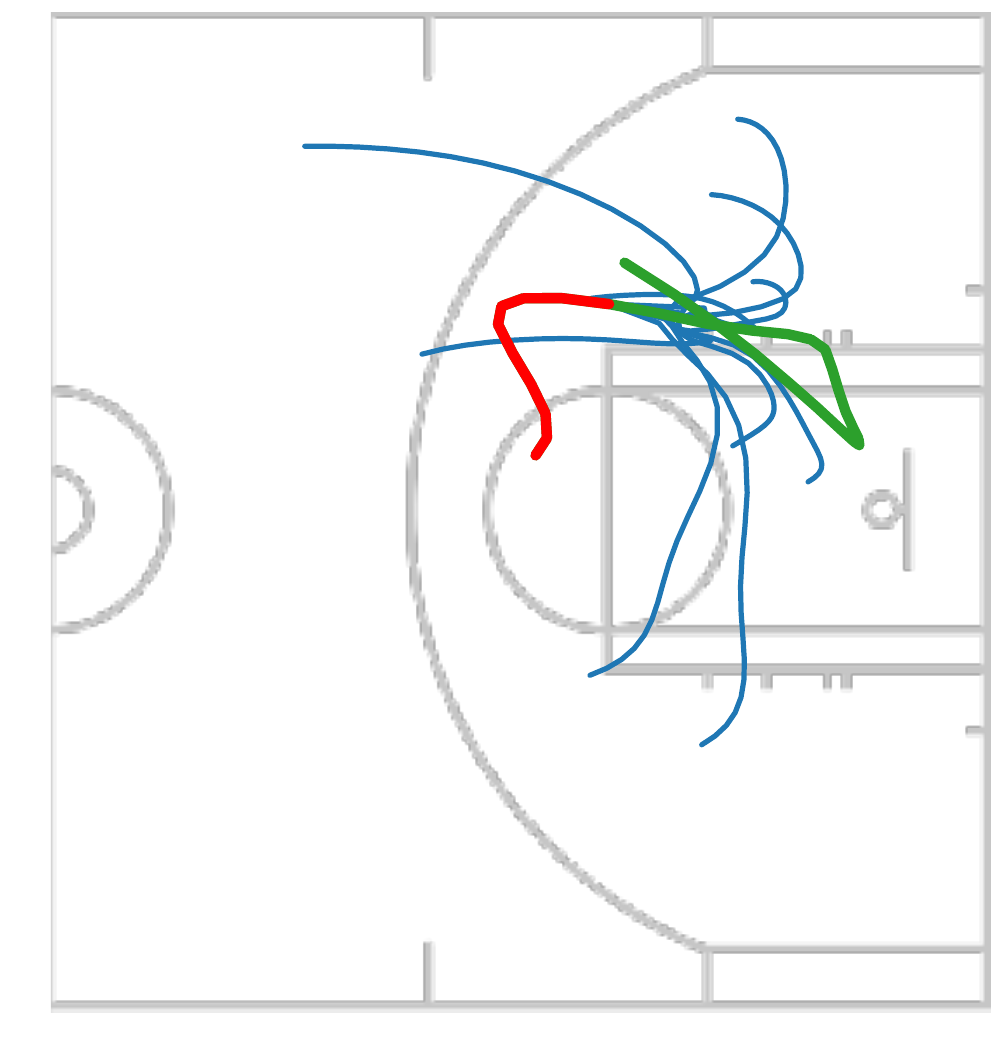}
		\caption{\acrshort{cfvae}}
		\label{fig:player_vae}
	\end{subfigure}
	\begin{subfigure}[t]{0.16\textwidth}
		\includegraphics[width=\textwidth]{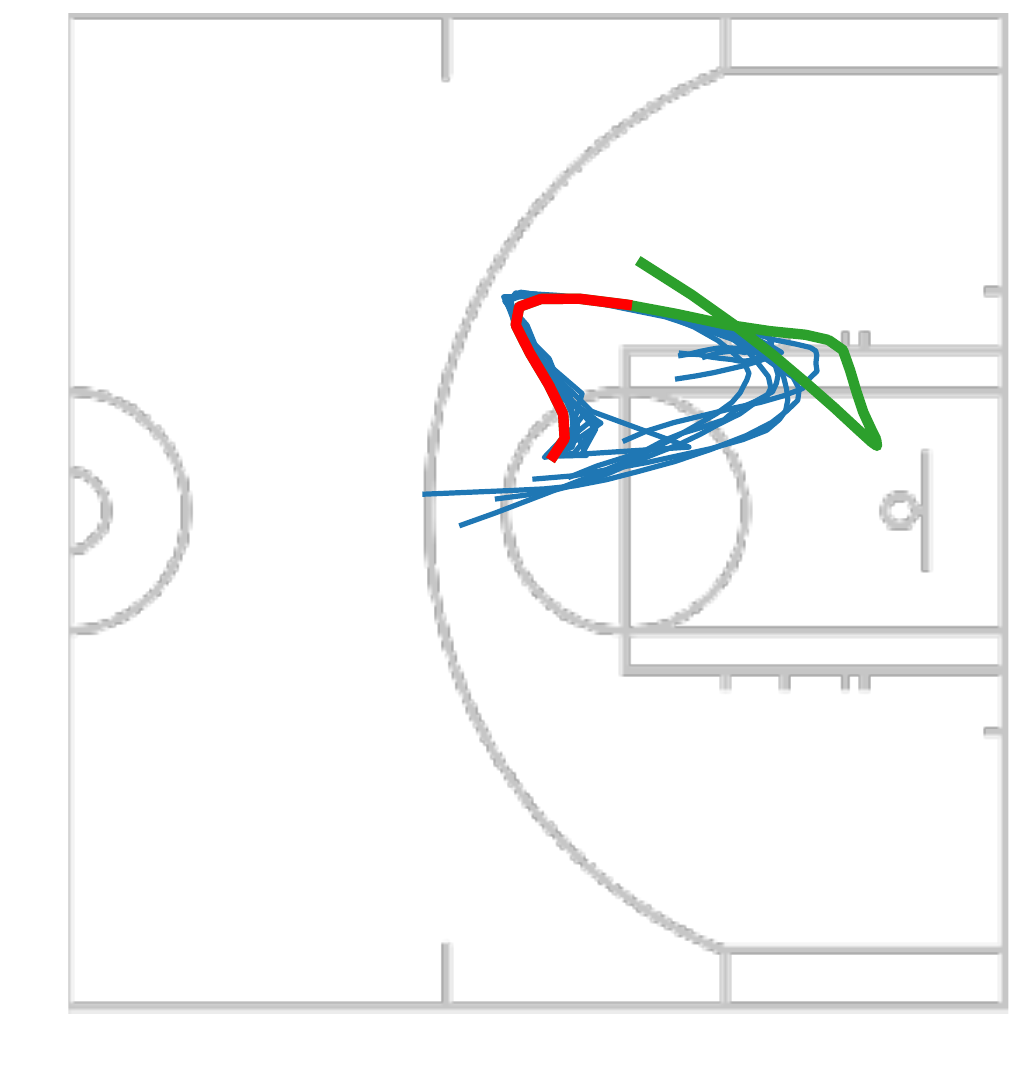}
		\caption{\acrshort{vrnn}}
		\label{fig:player_vrnn}
	\end{subfigure}
		\begin{subfigure}[t]{0.16\textwidth}
		\includegraphics[width=\textwidth]{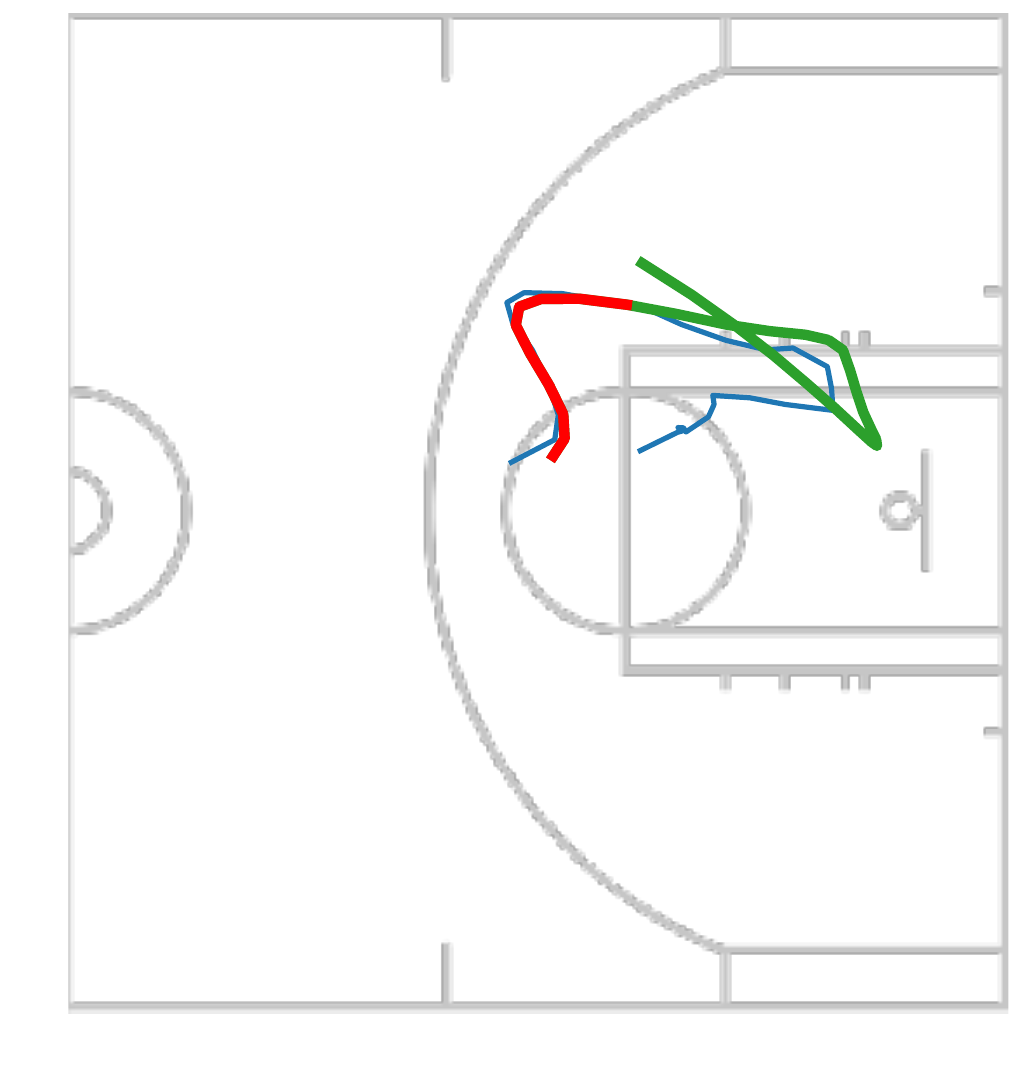}
		\caption{\acrshort{rkn} }
		\label{fig:player_rkn}
	\end{subfigure}
	\caption{\gls{vdm} and \gls{cfvae} generate plausible multi-modal trajectories of basketball plays. Each model's forecasts (blue) are based on the first 10 observations (red). Ground truth data is green.}
	\label{fig:players}
\end{figure}

\paragraph*{NBA SportVu data.}
This dataset\footnote{A version of the dataset is available at https://www.stats.com/data-science/} of sequences of 2D coordinates describes the movements of basketball players and the ball. We extract the trajectories and cut them to a fixed length of 30 to simplify the comparisons (details in \cref{sec:player_exp_appendix}). The task is to predict the next 20 observations given 10 initial observations. Players can move anywhere on the court and hence their movement is less structured than the taxi trajectories which are constrained by the underlying street map. Due to this, the initial movement patters are not similar enough to each other to evaluate empirical W-distance. In multi-step and one-step predictions, \gls{vdm} outperforms the other baselines (\cref{tab:basketball_eva}). \cref{fig:players} illustrates qualitatively that \gls{vdm} (\cref{fig:player_vdm}) and \gls{cfvae} (\cref{fig:player_vae}) succeed in capturing the multi-modal dynamics. The forecasts of \gls{aesmc} (\cref{fig:player_smc}) are less plausible (not as smooth as data), and \gls{vrnn} (\cref{fig:player_vrnn}) and \gls{rkn} (\cref{fig:player_rkn}) fail in capturing the multi-modality.

\begin{table}[t!]
	\caption{Prediction error on U.S. pollution data for two evaluation metrics (details in main text). \gls{vdm} makes the most accurate multi-step and one-step predictions.}
	\label{tab:pollution_eva}
	\centering
	\vspace{-5pt}
	\resizebox{\linewidth}{!}{
	\begin{tabular}{l|cccc|cc}

		     & \acrshort{rkn}    &\acrshort{vrnn}  &\acrshort{cfvae}  &\acrshort{aesmc} &$\text{VDM}(k=1)$&$\text{VDM}(k=17)$\\
		\hline
        Multi-steps & 53.13  & 49.32$\pm$0.13  & 45.86$\pm$0.04&41.14$\pm$0.13&42.33$\pm$0.11& \textbf{36.72}$\pm$0.08\\
        One-step &  6.98& 8.69 & n.a &6.93&7.97& \textbf{6.05}\\

	\end{tabular}}
\end{table}

\begin{table}[t!]
	\caption{Prediction error on basketball players' trajectories (details in main text). \gls{vdm} makes the most accurate multi-step and one-step predictions.}
	\label{tab:basketball_eva}
	\centering
	\vspace{-5pt}
	\resizebox{\linewidth}{!}{
	\begin{tabular}{l|cccc|cc}

		     & \acrshort{rkn}    &\acrshort{vrnn}  &\acrshort{cfvae}  &\acrshort{aesmc} &$\text{VDM}(k=1)$&$\text{VDM}(k=13)$\\
		\hline
        Multi-steps & 4.88 & 5.42$\pm$0.009  & 3.24$\pm$0.003 &3.74$\pm$0.003&3.56$\pm$0.005& \textbf{3.18}$\pm$0.005\\
        One-step &  1.55 & -2.78 & n.a &-3.91&-4.26& \textbf{-5.18}\\

	\end{tabular}}
\end{table}

%% file: conclusion.tex
\section{Conclusion}
\label{sec:conclusion}
We have presented \acrfull{vdm}, a sequential latent variable model for multi-modal dynamics. The main contribution is a new variational family. It propagates multiple samples through an RNN to parametrize the posterior approximation with a mixture density network. Additionally, we have introduced the empirical Wasserstein distance for the evaluation of multi-modal forecasting tasks, since it accounts for forecast accuracy and diversity. \gls{vdm} succeeds in learning challenging multi-modal dynamics and outperforms existing work in various applications.

%% file: supplement.tex
\appendix
\section{Supplementary to Weighting function}
\label{sec:omega_appendix}
In this Appendix we give intuition for our choice of weighting function \cref{eqn:pi}. Since we approximate the integrals in \cref{eqn:posterior,eqn:expected_h} with samples from $\Tilde{q}(\mathbf{s}_{t-1}\mid \mathbf{x}_{<t})$ \footnote{The $\sim$ just helps to visually distinguish the two distributions that appear in the main text.} instead of samples from $q(\mathbf{s}_{t-1}\mid \mathbf{x}_{\leq t})$, importance sampling tells us that the weigths should be  
\begin{align}
\omega(\mathbf{s}_{t-1},\mathbf{x}_{t}) &= \frac{q(\mathbf{s}_{t-1}\mid \mathbf{x}_{\leq t})}{\Tilde{q}(\mathbf{s}_{t-1}\mid \mathbf{x}_{<t})}
=\frac{q(\mathbf{x}_{t} \mid \mathbf{s}_{t-1},\mathbf{x}_{<t})}{q(\mathbf{x}_{t} \mid \mathbf{x}_{< t})}\frac{\Tilde{q}(\mathbf{s}_{t-1}\mid \mathbf{x}_{<t})}{\Tilde{q}(\mathbf{s}_{t-1}\mid \mathbf{x}_{<t})}\nonumber\\  
&= \frac{q(\mathbf{x}_{t} \mid \mathbf{s}_{t-1},\mathbf{x}_{<t})}{q(\mathbf{x}_{t} \mid \mathbf{x}_{< t})}\propto q(\mathbf{x}_{t} \mid \mathbf{s}_{t-1},\mathbf{x}_{<t})
\end{align}
This is consistent with out earlier definition of $q(\mathbf{s}_{t-1}\mid \mathbf{x}_{\leq t})=\omega(\mathbf{s}_{t-1},\mathbf{x}_{t})\Tilde{q}(\mathbf{s}_{t-1}\mid \mathbf{x}_{<t})$.
The weights are proportional to the likelihood of the variational model $q(\mathbf{x}_{t} \mid \mathbf{s}_{t-1},\mathbf{x}_{<t})$. We choose to parametrize it using the likelihood of the generative model $p(\mathbf{x}_{t} \mid \mathbf{h}_{t-1}=\mathbf{s}_{t-1})$ and get 
\begin{align}
        \omega_t^{(i)}&=\omega(\mathbf{s}_{t-1}^{(i)}, \mathbf{x}_{t})/k \coloneqq \mathbbm{1}(i=\argmax_{j} p(\mathbf{x}_t\mid \mathbf{h}_{t-1}= \mathbf{s}_{t-1}^{(j)})).
\end{align}
With this choice of the weighting function, only the mixture component with the highest likelihood is selected to be in charge of modeling the current observation $\mathbf{x}_t$. As a result, other mixture components have the capacity to focus on different modes. This helps avoid the effect of mode-averaging. An alternative weight function is given in \cref{sec:appendix_results}.

\section{Supplementary to Lower Bound}
\label{sec:elbo_appendix}
\begin{claim-non}
The \gls{elbo} in \cref{eqn:elbo} is a lower bound on the log evidence $\log p(\mathbf{x}_{t} \mid \mathbf{x}_{<t})$,
\begin{equation}
        \log p(\mathbf{x}_{t} \mid \mathbf{x}_{<t})\geq \mathcal{L}_{\mathrm{ELBO}}(\mathbf{x}_{\leq t},\phi)\,.
\end{equation}
\end{claim-non}

\begin{proof}
We write the data evidence as the double integral over the latent variables $\mathbf{z}_{t}$, and $\mathbf{z}_{<t}$. 
\begin{align}
    \log p(\mathbf{x}_{t} \mid \mathbf{x}_{<t})&=\log \iint p(\mathbf{x}_{t} \mid \mathbf{z}_{\leq t},\mathbf{x}_{<t})p(\mathbf{z}_{t} \mid \mathbf{z}_{<t},\mathbf{x}_{<t})p(\mathbf{z}_{<t} \mid \mathbf{x}_{<t}) \mathrm{d}\mathbf{z}_{t}\mathrm{d}\mathbf{z}_{<t}
\end{align}
We multiply the posterior at the previous time step $p(\mathbf{z}_{<t} \mid \mathbf{x}_{<t})$ with the ratio of the approximated posterior $\frac{q(\mathbf{z}_{<t} \mid \mathbf{x}_{<t})}{q(\mathbf{z}_{<t} \mid \mathbf{x}_{<t})}$ and the ratio $\frac{f(\mathbf{a},\mathbf{b})}{f(\mathbf{a},\mathbf{b})}$, where $f$ is any suitable function of two variables $\mathbf{a}$ and $\mathbf{b}$. The following equality holds, since the ratios equal to one.
\begin{align}
    &\log p(\mathbf{x}_{t} \mid \mathbf{x}_{<t})\nonumber\\
    &= \log \int\frac{f(\mathbf{a},\mathbf{b})}{f(\mathbf{a},\mathbf{b})}\frac{q(\mathbf{z}_{<t} \mid \mathbf{x}_{<t})}{q(\mathbf{z}_{<t} \mid \mathbf{x}_{<t})}p(\mathbf{z}_{<t} \mid \mathbf{x}_{<t})\int p(\mathbf{x}_{t} \mid \mathbf{z}_{\leq t},\mathbf{x}_{<t})p(\mathbf{z}_{t} \mid \mathbf{z}_{<t},\mathbf{x}_{<t})\mathrm{d}\mathbf{z}_{t}\mathrm{d}\mathbf{z}_{<t}
\end{align}
We move the integral over $\mathbf{z}_{<t}$ with respect to $f(\mathbf{a},\mathbf{b})q(\mathbf{z}_{<t} \mid \mathbf{x}_{<t})$ out of the log operation with applying the Jensen's inequality.
\begin{align}
    \log p(\mathbf{x}_{t} \mid \mathbf{x}_{<t}) &\geq \mathbb{E}_ {f(\mathbf{a},\mathbf{b})q(\mathbf{z}_{<t} \mid \mathbf{x}_{<t})} \left[\log \int p(\mathbf{x}_{t} \mid \mathbf{z}_{\leq t},\mathbf{x}_{<t})p(\mathbf{z}_{t} \mid \mathbf{z}_{<t},\mathbf{x}_{<t})\mathrm{d}\mathbf{z}_{t}\right]\\
    &
    - \mathbb{E}_ {f(\mathbf{a},\mathbf{b})q(\mathbf{z}_{<t} \mid \mathbf{x}_{<t})} \left[\log f(\mathbf{a},\mathbf{b})+\log\frac{q(\mathbf{z}_{<t} \mid \mathbf{x}_{<t})}{p(\mathbf{z}_{<t} \mid \mathbf{x}_{<t})} \right]\nonumber
\end{align}
We introduce the variational posterior $q(\mathbf{z}_{t} \mid \mathbf{z}_{<t},\mathbf{x}_{\leq t})$, and apply Jensen's inequality to replace the intractable integral $\log \int p(\mathbf{x}_{t} \mid \mathbf{z}_{\leq t},\mathbf{x}_{<t})p(\mathbf{z}_{t} \mid \mathbf{z}_{<t},\mathbf{x}_{<t})\mathrm{d}\mathbf{z}_{t}$ with its lower bound.
\begin{align}
    \log p(\mathbf{x}_{t} \mid \mathbf{x}_{<t}) &\geq \mathbb{E}_ {f(\mathbf{a},\mathbf{b})q(\mathbf{z}_{<t} \mid \mathbf{x}_{<t})} \left[\mathbb{E}_{q(\mathbf{z}_{t}\mid \mathbf{z}_{<t},\mathbf{x}_{\leq t})}\left[  \log\frac{p(\mathbf{x}_{t} \mid \mathbf{z}_{\leq t},\mathbf{x}_{<t})p(\mathbf{z}_{t} \mid  \mathbf{z}_{<t},\mathbf{x}_{<t})}{q(\mathbf{z}_{t}\mid \mathbf{z}_{<t},\mathbf{x}_{\leq t})}\right]\right]\nonumber\\
    &
    - \mathbb{E}_ {f(\mathbf{a},\mathbf{b})q(\mathbf{z}_{<t} \mid \mathbf{x}_{<t})} \left[\log f(\mathbf{a},\mathbf{b})+\log\frac{q(\mathbf{z}_{<t} \mid \mathbf{x}_{<t})}{p(\mathbf{z}_{<t} \mid \mathbf{x}_{<t})} \right]\,.  
\end{align}
The expectation with respect to  $f(\mathbf{a},\mathbf{b})q(\mathbf{z}_{<t} \mid \mathbf{x}_{<t})$ is approximated with samples. Instead of resampling the entire history, samples from previous time steps are reused (they have been aggregated by the RNN) and we sample according to \cref{eqn:qtilde}. We plugg in the weighting function $\omega(\mathbf{s}_{t-1}^{(i)},\mathbf{x}_{t})$ for $f(\mathbf{a},\mathbf{b})$. The term $\log\frac{q(\mathbf{z}_{<t} \mid \mathbf{x}_{<t})}{p(\mathbf{z}_{<t} \mid \mathbf{x}_{<t})}$ is not affected by the incoming observation $\mathbf{x}_{t}$ and can be treated as a constant. 

In this step, we plug in our generative model and inference model as they are described in the main text for $p$ and $q$. The conditional independence assumptions can be read of \cref{fig:schematic_illustration}.
 In the generative model $\mathbf{h}_{t-1}$  and in the inference model $\mathbf{s}_{t-1}$ summarize the dependencies of $\mathbf{z}_{t}$ on the previous latent variables $\mathbf{z}_{<t}$ and observations $\mathbf{x}_{<t}$. In other words, we assume $\mathbf{z}_{t}$ is conditionally independent on $\mathbf{z}_{<t}$ and $\mathbf{x}_{<t}$ given $\mathbf{s}_{t-1}^{(i)}$ in the inference model (or given $\mathbf{h}_{t-1}$ in the generative model).
\begin{align}
    \log p(\mathbf{x}_{t} \mid \mathbf{x}_{<t}) &\geq \frac{1}{k}\sum_i^k \omega(\mathbf{s}_{t-1}^{(i)},\mathbf{x}_{t}) \mathbb{E}_{q(\mathbf{z}_{t}\mid \mathbf{s}_{t-1}^{(i)},\mathbf{x}_{ t})}\left[\log p(\mathbf{x}_{t} \mid \mathbf{z}_{t},\mathbf{h}_{t-1} = \mathbf{s}_{t-1}^{(i)})\right]\nonumber\\
   & +  \frac{1}{k}\sum_i^k \omega(\mathbf{s}_{t-1}^{(i)},\mathbf{x}_{t}) \mathbb{E}_{q(\mathbf{z}_{t}\mid \mathbf{s}_{t-1}^{(i)},\mathbf{x}_{ t})}\left[\log\frac{p(\mathbf{z}_{t} \mid  \mathbf{h}_{t-1} = \mathbf{s}_{t-1}^{(i)})}{q(\mathbf{z}_{t}\mid \mathbf{s}_{t-1}^{(i)},\mathbf{x}_{t})}\right]\nonumber\\
    &- \frac{1}{k}\sum_i^k \omega(\mathbf{s}_{t-1}^{(i)},\mathbf{x}_{t})\left[\log \omega(\mathbf{s}_{t-1}^{(i)},\mathbf{x}_{t})+\mathbf{C}\right] 
\end{align}
\end{proof}

\section{Algorithms of Generative Model and Inference Model}
\label{sec:algo_appendix}
\begin{minipage}{0.4\textwidth}
	\begin{algorithm}[H]
		\caption{Generative model}
		\label{alg:gen_pcode}
		\small
		\begin{algorithmic}
			\STATE {\bf Inputs:} 
			$[\mathbf{\mu}_{z,\tau}, \mathbf{\sigma}_{z,\tau}^2], \mathbf{h}_{\tau-1}$
			\STATE {\bf Outputs:} 
			$\mathbf{x}_{\tau+1:T}$
			\STATE $\mathbf{z}_\tau \sim \mathcal{N}(\mathbf{\mu}_{z,\tau}, \mathbf{\sigma}_{z,\tau}^2\mathbb{I})$
			\STATE $\mathbf{h}_{\tau} = \phi^{\text{GRU}}(\mathbf{z}_{\tau}, \mathbf{h}_{\tau-1})$
			\FOR{$t= \tau+1 : T$}
			\STATE $[\mathbf{\mu}_{0,t}, \mathbf{\sigma}_{0,t}^2] = \phi^{tra}(\mathbf{h}_{t-1})$  
			\STATE $\mathbf{z}_t \sim \mathcal{N}(\mathbf{\mu}_{0,t}, \mathbf{\sigma}_{0,t}^2\mathbb{I})$
			\STATE $\mathbf{h}_{t} = \phi^{\text{GRU}}(\mathbf{z}_{t}, \mathbf{h}_{t-1})$
			\STATE $[\mathbf{\mu}_{x,t}, \mathbf{\sigma}_{x,t}^2] = \phi^{dec}(\mathbf{z}_{t},\mathbf{h}_{t-1})$
			\STATE $\mathbf{x}_t \sim \mathcal{N}(\mathbf{\mu}_{x,t}, \mathbf{\sigma}_{x,t}^2\mathbb{I})$
			\ENDFOR
		\end{algorithmic}
	\end{algorithm}
\end{minipage}
\hfill
\begin{minipage}{0.53\textwidth}
	\begin{algorithm}[H]
		\caption{Inference model}
		\label{alg:inf_pcode}
		\small
		\begin{algorithmic}
			\STATE {\bf Inputs:} $\mathbf{x}_{1:\tau},\mathbf{h}_0$
			\STATE {\bf Outputs:} 
			$[\mathbf{\mu}_{z,1:\tau}, \mathbf{\sigma}_{z,1:\tau}^2],\mathbf{h}_{\tau-1}$
			\STATE 
			$[\mathbf{\mu}_{z,1}, \mathbf{\sigma}_{z,1}^2] = \phi^{inf}(\mathbf{h}_{0},\mathbf{x}_1)$

			\FOR{$t= 2 : \tau$}
            \STATE
            $\mathbf{z}_{t-1}^{(i)} \sim \mathcal{N}(\mathbf{\mu}_{z,t-1}, \mathbf{\sigma}_{z,t-1}^2\mathbb{I})$
			\STATE 
			$\mathbf{s}_{t-1}^{(i)} = \phi^{\text{GRU}}(\mathbf{z}_{t-1}^{(i)}, \mathbf{h}_{t-2})$
            \STATE 
            $[\mathbf{\mu}_{z,t}^{(i)}, \mathbf{\sigma}_{z,t}^{(i)2}] = \phi^{inf}( \mathbf{s}_{t-1}^{(i)},\mathbf{x}_{t})$
			\STATE 
			$\omega_t^{(i)} \coloneqq \mathbbm{1}(i=\argmax_{j} p(\mathbf{x}_t\mid \mathbf{h}_{t-1}= \mathbf{s}_{t-1}^{(j)}))$
			\STATE 
			$[\mathbf{\mu}_{z,t}, \mathbf{\sigma}_{z,t}^2]=\sum_i^k \omega_t^{(i)} \mathcal{N}(\mathbf{\mu}_{z,t}^{(i)}, \mathbf{\sigma}_{z,t}^{(i)2}\mathbb{I})$
			\STATE 
			$\mathbf{h}_{t-1} \approx \sum_i^k \omega_t^{(i)}\mathbf{s}_{t-1}^{(i)}$
			\ENDFOR
		\end{algorithmic}
	\end{algorithm}
\end{minipage}
\section{Supplementary to Stochastic Cubature Approximation}
\label{sec:cubature_appendix}
\paragraph*{Cubature approximation.} The cubature approximation is widely used in the engineering community as a deterministic method to numerically integrate a nonlinear function $f(\cdot)$ of Gaussian random variable $z\sim \mathcal{N}( \mu_\mathbf{z}, \sigma^2_\mathbf{z}\mathbb{I})$, with $\mathbf{z} \in \mathbb{R}^d$. The method proceeds by constructing $2d + 1$ sigma points $\mathbf{z}^{(i)}=\mu_\mathbf{z} + \sigma_\mathbf{z}\xi^{(i)}$.
The cubature approximation is simply a weighted sum of the sigma points propagated through the nonlinear function $f(\cdot)$, 
\begin{equation}
\int f(\mathbf{z})\mathcal{N}(\mathbf{z} \mid \mu_\mathbf{z}, \sigma^2_\mathbf{z}\mathbb{I}) \mathrm{d}\mathbf{z} \approx
\sum_{i = 1}^{2d + 1} \gamma^{(i)} f(\mathbf{z}^{(i)})\,.
\end{equation}
Simple analytic formulas determine the computation of weights $\gamma^{(i)}$ and the locations of $\xi^{(i)}$.

\begin{tabularx}{\textwidth}{XX}
{\begin{align*}
\label{eqn:cubature_general}
\gamma^{(i)} = 
\begin{cases}
\frac{1}{2(n+\kappa)} &, i=1, ..., 2n\\
\frac{\kappa}{n+\kappa} &, i=0
\end{cases}
\end{align*}}
&\quad
{\begin{align}
\xi^{(i)}= 
\begin{cases}
~~~\sqrt{n+\kappa}\mathbf{e}_i&, i=1, ..., n\\
-\sqrt{n+\kappa}\mathbf{e}_{i-n}&, i=n+1, ..., 2n\\
~~~~~~~~~~0&, i=0\,,
\end{cases}
\end{align}}
\end{tabularx}
where $\kappa$ is a hyperparameter controlling the spread of the sigma points in the $n$-dimensional sphere. Further $\mathbf{e}_i$ represents a basis in the $n$-dimensional space, which is choosen to be a unit vector in cartesian space, e.g. $\mathbf{e}_1 = [1, 0, ..., 0]$.

\paragraph*{Stochastic cubature approximation.} In \gls{sca}, we adopt the computation of $\xi^{(i)}$ in \cref{eqn:cubature_general}, and infuse the sigma points with standard Gaussian noise $\epsilon \sim \mathcal{N}(0, \mathbb{I})$ to obtain stochastic {\em sigma variables} $\mathbf{s}^{(i)}=\mu_\mathbf{z} + \sigma_\mathbf{z}(\xi^{(i)}+\epsilon)$. We choose $\kappa=0.5$ to set the weights $\gamma^{(i)}$ equally. 
\section{Supplementary to Ablation Study of Regularization Terms}
\label{sec:study_regularizations}
We investigate the effect of the regularization terms using the synthetic data from \cref{fig:toy}. We can see in \cref{tab:toy}, $\text{VDM}(k=9)$ can be trained successfully with $\mathcal{L}_{\mathrm{ELBO}}$ only, and both regularization terms improve the performance (negative log-likelihood of multi-steps ahead prediction), while $\text{VDM}(k=1)$ doesn't work whatever the regularization terms. Additionally, we tried to train the model only with the regularization terms (each separate or together) but these options diverged during training.
\begin{table}[h]  
 	\caption{ Ablation study of the regularization terms for synthetic data from \cref{fig:toy}}
    \label{tab:toy}
	\centering
\begin{tabular}{ l|cccc } 
\hline
 & $\mathcal{L}_{\mathrm{ELBO}}$ 
 & $\mathcal{L}_{\mathrm{ELBO}}\&\mathcal{L}_{pred}$
 &$\mathcal{L}_{\mathrm{ELBO}}\&\mathcal{L}_{adv}$
 &$\mathcal{L}_{\mathrm{VDM}}$
 \\
\hline
$\text{VDM}(k=9)$ & 2.439$\pm$0.005 & 2.379$\pm$0.008&2.381$\pm$0.006 &\textbf{2.363}$\pm$0.004 \\
\hline
$\text{VDM}(k=1)$ &3.756$\pm$0.003  &3.960$\pm$0.008&3.743$\pm$0.005 &3.878$\pm$0.007 \\
\hline
\end{tabular}
\end{table}

\section{Supplementary to Experiments Setup}
\subsection{Stochastic lorenz attractor setup}
\label{sec:lorenz_exp_appendix}
Lorenz attractor is a system of three ordinary differential equations:
\begin{equation}
    \frac{\mathrm{d}\mathbf{x}}{\mathrm{d}t}=\sigma(\mathbf{y}-\mathbf{x}),\quad \frac{\mathrm{d}\mathbf{y}}{\mathrm{d}t}=\mathbf{x}(\rho-\mathbf{z})-\mathbf{y},\quad
    \frac{\mathrm{d}\mathbf{z}}{\mathrm{d}t}=\mathbf{x}\mathbf{y}-\beta\mathbf{z} \,,   
\end{equation}
where $\sigma$, $\rho$, and $\beta$ are system parameters. We set $\sigma=10,\,\rho=28$ and $\beta=8/3$ to make the system chaotic. We simulate the trajectories by RK4 with a step size of 0.01. To make it stochastic, we add process noise to the transition, which is a mixture of two Gaussians $0.5\mathcal{N}(\mathbf{m}_0,\mathbf{P})+0.5\mathcal{N}(\mathbf{m}_2,\mathbf{P})$, where
\begin{equation}
    \mathbf{m}_0= \begin{bmatrix} 
	0  \\
	1 \\
	0
	\end{bmatrix},\quad
    \mathbf{m}_1= \begin{bmatrix} 
	0  \\
	-1 \\
	0
	\end{bmatrix},\quad	
	\mathbf{P}= \begin{bmatrix} 
	0.06&0.03&0.01\\
	0.03&0.03&0.03 \\
	0.01&0.03&0.05
	\end{bmatrix}.
\end{equation}
Besides, we add a Gaussian noise with zero mean and diagonal standard deviation $[0.6,0.4,0.8]$ as the observation noise. Totally, we simulate 5000 sequences as training set, 200 sequences as validation set, and 800 sequences as test set. For evaluation of Wasserstein distance, we simulate 10 groups of sequences additionally. Each group has 100 sequences with similar initial observations.

\subsection{Taxi trajectories setup}
\label{sec:taxi_exp_appendix}
The full dataset is very large and the length of trajectories varies. We select the trajectories inside the Porto city area with length in the range of 30 and 45, and only extract the first 30 coordinates of each trajectory. Thus we obtain a dataset with a fixed sequence length of 30. We split it into the training set of size 86386, the validation set of size 200, and the test set of size 10000.
\subsection{U.S. pollution data setup}
\label{sec:pollution_exp_appendix}
The U.S. pollution dataset consists of four pollutants (NO2, O3, SO2 and O3). Each of them has 3 major values (mean, max value, and air quality index). It is collected from counties in different states for every day from 2000 to 2016. Since the daily measurements are too noisy, we firstly compute the monthly average values of each measurement, and then extract non-overlapped segments with the length of 24 from the dataset. Totally we extract 1639 sequences as training set, 25 sequences as validation set, and 300 sequences as test set.
\subsection{NBA SportVu data setup}
\label{sec:player_exp_appendix}
We use a sliding window of the width 30, and the stride 30 to cut the long sequences to short sequences of a fixed length 30. We split them into the training set of size 8324, the validation set of size 489, and the test set of size 980.
\section{Implementation Details}
\label{sec:imp_appendix}
Here, we provide implementation details of \gls{vdm} models used across the three datasets in the main paper. \gls{vdm} consists of
\begin{itemize}
    \item encoder: embed the first observation $\mathbf{x}_0$ to the latent space as the initial latent state $\mathbf{z}_0$.
    \item transition network: propagate the latent states $\mathbf{z}_t$.
    \item decoder: map the latent states $\mathbf{z}_t$ and the recurrent states $\mathbf{h}_t$ to observations $\mathbf{x}_t$.
    \item inference network: update the latent states $\mathbf{z}_t$ given observations $\mathbf{x}_t$.
    \item latent GRU: summarize the historic latent states $\mathbf{z}_{\leq t}$ in the recurrent states $\mathbf{h}_t$.
    \item discriminator: be used for adversarial training.
\end{itemize}
The optimizer is Adam with the learning rate of $1e-3$. In all experiments, the networks have the same architectures but different sizes. The model size depends on observation dimension $\mathbf{d_x}$, latent state dimension $\mathbf{d_z}$, and recurrent state dimension $\mathbf{d_h}$. The number of samples used at each time step in the training is $2\mathbf{d_z}+1$. If the model output is variance, we use the exponential of it to ensure its non-negative.
\begin{itemize}
    \item Encoder: input size is $\mathbf{d_x}$; 3 linear layers of size 32, 32 and $2\mathbf{d_z}$, with 2 ReLUs.
    \item Transition network: input size is $\mathbf{d_h}$; 3 linear layers of size 64, 64, and $2\mathbf{d_z}$, with 3 ReLUs.
    \item Decoder: input size is $\mathbf{d_h}+\mathbf{d_z}$; 3 linear layers of size 32, 32 and $2\mathbf{d_x}$, with 2 ReLUs.
    \item Inference network: input size is $\mathbf{d_h}+\mathbf{d_x}$; 3 linear layers of size 64, 64, and $2\mathbf{d_z}$, with 3 ReLUs.
    \item Latent GRU: one layer GRU of input size $\mathbf{d_z}$ and hidden size $\mathbf{d_h}$
    \item Discriminator: one layer GRU of input size $\mathbf{d_x}$ and hidden size $\mathbf{d_h}$ to summarize the previous observations as the condition, and a stack of 3 linear layers of size 32, 32 and 1, with 2 ReLUs and one sigmoid as the output activation, whose input size is $\mathbf{d_h}+\mathbf{d_x}$. 
\end{itemize}
\paragraph*{Stochastic Lorenz attractor.} Observation dimension $\mathbf{d_x}$ is 3, latent state dimension $\mathbf{d_z}$ is 6, and recurrent state dimension $\mathbf{d_h}$ is 32.
\paragraph*{Taxi trajectories.} Observation dimension $\mathbf{d_x}$ is 2, latent state dimension $\mathbf{d_z}$ is 6, and recurrent state dimension $\mathbf{d_h}$ is 32.
\paragraph*{U.S. pollution data\footnote{https://www.kaggle.com/sogun3/uspollution}} Observation dimension $\mathbf{d_x}$ is 12, latent state dimension $\mathbf{d_z}$ is 8, and recurrent state dimension $\mathbf{d_h}$ is 48.
\paragraph*{NBA SportVu data.} Observation dimension $\mathbf{d_x}$ is 2, latent state dimension $\mathbf{d_z}$ is 6, and recurrent state dimension $\mathbf{d_h}$ is 32.

Here, we give the number of parameters for each model in different experiments in \cref{tab:num_params}.
\begin{table}[h]
	\caption{Number of parameters for each model in three experiments. \gls{vdm}, \gls{aesmc}, \gls{vrnn}, and \gls{rkn} have comparable number of parameters. \gls{cfvae} has much more parameters.}
	\label{tab:num_params}
	\centering
	\begin{tabular}{l|ccccc}
		\hline
        &\gls{rkn}&\gls{vrnn}&\gls{cfvae}&\gls{aesmc}&\gls{vdm}\\
        \hline
        Lorenz&23170&22506&7497468&22218&22218\\
        Taxi&23118&22248&7491123&22056&22056\\
        Pollution&35774&33192&8162850&31464&31464\\
        SportVu&23118&22248&7491123&22056&22056\\
		\hline 
	\end{tabular}
\end{table}
\section{Additional Evaluation Results}
\label{sec:appendix_results}
We evaluate more variants of \gls{vdm} in the chosen experiments to investigate the different choices of sampling methods (Monte Carlo method, and \gls{sca}) and weighting functions (\cref{eqn:omega_1,eqn:omega_2}). In addition to \cref{eqn:omega_1} described in the main text, we define one other choice in \cref{eqn:omega_2}.
\begin{align}
\label{eqn:omega_1}
    \omega_t^{(i)}&=\omega(\mathbf{s}_{t-1}^{(i)}, \mathbf{x}_{t})/k \coloneqq \mathbbm{1}(i=\argmax_{j} p(\mathbf{x}_t\mid \mathbf{h}_{t-1}= \mathbf{s}_{t-1}^{(j)}))
\\
\label{eqn:omega_2}
    \omega_t^{(i)}&=\omega(\mathbf{s}_{t-1}^{(i)}, \mathbf{x}_{ t})/k \coloneqq \mathbbm{1}(i=j\sim \text{Cat}(\cdot \mid \omega^1,\dots,\omega^k)), \quad \omega^j \propto p(\mathbf{x}_{t} \mid \mathbf{h}_{t-1}= \mathbf{s}_{t-1}^{(j)}),
\end{align}
We define the weighting function as an indicator function, in \cref{eqn:omega_1} we set the non-zero component by selecting the sample that achieves the highest likelihood, and in \cref{eqn:omega_2} the non-zero index is sampled from a categorical distribution with probabilities proportional to
the likelihood. The first choice (\cref{eqn:omega_1}) is named with $\delta$-function, and the second choice (\cref{eqn:omega_2}) is named with categorical distribution.
\begin{table}[h]
	\caption{Definition of \gls{vdm} variants}
	\label{tab:def_variants}
	\centering
	\begin{tabular}{l|cccc}
		\hline
        &$\text{VDM}(k=1)$&$\text{VDM-MC+}\delta$&$\text{VDM-SCA+Cat}$&$\text{VDM-SCA+}\delta$\\
        \hline
        Sampling method&Monte-Carlo&Monte-Carlo&\gls{sca}&\gls{sca}\\
        Weighting function& n.a.&$\delta$-function&Categorical distribution&$\delta$-function\\
		\hline 
	\end{tabular}
\end{table}
Besides, in $\text{VDM-Net}$, we evaluate the performance of replacing the closed-form inference of the weighting function with an additional inference network. In \cref{tab:def_variants}, we show the choices in different variants. All models are trained with $\mathcal{L}_{\mathrm{ELBO}}\&\mathcal{L}_{pred}$.
\subsection{Stochastic Lorenz attractor}
\label{sec:appendix_lorenz_results}
\begin{table}[ht]
	\caption{Ablation study of \gls{vdm}'s variants on stochastic Lorenz attractor for three distance metrics (see main text). The variants are defined in \cref{tab:def_variants}. All variants give comparable quantitative results.}
	\centering
	\resizebox{\linewidth}{!}{
	\begin{tabular}{l|ccccc}
		\hline
 &$\text{VDM}(k=1)$&$\text{VDM-Net}$&$\text{VDM-MC+}\delta$&$\text{VDM-SCA+Cat}$&$\text{VDM-SCA+}\delta$\\
		\hline
        Multi-steps &25.03$\pm$0.28&26.65$\pm$0.15 &24.67$\pm$0.16 &24.69$\pm$0.16 &24.49$\pm$0.16\\
        One-step &-1.81& -1.71 & -1.84 & -1.83 &-1.81\\
        W-distance  &7.31$\pm$0.002&7.68$\pm$0.002 &7.31$\pm$0.005 &7.30$\pm$0.009 &7.29$\pm$0.003 \\
		\hline 
	\end{tabular}}
\end{table}

\subsection{Taxi trajectories}
\label{sec:appendix_taxi_results}
\begin{table}[h]
	\caption{Ablation study of \gls{vdm}'s variants on taxi trajectories for three distance metrics (see main text). The variants are defined in \cref{tab:def_variants}. $\text{VDM-SCA+}\delta$ outperforms other variants and approaches our default \gls{vdm} (trained with $\mathcal{L}_{adv}$ additionally).}
	\centering
	\resizebox{\linewidth}{!}{
	\begin{tabular}{l|ccccc}
		\hline
 &$\text{VDM}(k=1)$&$\text{VDM-Net}$&$\text{VDM-MC+}\delta$&$\text{VDM-SCA+Cat}$&$\text{VDM-SCA+}\delta$\\
		\hline
        Multi-steps &3.26$\pm$0.001&3.68$\pm$0.002&3.17$\pm$0.001 &3.09$\pm$0.001 &2.88$\pm$0.002 \\
        One-step &-2.99&-2.74& -3.21 &  -3.24  & -3.68\\
        W-distance &0.69$\pm$0.0005&0.79$\pm$0.0003&0.70$\pm$0.0008 &0.64$\pm$0.0005 &0.59$\pm$0.0008 \\
		\hline 
	\end{tabular}}
\end{table}

\subsection{U.S. pollution data}
\label{sec:appendix_poll_results}
\begin{table}[h]
	\caption{Ablation study of \gls{vdm}'s variants on U.S. pollution data for two distance metrics (see main text). The variants are defined in \cref{tab:def_variants}. $\text{VDM-SCA+}\delta$ outperforms other variants.}
	\centering
	\resizebox{\linewidth}{!}{
	\begin{tabular}{l|ccccc}
		\hline
 &$\text{VDM}(k=1)$&$\text{VDM-Net}$&$\text{VDM-MC+}\delta$&$\text{VDM-SCA+Cat}$&$\text{VDM-SCA+}\delta$\\
		\hline
        Multi-steps &42.33$\pm$0.11&52.44$\pm$0.04 &40.33$\pm$0.03 &39.58$\pm$0.09 &37.64$\pm$0.07\\
        One-step &7.97&10.70& 8.12 & 7.82  &  6.91 \\
		\bottomrule 
	\end{tabular}}
\end{table}

\begin{figure}[h]
	\centering
	\begin{subfigure}[t]{0.19\textwidth}
		\includegraphics[width=\textwidth]{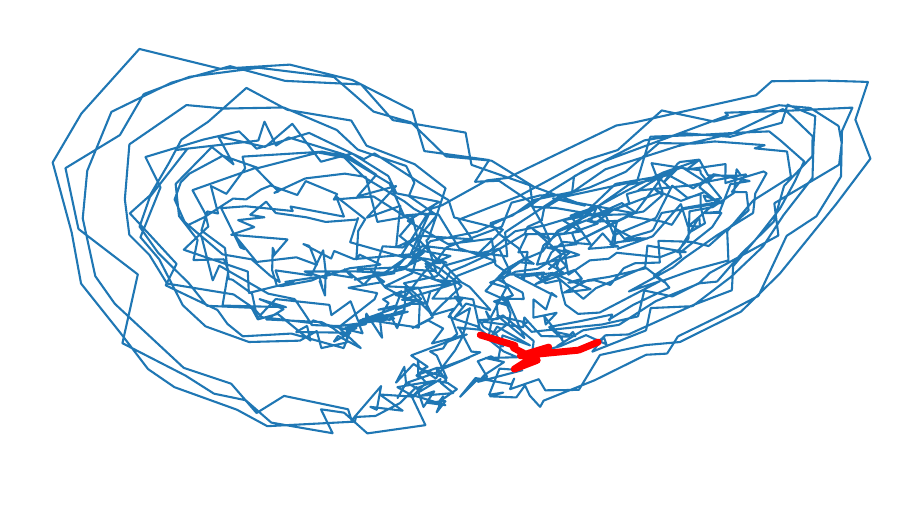}
		\caption{$\text{VDM-SCA+}\delta$}
	\end{subfigure}
		\begin{subfigure}[t]{0.19\textwidth}
		\includegraphics[width=\textwidth]{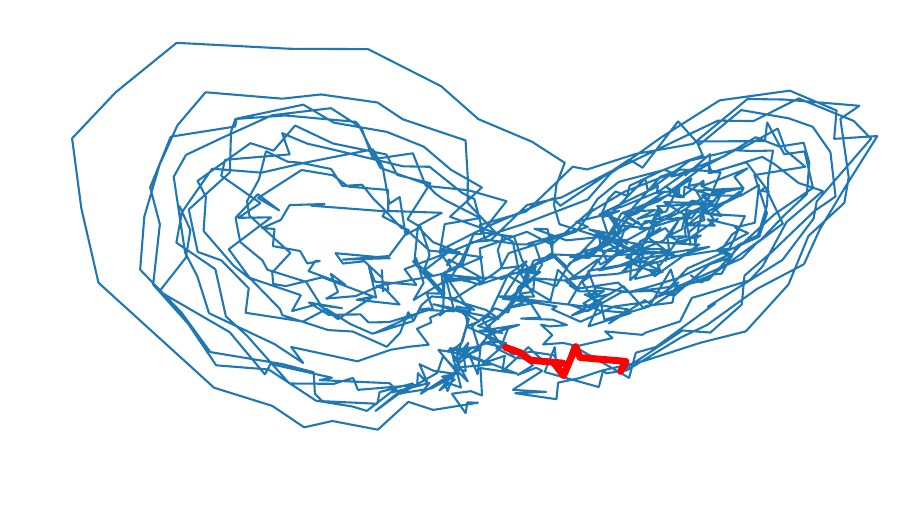}
		\caption{$\text{VDM-SCA+Cat}$}
	\end{subfigure}
		\begin{subfigure}[t]{0.19\textwidth}
		\includegraphics[width=\textwidth]{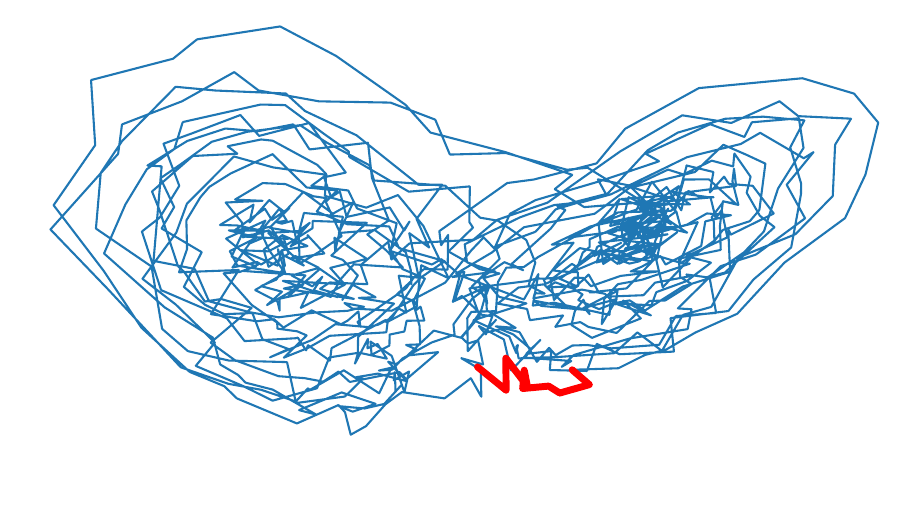}
		\caption{$\text{VDM-MC+}\delta$}
	\end{subfigure}
			\begin{subfigure}[t]{0.19\textwidth}
		\includegraphics[width=\textwidth]{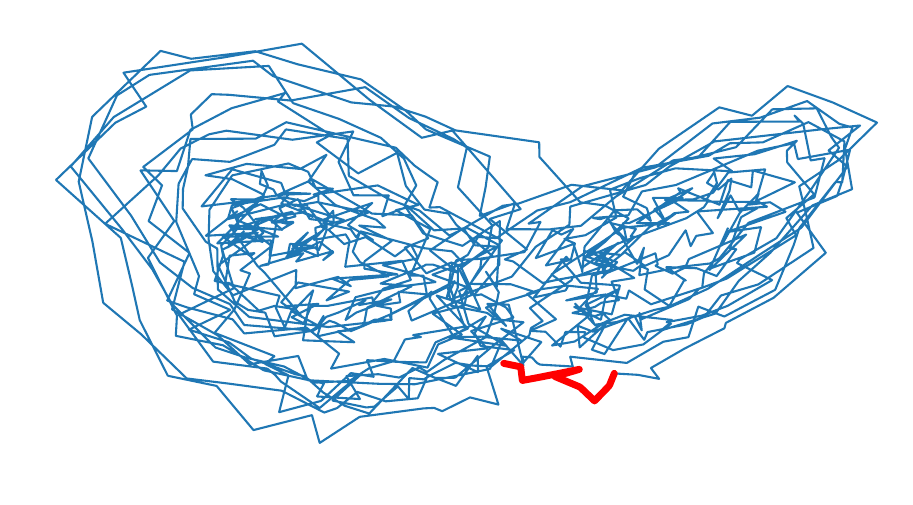}
		\caption{$\text{VDM-Net}$}
	\end{subfigure}
			\begin{subfigure}[t]{0.19\textwidth}
		\includegraphics[width=\textwidth]{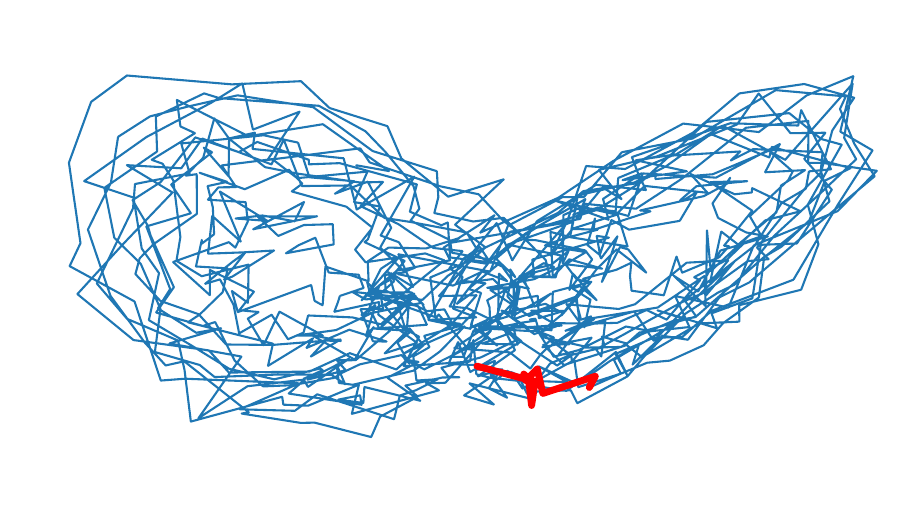}
		\caption{$\text{VDM}(k=1)$}
	\end{subfigure}
	\caption{Generated trajectories of stochastic Lorenz attractor from \gls{vdm} variants. The first ten observations (red) are obtained from models given the first 10 true observations. The rest 990 observations (blue) are predicted. We can see, all variants give very good qualitative results. Since the fundamental dynamics is govern by ordinary differential equations, the transition at each time step is not highly multi-modal. Once the model is equipped with a stochastic transition, it is able to model this dynamics.}
	\label{fig:appendix_lorenz}
\end{figure}
\begin{figure}[h]
	\centering
	\begin{subfigure}[t]{0.19\textwidth}
		\includegraphics[width=\textwidth]{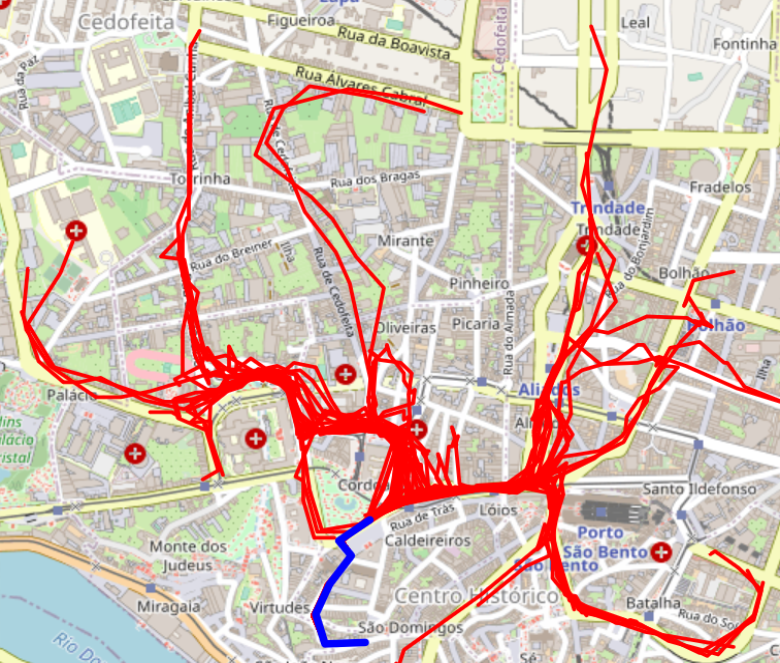}
	\end{subfigure}
	\begin{subfigure}[t]{0.19\textwidth}
		\includegraphics[width=\textwidth]{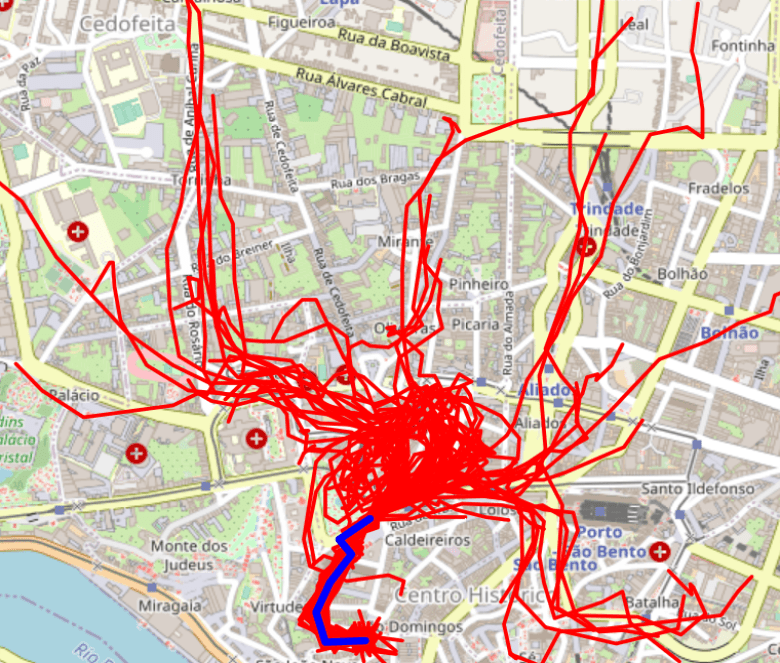}
	\end{subfigure}
	\begin{subfigure}[t]{0.19\textwidth}
		\includegraphics[width=\textwidth]{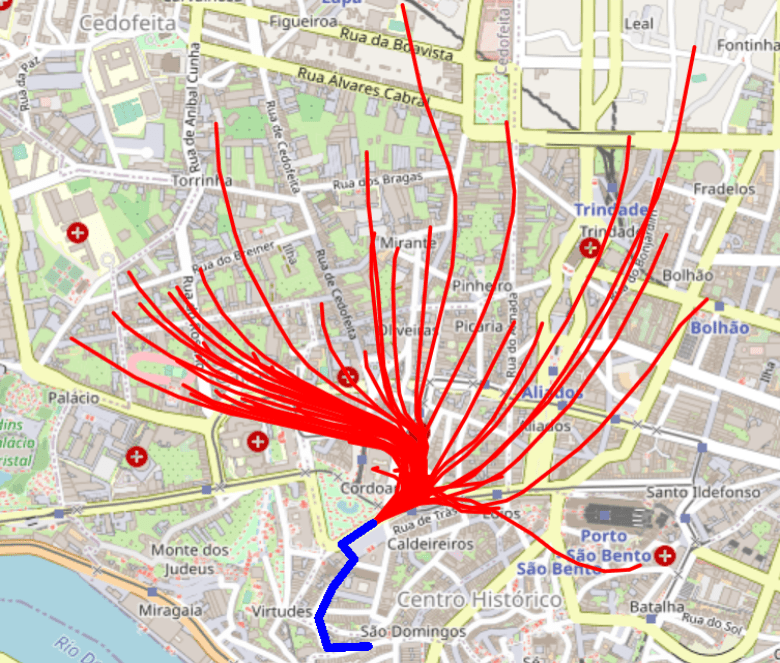}
	\end{subfigure}
	\begin{subfigure}[t]{0.19\textwidth}
		\includegraphics[width=\textwidth]{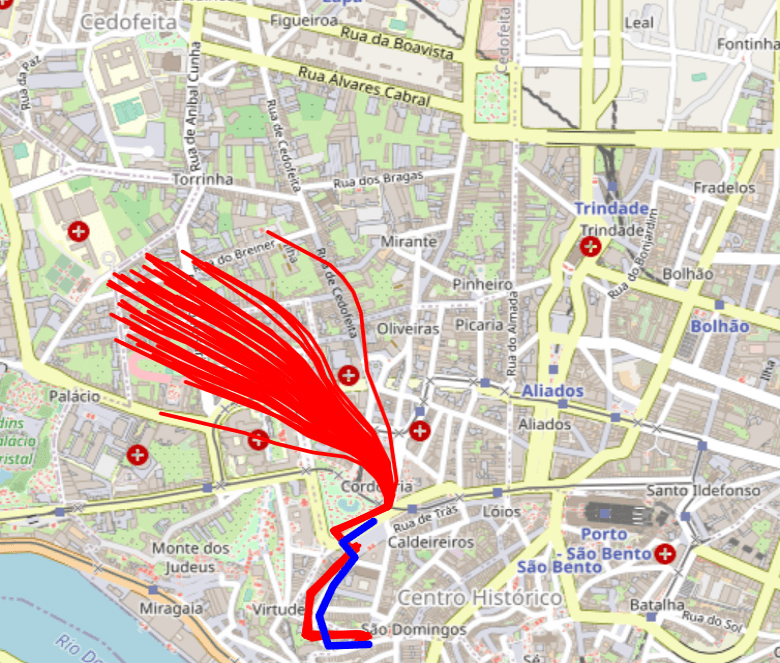}
	\end{subfigure}
	\begin{subfigure}[t]{0.19\textwidth}
		\includegraphics[width=\textwidth]{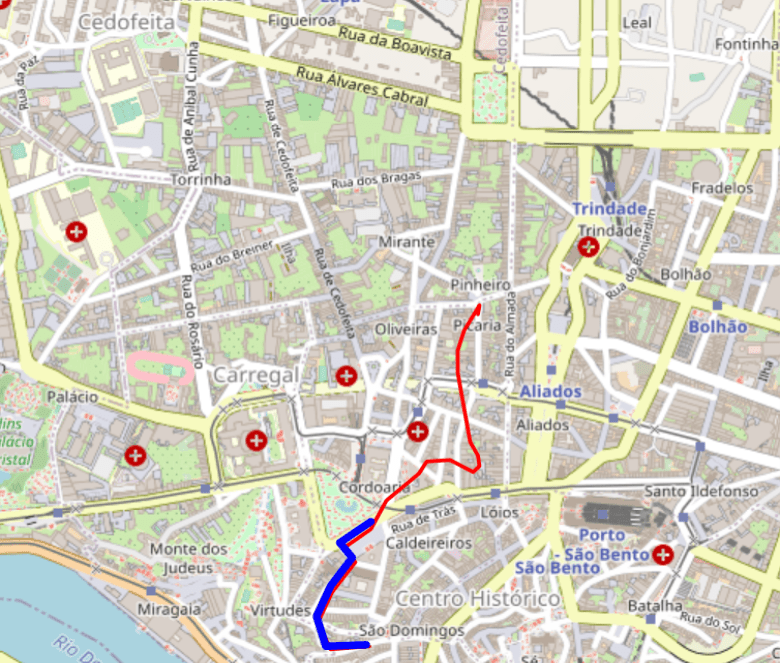}
	\end{subfigure}\\
	\begin{subfigure}[t]{0.19\textwidth}
		\includegraphics[width=\textwidth]{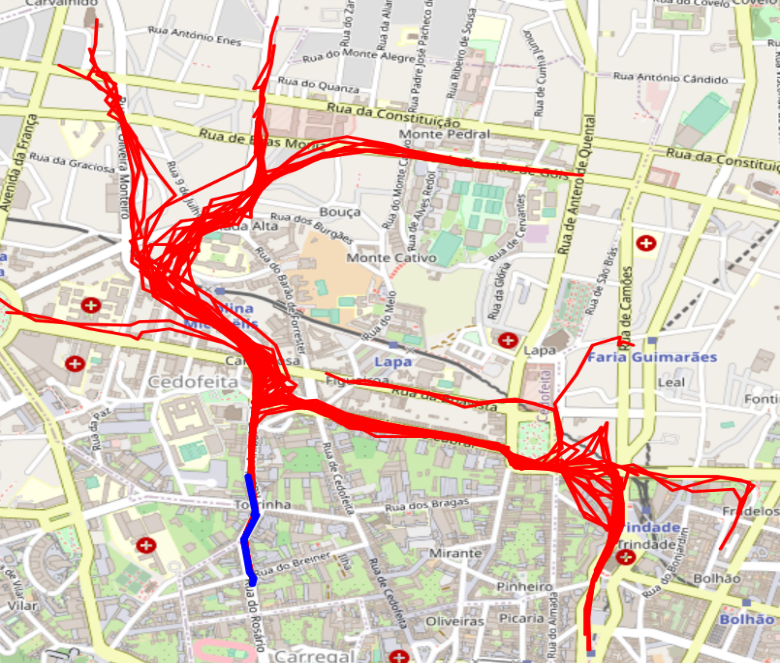}
	\end{subfigure}
	\begin{subfigure}[t]{0.19\textwidth}
		\includegraphics[width=\textwidth]{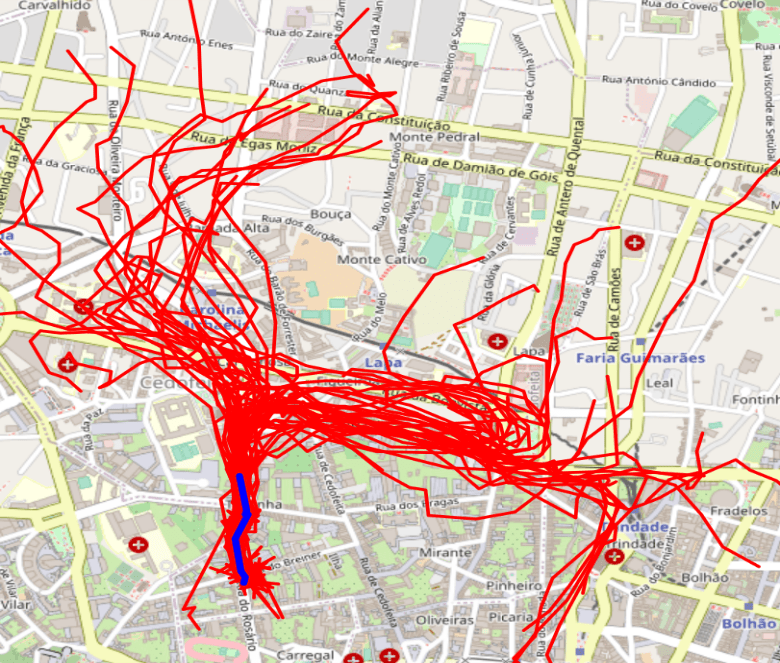}
	\end{subfigure}
	\begin{subfigure}[t]{0.19\textwidth}
		\includegraphics[width=\textwidth]{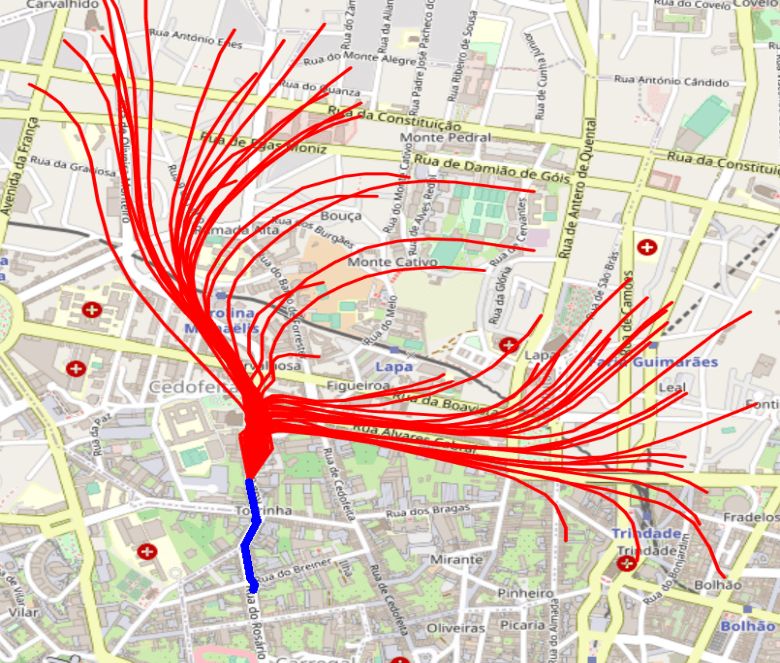}
	\end{subfigure}
	\begin{subfigure}[t]{0.19\textwidth}
		\includegraphics[width=\textwidth]{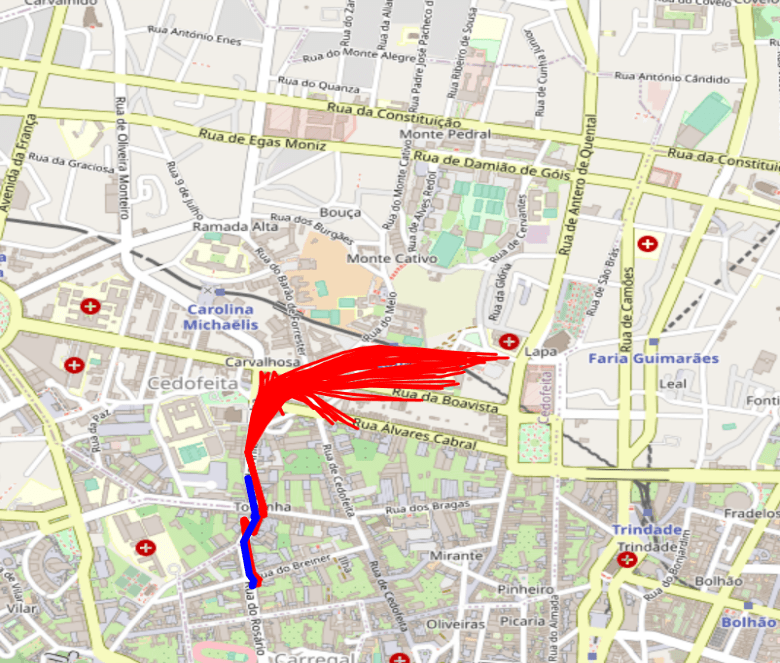}
	\end{subfigure}
	\begin{subfigure}[t]{0.19\textwidth}
		\includegraphics[width=\textwidth]{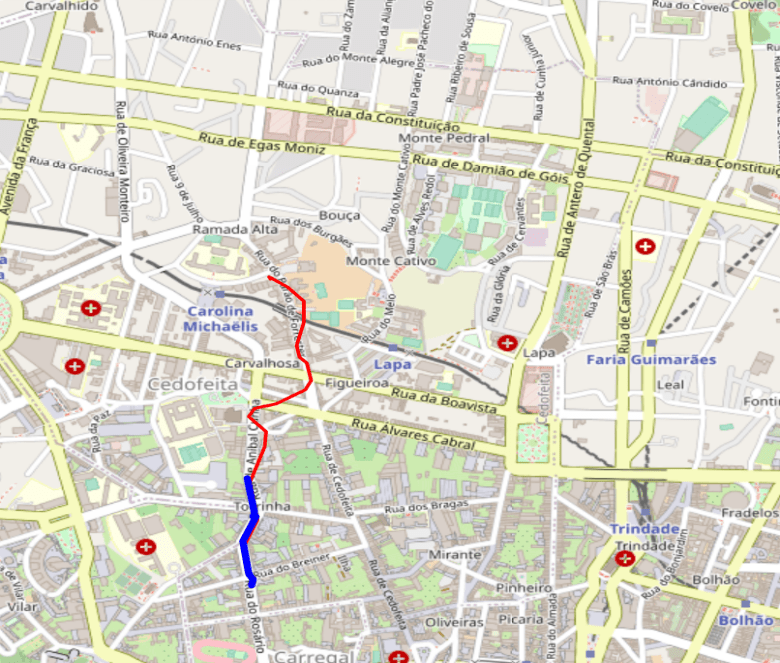}
	\end{subfigure}\\
	\begin{subfigure}[t]{0.19\textwidth}
		\includegraphics[width=\textwidth]{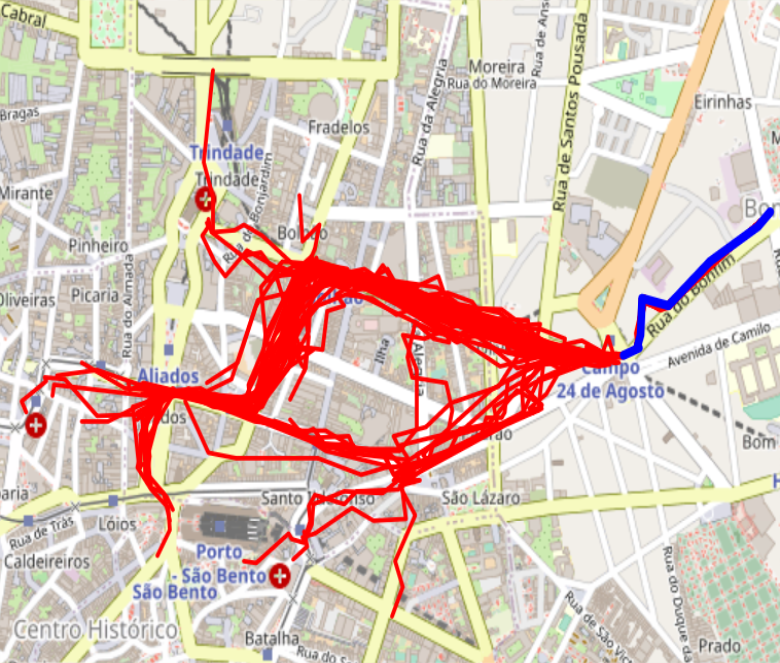}
		\caption{\gls{vdm} (ours)}
	\end{subfigure}
	\begin{subfigure}[t]{0.19\textwidth}
		\includegraphics[width=\textwidth]{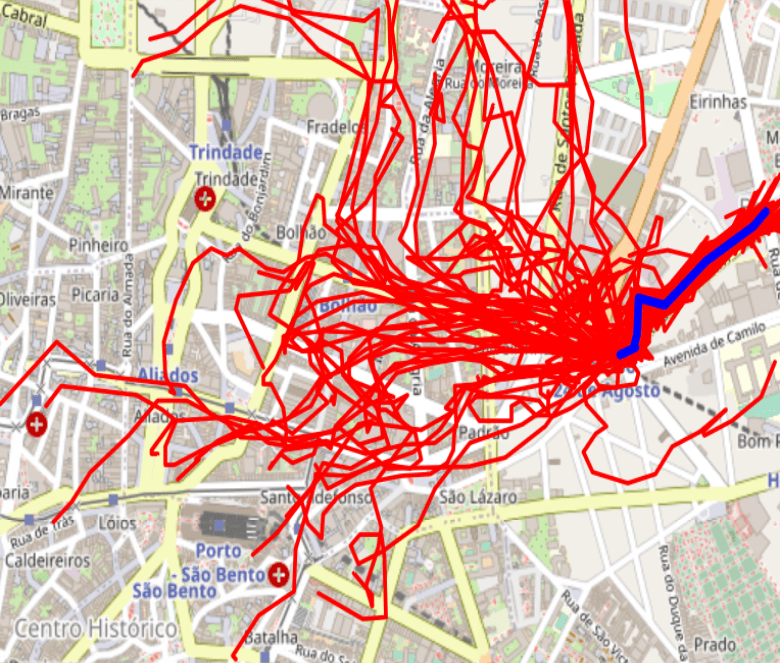}
		\caption{\acrshort{aesmc}}
	\end{subfigure}
	\begin{subfigure}[t]{0.19\textwidth}
		\includegraphics[width=\textwidth]{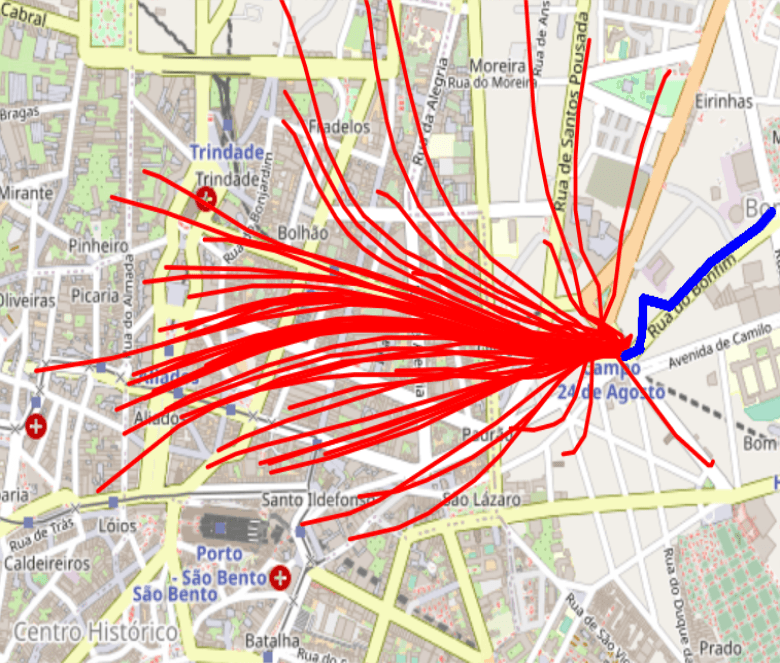}
		\caption{\acrshort{cfvae}}
	\end{subfigure}
	\begin{subfigure}[t]{0.19\textwidth}
		\includegraphics[width=\textwidth]{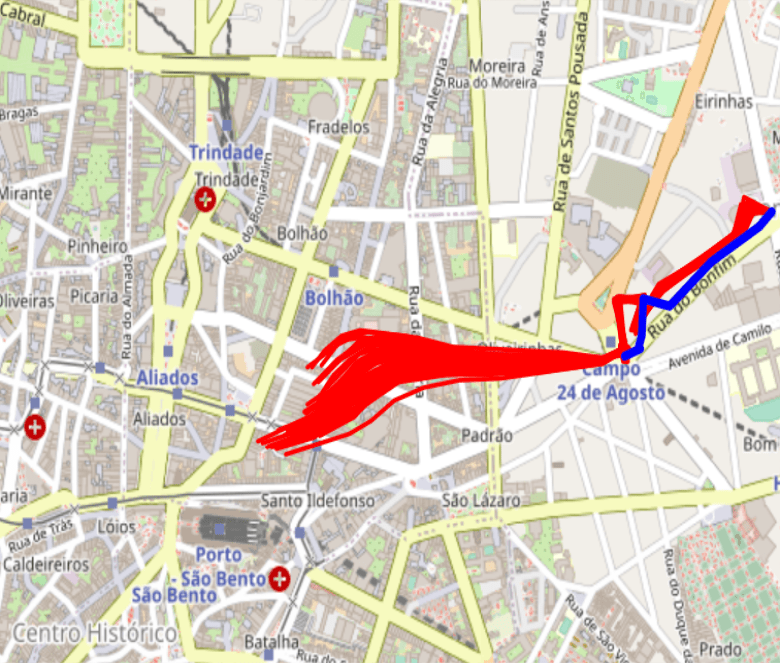}
		\caption{\acrshort{vrnn}}
	\end{subfigure}
	\begin{subfigure}[t]{0.19\textwidth}
		\includegraphics[width=\textwidth]{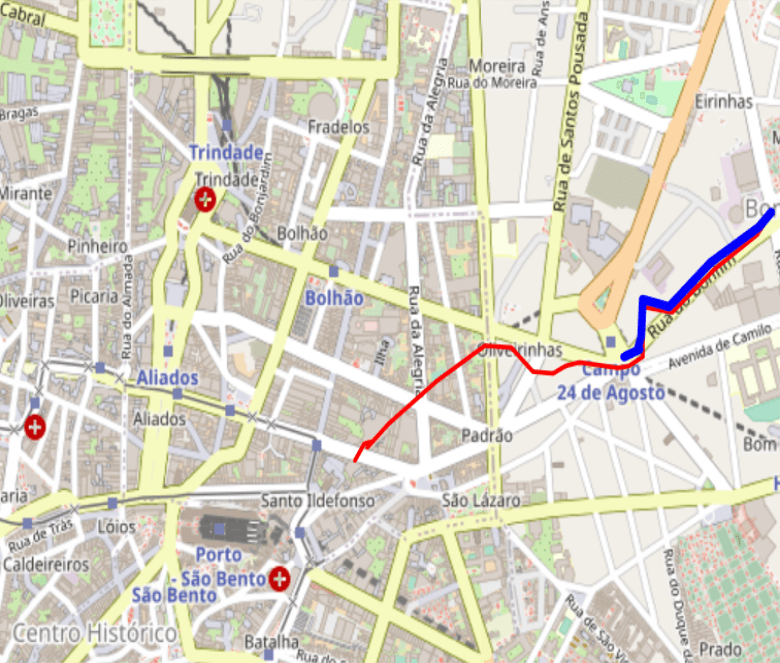}
		\caption{\acrshort{rkn}}
	\end{subfigure}\\
	\caption{Generated 50 taxi trajectories in 3 different areas from \gls{vdm} and the baselines. All models are required to predict the future continuations (red), based the beginning of a trajectory (blue). \gls{vdm} generates more plausible trajectories compared with the baselines. While the generated trajectories from \gls{vdm} follow the street map, the generated trajectories from all baselines are physically impossible. \gls{aesmc} and \gls{cfvae} can capture the general evolving direction, but suffer from capturing the multi-modality at each time step.}
\end{figure}

\begin{figure}[h]
	\centering
	\begin{subfigure}[t]{0.19\textwidth}
		\includegraphics[width=\textwidth]{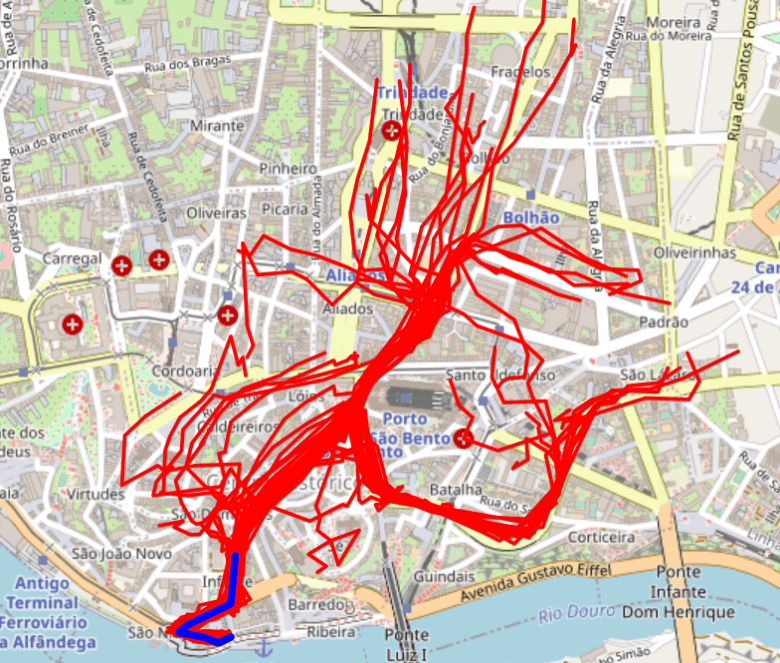}
	\end{subfigure}
	\begin{subfigure}[t]{0.19\textwidth}
		\includegraphics[width=\textwidth]{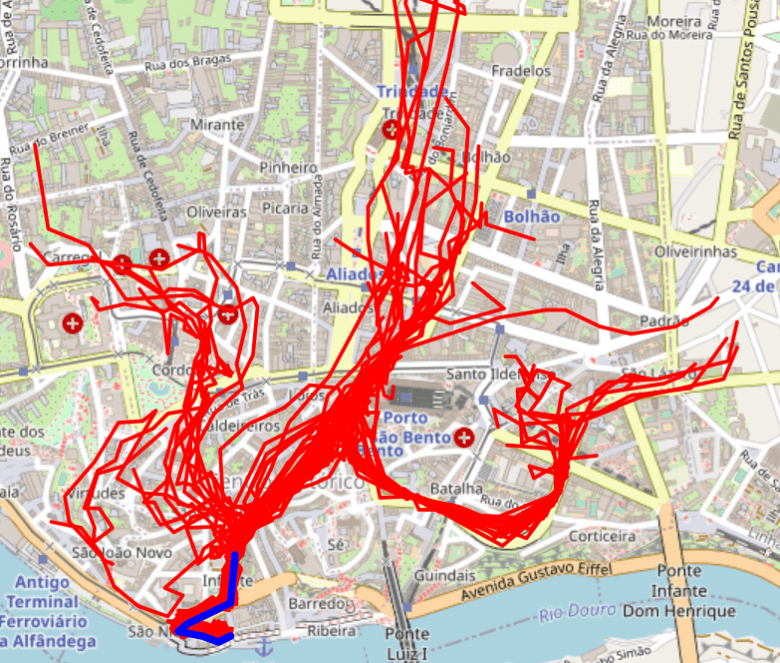}
	\end{subfigure}
	\begin{subfigure}[t]{0.19\textwidth}
		\includegraphics[width=\textwidth]{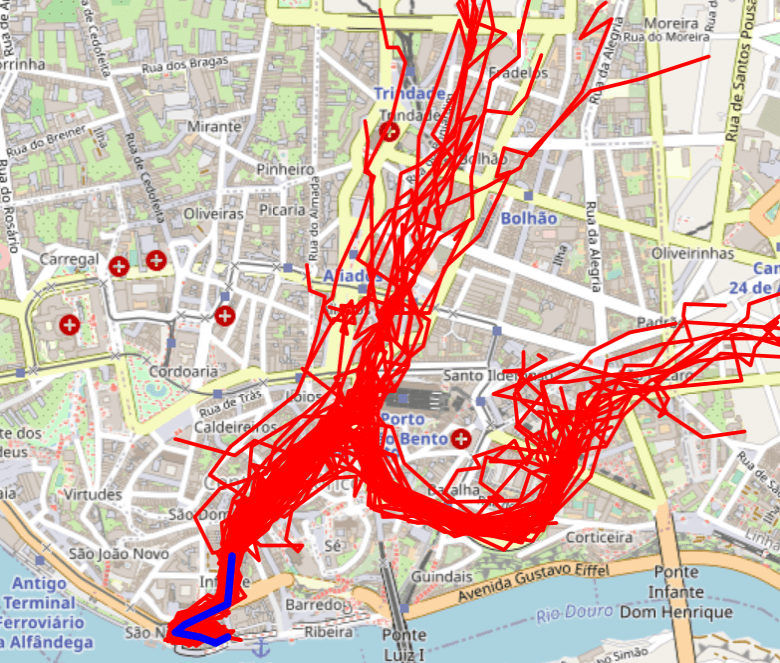}
	\end{subfigure}
	\begin{subfigure}[t]{0.19\textwidth}
		\includegraphics[width=\textwidth]{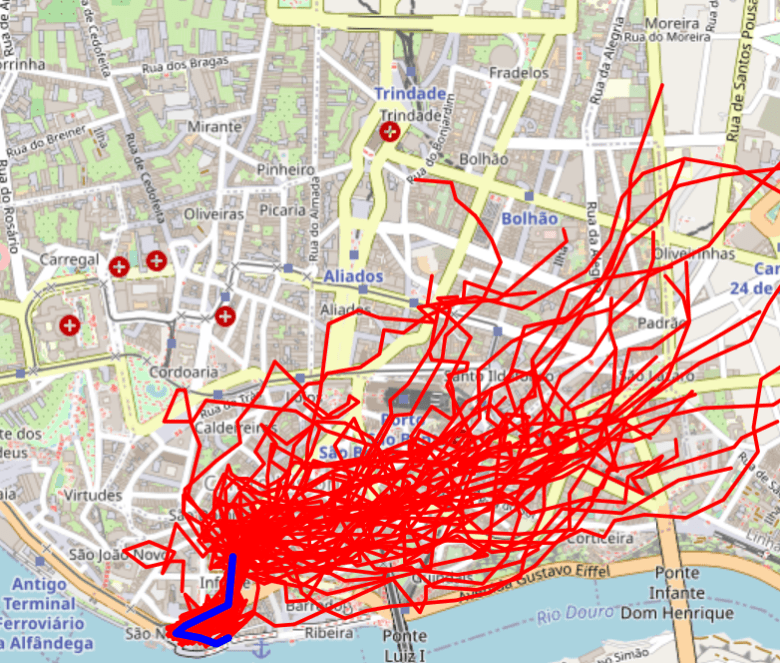}
	\end{subfigure}
	\begin{subfigure}[t]{0.19\textwidth}
		\includegraphics[width=\textwidth]{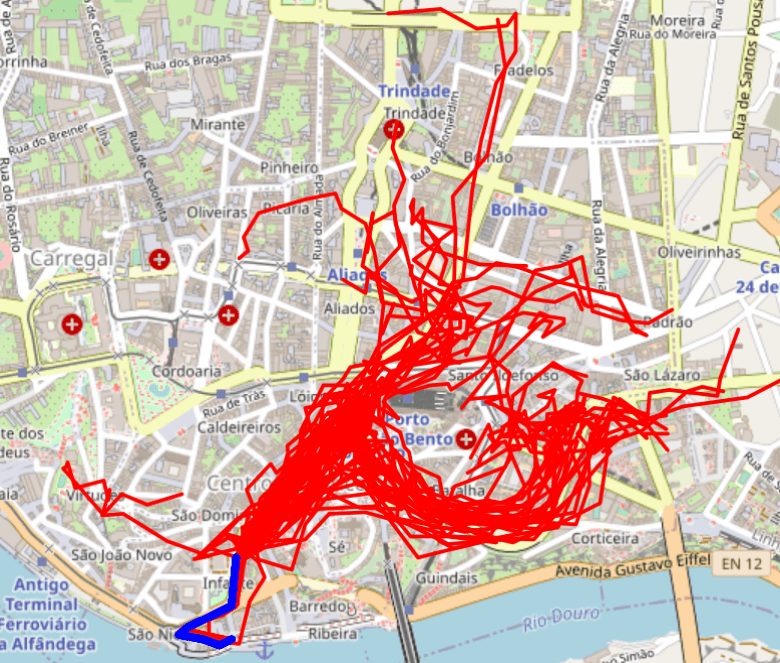}
	\end{subfigure}\\
	\begin{subfigure}[t]{0.19\textwidth}
		\includegraphics[width=\textwidth]{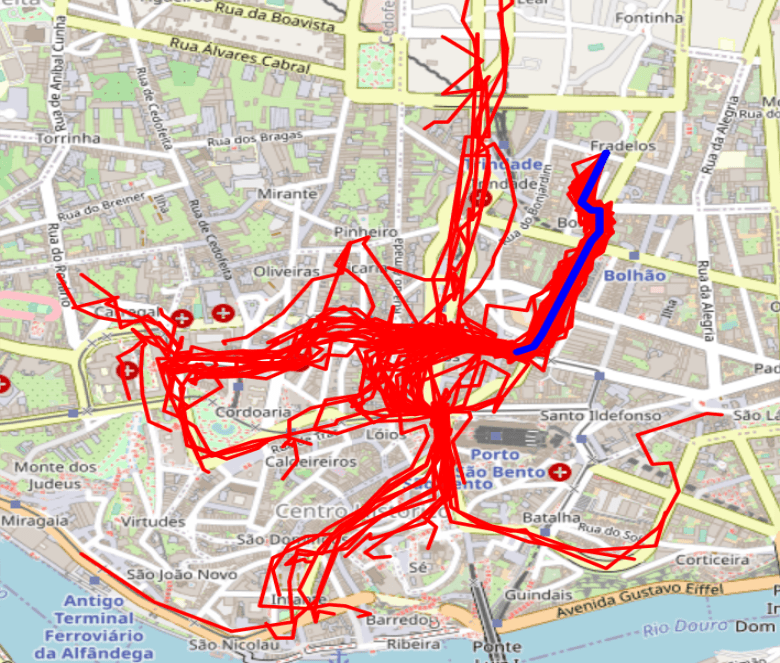}
	\end{subfigure}
	\begin{subfigure}[t]{0.19\textwidth}
		\includegraphics[width=\textwidth]{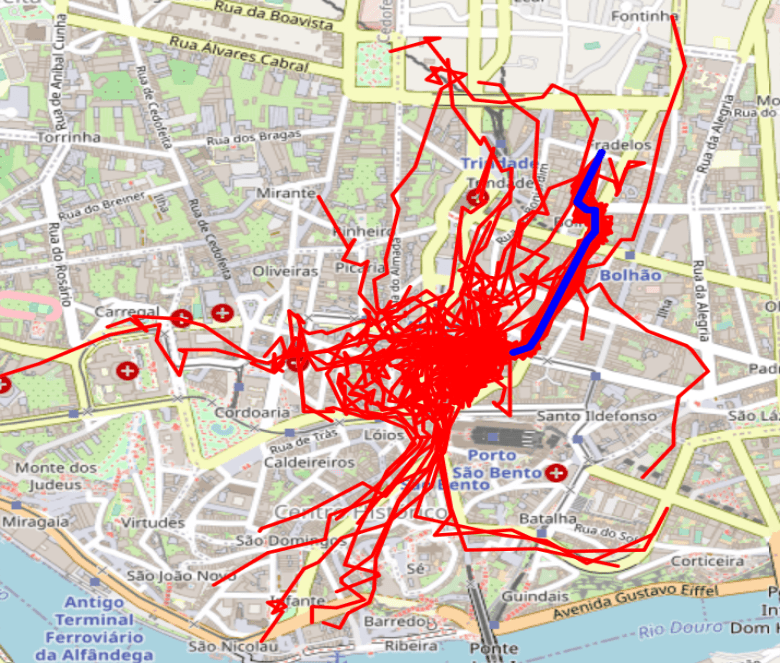}
	\end{subfigure}
	\begin{subfigure}[t]{0.19\textwidth}
		\includegraphics[width=\textwidth]{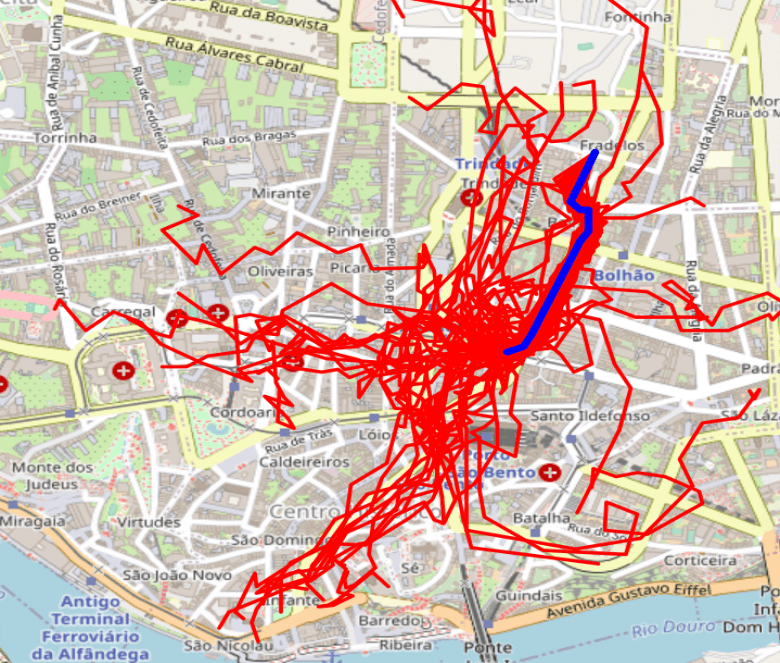}
	\end{subfigure}
	\begin{subfigure}[t]{0.19\textwidth}
		\includegraphics[width=\textwidth]{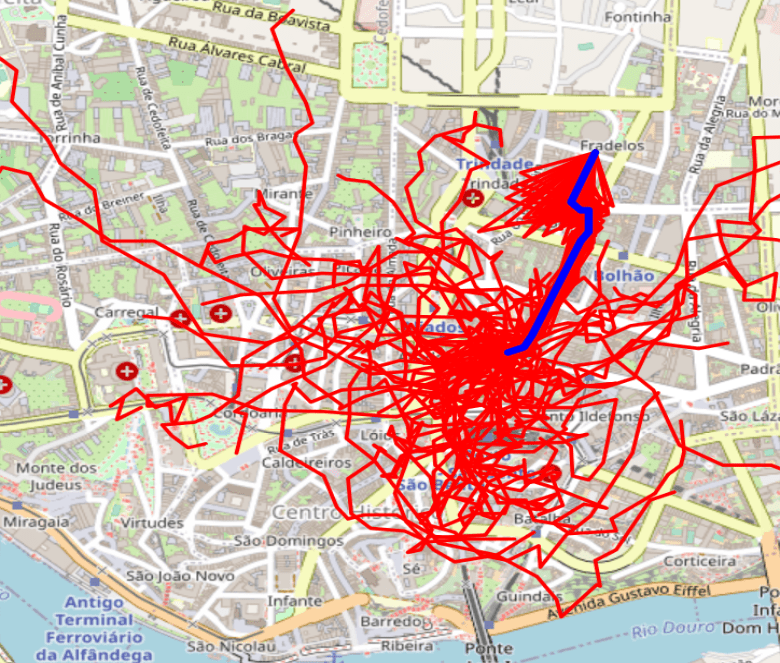}
	\end{subfigure}
	\begin{subfigure}[t]{0.19\textwidth}
		\includegraphics[width=\textwidth]{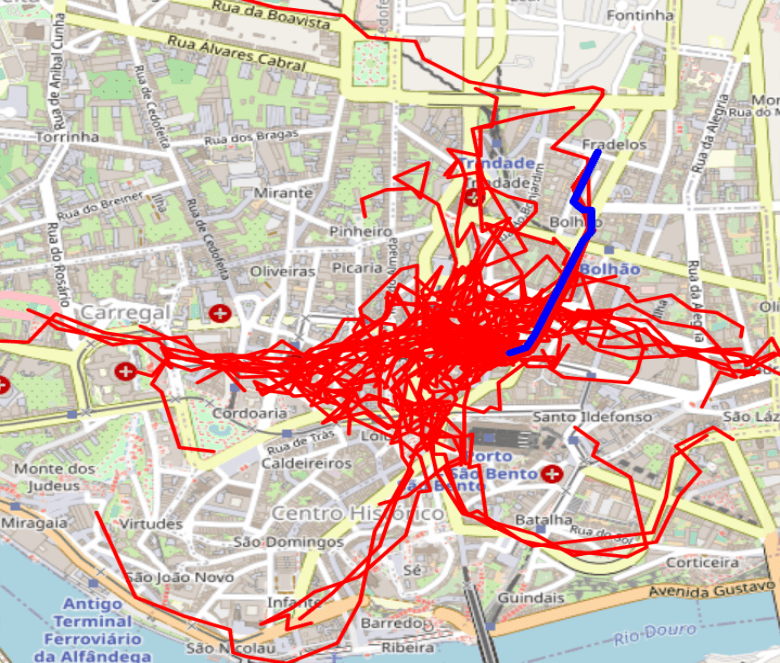}
	\end{subfigure}\\
		\begin{subfigure}[t]{0.19\textwidth}
		\includegraphics[width=\textwidth]{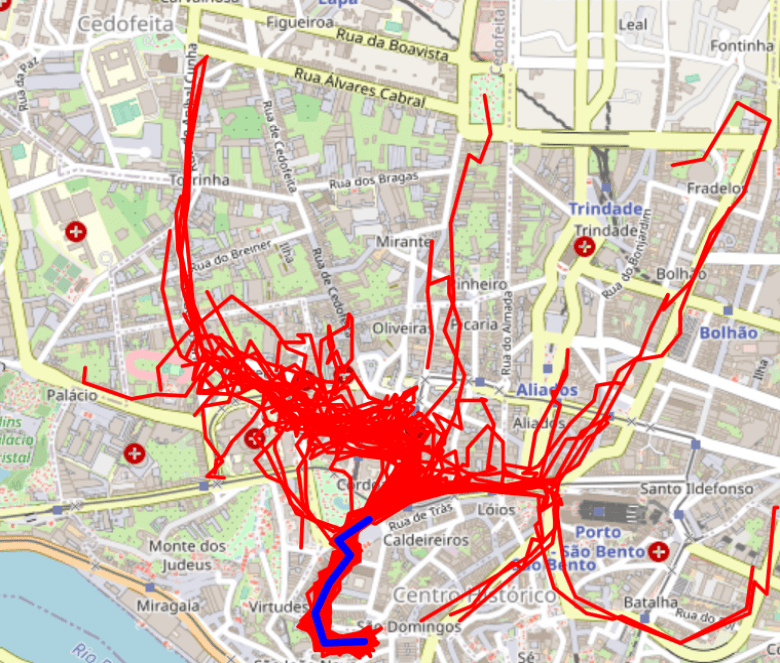}
	\end{subfigure}
	\begin{subfigure}[t]{0.19\textwidth}
		\includegraphics[width=\textwidth]{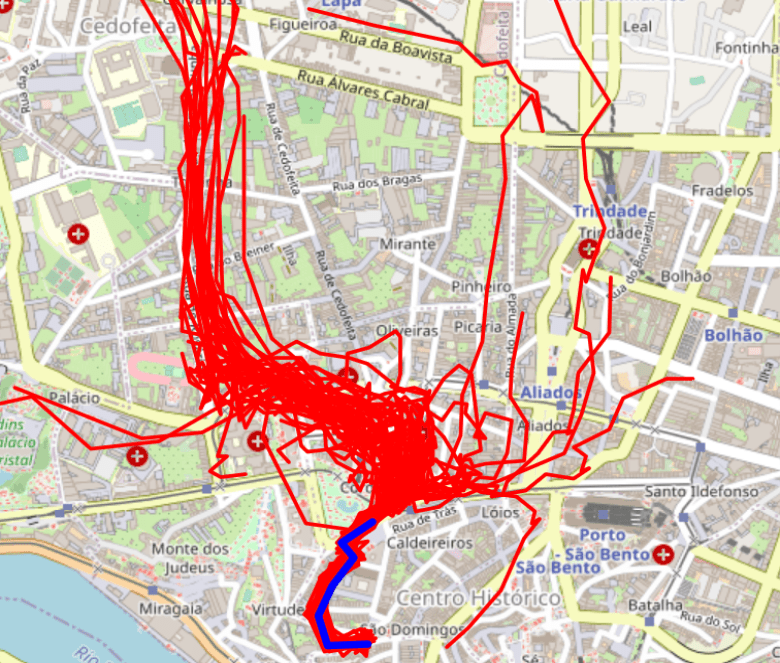}
	\end{subfigure}
	\begin{subfigure}[t]{0.19\textwidth}
		\includegraphics[width=\textwidth]{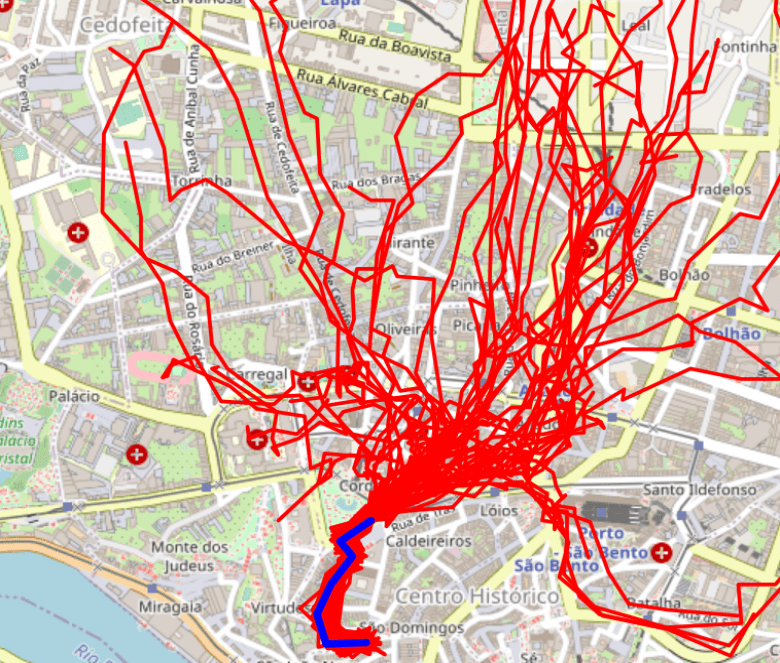}
	\end{subfigure}
	\begin{subfigure}[t]{0.19\textwidth}
		\includegraphics[width=\textwidth]{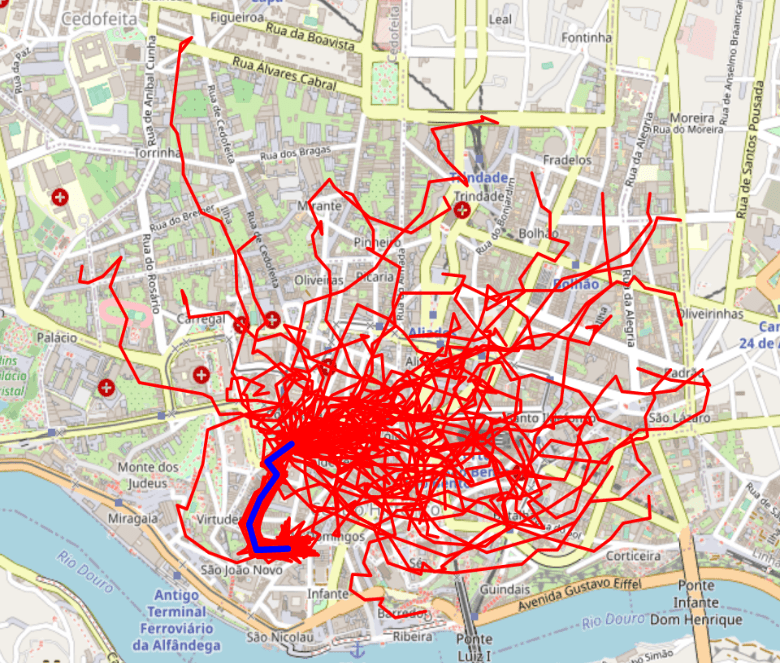}
	\end{subfigure}
	\begin{subfigure}[t]{0.19\textwidth}
		\includegraphics[width=\textwidth]{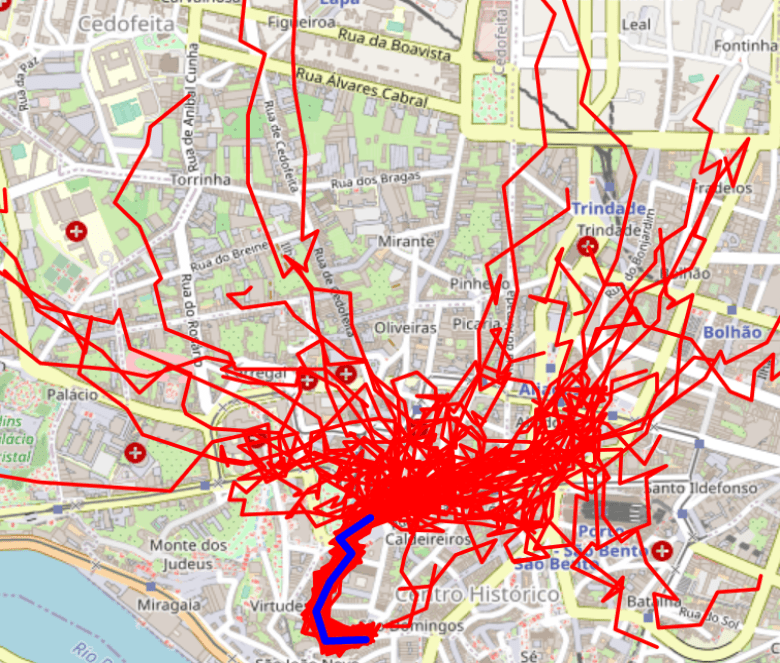}
	\end{subfigure}\\
		\begin{subfigure}[t]{0.19\textwidth}
		\includegraphics[width=\textwidth]{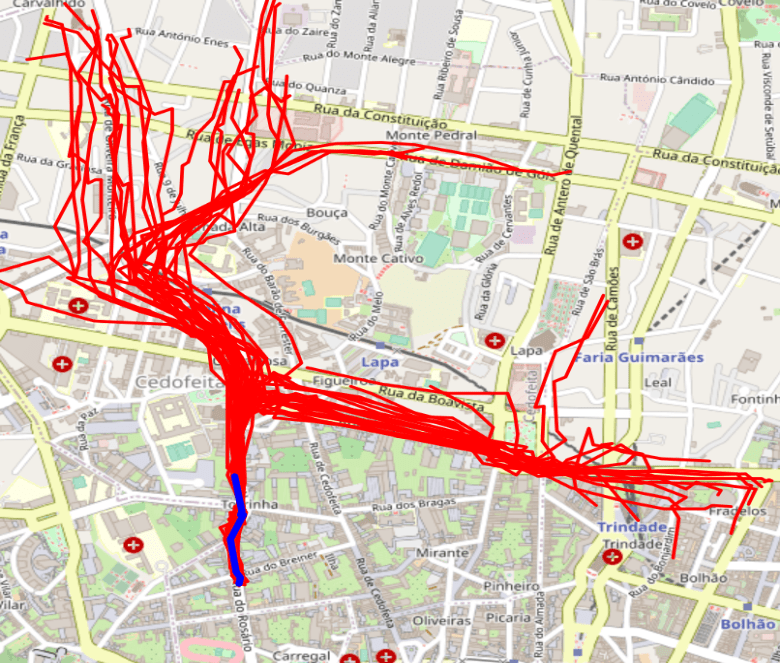}
	\end{subfigure}
	\begin{subfigure}[t]{0.19\textwidth}
		\includegraphics[width=\textwidth]{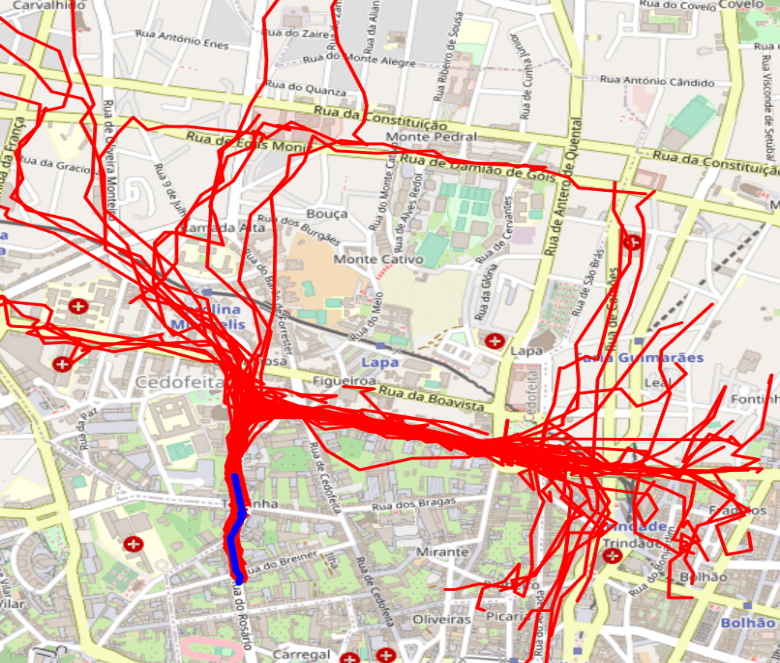}
	\end{subfigure}
	\begin{subfigure}[t]{0.19\textwidth}
		\includegraphics[width=\textwidth]{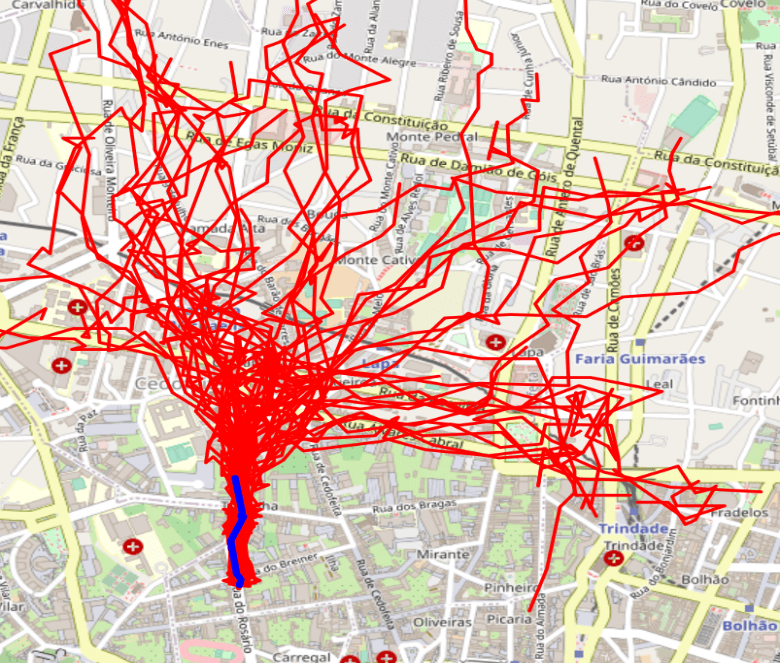}
	\end{subfigure}
	\begin{subfigure}[t]{0.19\textwidth}
		\includegraphics[width=\textwidth]{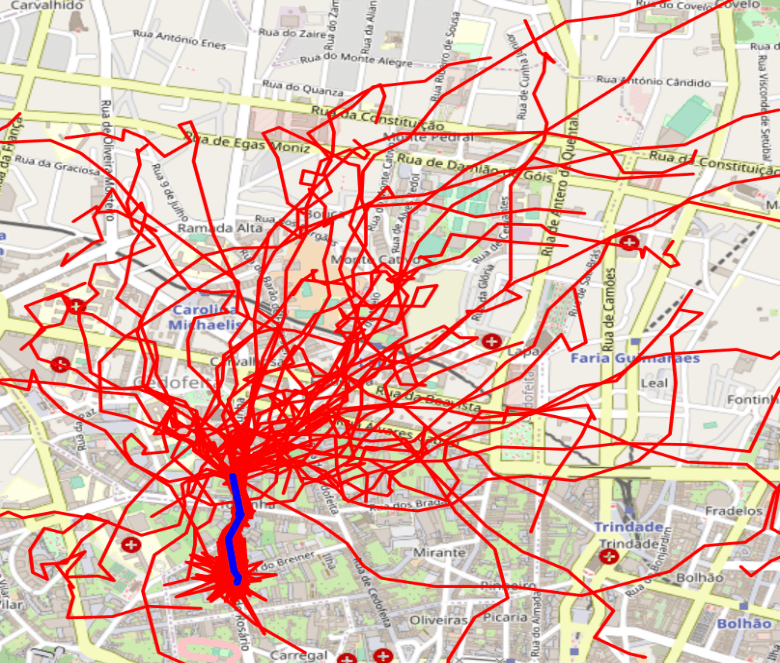}
	\end{subfigure}
	\begin{subfigure}[t]{0.19\textwidth}
		\includegraphics[width=\textwidth]{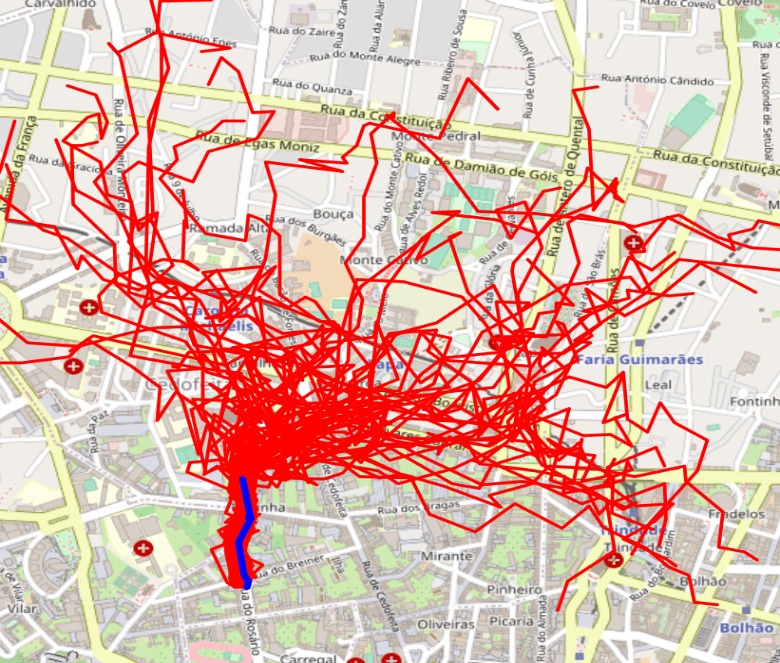}
	\end{subfigure}\\
		\begin{subfigure}[t]{0.19\textwidth}
		\includegraphics[width=\textwidth]{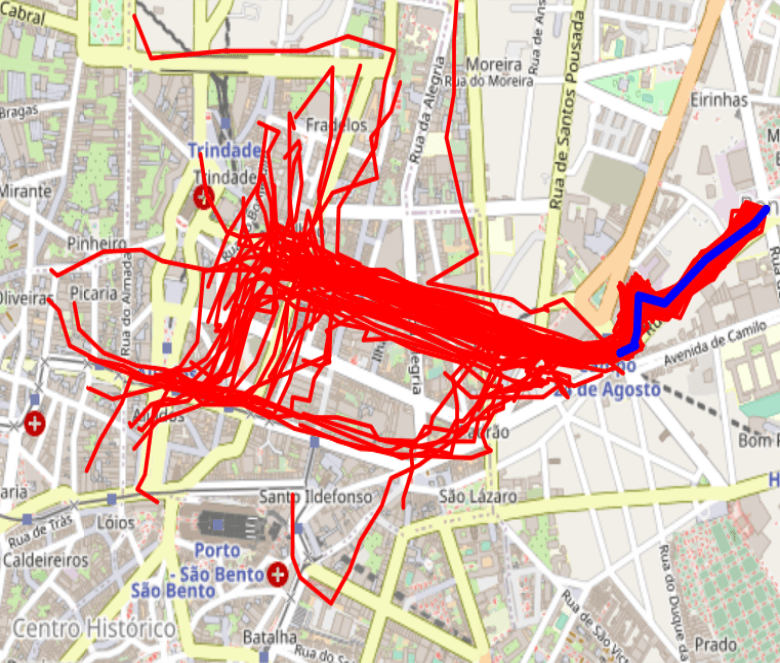}
		\caption{$\text{VDM-SCA+}\delta$}
	\end{subfigure}
	\begin{subfigure}[t]{0.19\textwidth}
		\includegraphics[width=\textwidth]{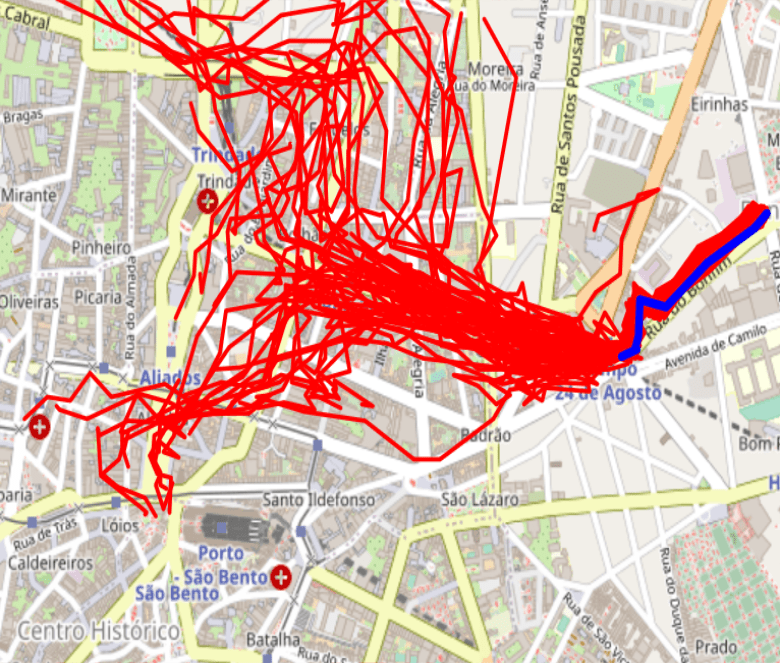}
		\caption{$\text{VDM-SCA+Cat}$}
	\end{subfigure}
	\begin{subfigure}[t]{0.19\textwidth}
		\includegraphics[width=\textwidth]{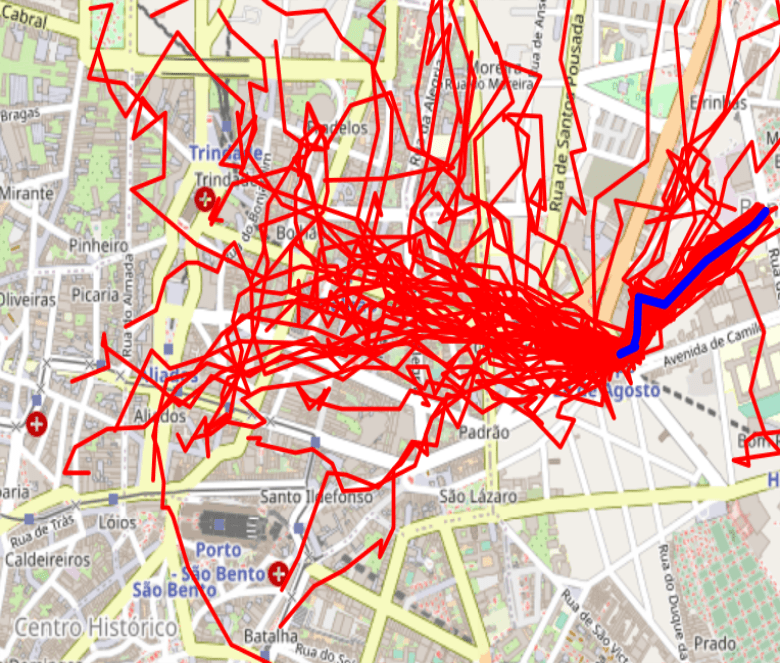}
		\caption{$\text{VDM-MC+}\delta$}
	\end{subfigure}
	\begin{subfigure}[t]{0.19\textwidth}
		\includegraphics[width=\textwidth]{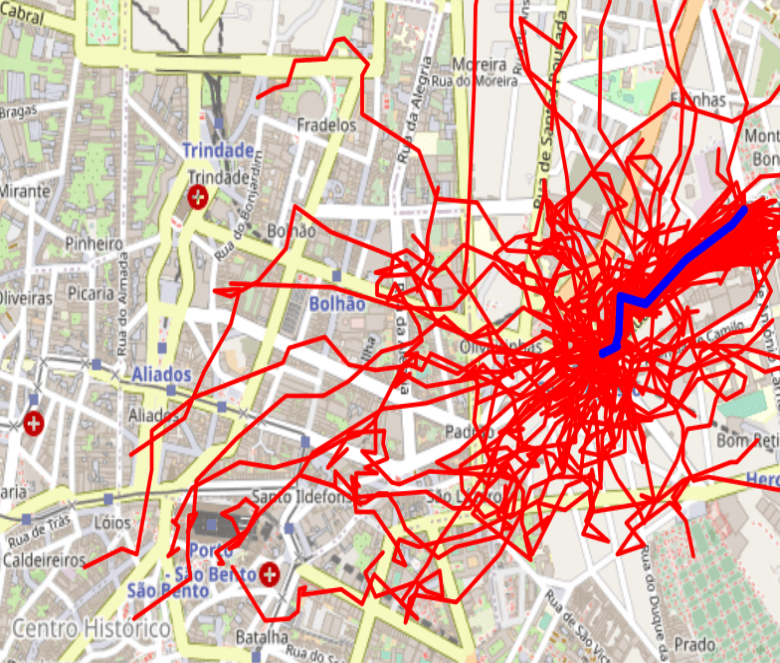}
		\caption{$\text{VDM-Net}$}
	\end{subfigure}
	\begin{subfigure}[t]{0.19\textwidth}
		\includegraphics[width=\textwidth]{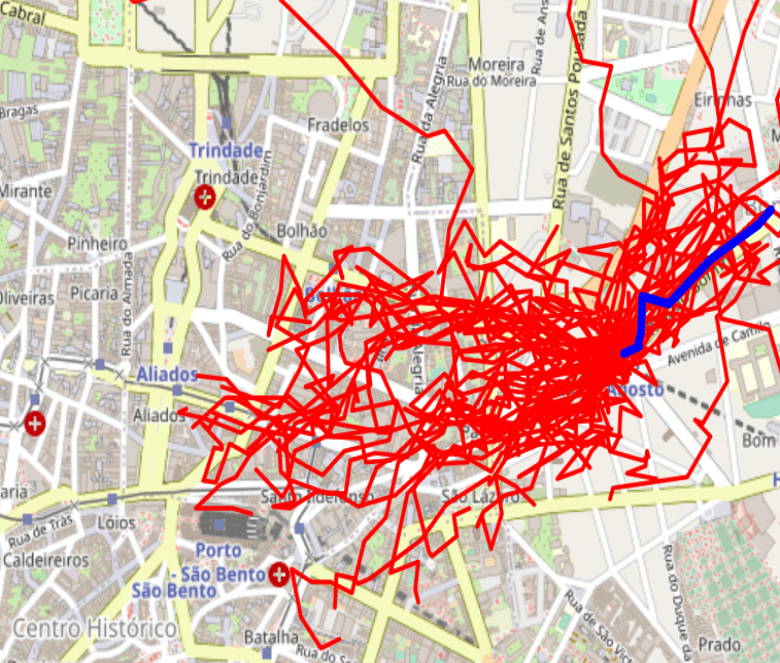}
		\caption{$\text{VDM}(k=1)$}
	\end{subfigure}\\
	\caption{Generated 50 taxi trajectories from \gls{vdm} variants. All models are required to predict the future continuations (red), based the beginning of a trajectory (blue). $\text{VDM-SCA+}\delta$ achieves the best qualitative results among all variants. $\text{VDM-SCA+}\delta$ can generate plausible trajectories, even it is trained without the adversarial term $\mathcal{L}_{adv}$. We can see, for the weighting function, \cref{eqn:omega_1} is better than \cref{eqn:omega_2}, and for the sampling method, \gls{sca} is better than Monte-Carlo method.}
\end{figure}